%% file: colm2024_conference.tex
\documentclass{article} 
\usepackage{colm2024_conference}

\usepackage{microtype}
\usepackage{hyperref}
\usepackage{url}
\usepackage{booktabs}
\usepackage{wrapfig} 
\usepackage{graphicx}
\usepackage{caption}
\usepackage{multirow}
\usepackage{diagbox}
\usepackage{makecell}
\definecolor{darkblue}{rgb}{0, 0, 0.5}
\hypersetup{colorlinks=true, citecolor=darkblue, linkcolor=darkblue, urlcolor=darkblue}

\usepackage[utf8]{inputenc} %
\usepackage[T1]{fontenc}    %
\usepackage{url}            %
\usepackage{amsfonts}       %
\usepackage{nicefrac}       %
\usepackage{microtype}      %
\usepackage{colortbl}
\usepackage{amsmath}
\usepackage[most]{tcolorbox}
\usepackage{dialogue}
\usepackage{pifont}%

\usepackage{siunitx}
\usepackage{graphicx}

\usepackage{wrapfig}
\usepackage{enumitem}
\usepackage{xspace}
\usepackage{adjustbox}
\usepackage{pifont}
\usepackage{caption}
\usepackage{makecell}
\usepackage{subcaption}
\usepackage{hyperref}

\usepackage{nicematrix}
\usepackage{listings}

\usepackage{tikz}

\renewcommand{\paragraph}[1]{\noindent \textbf{#1}}

\usepackage{xcolor}
\definecolor{sailor2lightblue}{HTML}{d8eafb}

\newcommand{\huggingface}{\raisebox{-1.5pt}{\includegraphics[height=1.05em]{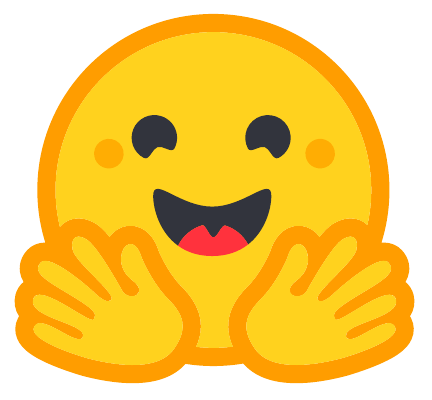}}\xspace}
\newcommand{\github}{\raisebox{-1.5pt}{\includegraphics[height=1.05em]{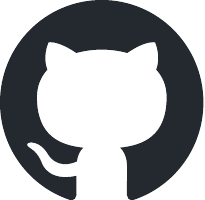}}\xspace}

\usepackage{nicematrix}

\newcommand{\githubreponologo}[1]{%
  \href{https://www.github.com/#1}{\small{\texttt{#1}}}%
}

\newcommand{\huggingfacemodeltinynologo}[1]{%
  \href{https://huggingface.co/#1}{\footnotesize{\texttt{#1}}}%
}

\newcommand{\huggingfacedatasettinynologo}[1]{%
  \href{https://huggingface.co/datasets/#1}{\footnotesize{\texttt{\detokenize{#1}}}}%
}

\usepackage{longtable}
\usepackage{array}
\usepackage{xtab}
\usepackage{arydshln}

\newcolumntype{L}[1]{>{\raggedright\let\newline\\\arraybackslash\hspace{0pt}}m{#1}}
\newcolumntype{C}[1]{>{\centering\let\newline\\\arraybackslash\hspace{0pt}}m{#1}}
\newcolumntype{R}[1]{>{\raggedleft\let\newline\\\arraybackslash\hspace{0pt}}m{#1}}
\newcolumntype{P}[1]{>{\centering\let\newline\\\arraybackslash\columncolor{sailor2lightblue}}m{#1}}
\newcolumntype{W}[1]{>{\columncolor{white}}c}  %

\newcommand{\pplup}[1]{\textcolor{red}{\scriptsize $\uparrow${#1}}}
\definecolor{fGreen}{RGB}{34,139,34}
\newcommand{\ppldown}[1]{\textcolor{fGreen}{\scriptsize $\downarrow${#1}}}

\NewDocumentCommand\emojisailor{}{
        \includegraphics[scale=0.025]{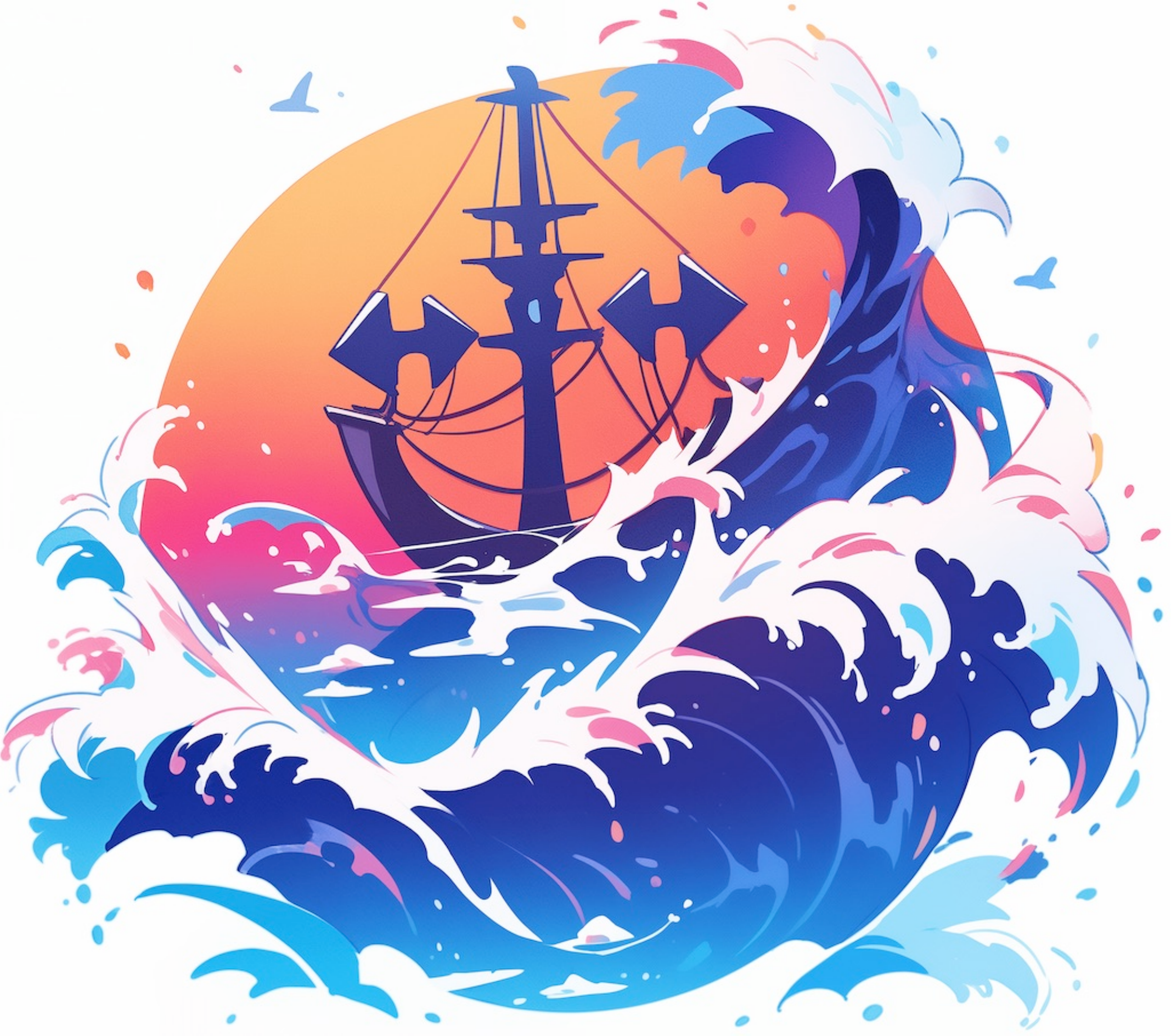}
}

\title{\emojisailor Sailor2: Sailing in South-East Asia with Inclusive \\ Multilingual LLMs}

\input{author}

\colmfinalcopy 
\begin{document}

\maketitle
\begin{abstract}
\vspace{-0.3cm}
Sailor2 is a family of cutting-edge multilingual language models for South-East Asian (SEA) languages, available in 1B, 8B, and 20B sizes to suit diverse applications. 
Building on Qwen2.5, Sailor2 undergoes continuous pre-training on 500B tokens (400B SEA-specific and 100B replay tokens) to support 13 SEA languages while retaining proficiency in Chinese and English. 
Sailor2-20B model achieves a 50-50 win rate against GPT-4o across  SEA languages.
We also deliver a comprehensive cookbook on how to develop the multilingual model in an efficient manner, including five key aspects: data curation, pre-training, post-training, model customization and evaluation. 
We hope that Sailor2 model (Apache 2.0 license) will drive language development in the SEA region, and Sailor2 cookbook will inspire researchers to build more inclusive LLMs for other under-served languages.


\end{abstract}
\vspace{-0.3cm}

\begin{figure}[htbp]
    \centering
    \includegraphics[width=1.0\textwidth]{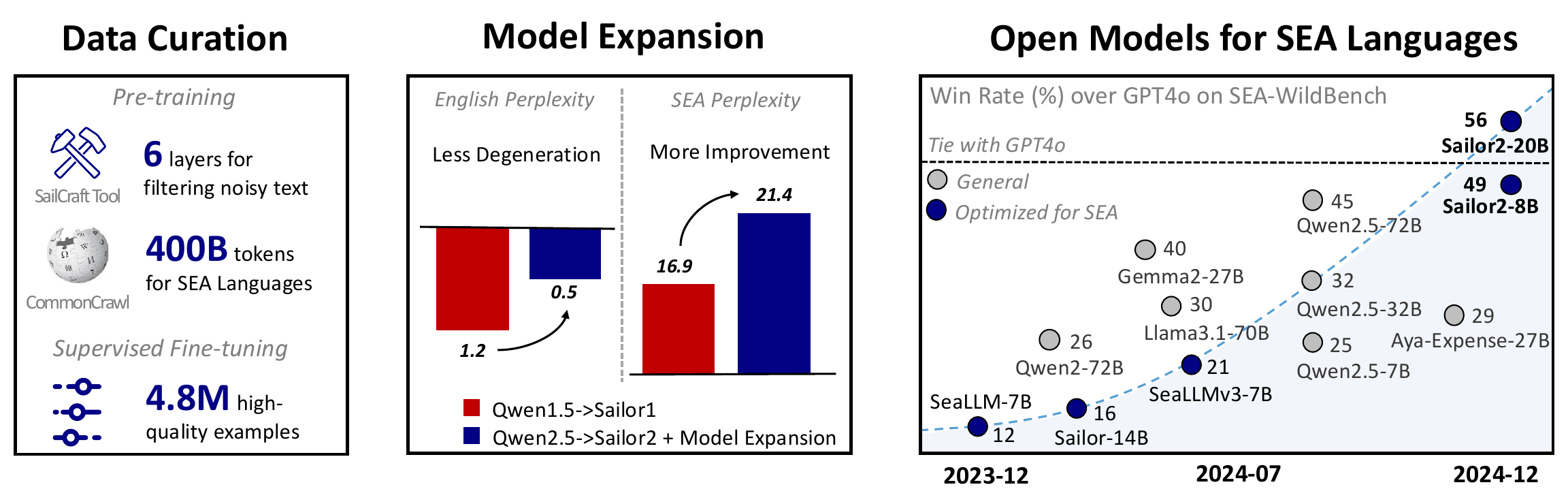}
    \caption{
    With rigorous data curation and efficient model expansion, 
    Sailor2-20B achieves the 50-50 win rate over GPT4o on SEA languages, marking a new milestone of open LLMs. 
    }
    \label{fig:sailor2_figure1}
\end{figure}

\newpage
\tableofcontents
\newpage

\newpage
\input{tables/artifacts}
\newpage

\input{sections/1.intro}
\input{sections/2.related_work}
\input{sections/3.data_cruation}
\input{sections/4.model_pre_training}

\input{sections/5.model_post_training}
\input{sections/6.model_customization}

\input{sections/7.evaluation}
\input{sections/8.analysis}

\input{sections/9.conclusion}

\newpage
\input{appendices/contribution}

\newpage

\bibliography{colm2024_conference}
\bibliographystyle{colm2024_conference}

\newpage
\input{appendices/prompt}
\newpage
\input{appendices/ppl_shift_figure}
\input{appendices/culture_eval}

\end{document}

%% file: author.tex
\author{\textbf{Longxu Dou}$^1$\thanks{Equal Contributors.  
Contact: \href{mailto:doulx@sea.com}{doulx@sea.com}, \href{mailto:doulx@sea.com}{liuqian.sea@gmail.com}}\quad
\textbf{Qian Liu}$^{1*}$\thanks{Qian Liu is the project leader of Sailor2.}\quad
\textbf{Fan Zhou}$^{6*}$\quad
\textbf{Changyu Chen}$^{7*}$\quad
\textbf{Zili Wang}\quad 
\textbf{Ziqi Jin}$^{5}$\quad \\
\textbf{Zichen Liu}$^{1,8}$\quad
\textbf{Tongyao Zhu}$^{1,8}$\quad
\textbf{Cunxiao Du}$^1$\quad
\textbf{Penghui Yang}$^{9}$\quad 
\textbf{Haonan Wang}$^{8}$\quad \\
\textbf{Jiaheng Liu}\quad 
\textbf{Yongchi Zhao}\quad 
\textbf{Xiachong Feng}$^{10}$\quad
\textbf{Xin Mao}$^{9}$\quad 
\textbf{Man Tsung Yeung}$^{8}$\quad  \\
\textbf{Kunat Pipatanakul}$^2$\quad
\textbf{Fajri Koto}$^{13}$\quad
\textbf{Min Si Thu}$^{12}$\quad
\textbf{Hynek Kydlíček} $^{4}$\quad
\textbf{Zeyi Liu}$^{11}$\quad \\
\textbf{Qunshu Lin}$^{11}$\quad 
\textbf{Sittipong Sripaisarnmongkol}$^2$\quad
\textbf{Kridtaphad Sae-Khow}$^3$\quad \\
\textbf{Nirattisai Thongchim}$^3$\quad 
\textbf{Taechawat Konkaew}$^3$\quad
\textbf{Narong Borijindargoon}$^3$\quad 
\textbf{Anh Dao}$^{14}$\quad \\
\textbf{Matichon Maneegard}$^{15}$\quad 
\textbf{Phakphum Artkaew}$^{16}$\quad 
\textbf{Zheng-Xin Yong}$^{17}$\quad 
\textbf{Quan Nguyen}$^{18}$\quad  \\
\textbf{Wannaphong Phatthiyaphaibun}$^{19}$\quad 
\textbf{Hoang H. Tran}$^{20}$\quad 
\textbf{Mike Zhang}$^{21}$\quad 
\textbf{Shiqi Chen}$^{22}$\quad  \\
\textbf{Tianyu Pang}$^1$\quad
\textbf{Chao Du}$^1$\quad
\textbf{Xinyi Wan}$^1$\quad 
\textbf{Wei Lu}$^5$\quad
\textbf{Min Lin}$^1$\quad \\ \\
$^{1}$Sea AI Lab \quad
$^{2}$SCB 10X \quad 
$^{3}$WiseSight \quad
$^{4}$Hugging Face \quad 
$^{5}$SUTD \quad 
$^{6}$SJTU \quad 
$^{7}$SMU \quad \\
$^{8}$NUS \quad 
$^{9}$NTU \quad
$^{10}$HKU \quad
$^{11}$ABAKA AI \quad
$^{12}$Peafowl.ai \quad
$^{13}$MBZUAI \quad \\
$^{14}$Michigan State University \quad 
$^{15}$Float16.cloud \quad 
$^{16}$NYU \quad 
$^{17}$Brown University \quad \\
$^{18}$Umeå University \quad
$^{19}$PyThaiNLP \quad 
$^{20}$HCMUT \quad
$^{21}$Aalborg University \quad
$^{22}$CityU \quad
\\ \\
\textbf{Home Page}: \url{https://sea-sailor.github.io/blog/sailor2/}
}

%% file: tables/artifacts.tex
\begin{table}[h!]
\centering

\caption{Models, resource, and code released with Sailor2 under Apache 2.0 License. \\\vspace{3pt}\textbf{Demo}: \url{https://huggingface.co/spaces/sail/Sailor2-20B-Chat}
}
\vspace{3pt}
\label{tab:artifacts}
\setlength\tabcolsep{5pt}

{\small

\begin{tabular}{L{60pt}L{90pt}L{90pt}L{110pt}}
\multicolumn{4}{c}{\normalsize \textbf{Model Checkpoints}} \\
\toprule
\textbf{Stage} & \textbf{Sailor2-1B} & \textbf{Sailor2-8B} & \textbf{Sailor2-20B} \\
\midrule
\rowcolor{sailor2lightblue} Pre-Annealing & \huggingfacemodeltinynologo{sail/Sailor2-1B-Pre} & \huggingfacemodeltinynologo{sail/Sailor2-8B-Pre}  & \huggingfacemodeltinynologo{sail/Sailor2-20B-Pre}\\
Base & \huggingfacemodeltinynologo{sail/Sailor2-1B} & \huggingfacemodeltinynologo{sail/Sailor2-8B}  & \huggingfacemodeltinynologo{sail/Sailor2-20B}\\
\rowcolor{sailor2lightblue} SFT & \huggingfacemodeltinynologo{sail/Sailor2-1B-SFT} & \huggingfacemodeltinynologo{sail/Sailor2-8B-SFT}  & \huggingfacemodeltinynologo{sail/Sailor2-20B-SFT} \\
Chat & \huggingfacemodeltinynologo{sail/Sailor2-1B-Chat} & \huggingfacemodeltinynologo{sail/Sailor2-8B-Chat} & \huggingfacemodeltinynologo{sail/Sailor2-20B-Chat} \\
\bottomrule
\end{tabular}
\vspace{5pt}

\begin{tabular}{L{120pt}L{250pt}}
\multicolumn{2}{c}{\normalsize\textbf{Codebases / Tools}} \\
\toprule
\textbf{Type}  & \textbf{\small{\github} Link} \\
\midrule
\rowcolor{sailor2lightblue} Data Cleaning & \githubreponologo{sail-sg/sailcraft} \\
Data Mixture & \githubreponologo{sail-sg/regmix} \\
\rowcolor{sailor2lightblue} Pre-training & \githubreponologo{sail-sg/Megatron-Sailor2} \\
Post-training & \githubreponologo{sail-sg/oat} \\
\rowcolor{sailor2lightblue} Evaluation & \githubreponologo{sail-sg/sailcompass} \\
\bottomrule
\end{tabular}
\vspace{5pt}

\begin{tabular}{L{120pt}L{250pt}}
\multicolumn{2}{c}{\normalsize\textbf{Post-Training Dataset}} \\
\toprule
\textbf{Domain} & \textbf{\small{\huggingface} Link} \\
\midrule
 \rowcolor{sailor2lightblue} SFT-Stage1 &  \huggingfacedatasettinynologo{sailor2/sailor2-sft-stage1} \\
SFT-Stage2 & \huggingfacedatasettinynologo{sailor2/sailor2-sft-stage2} \\
\rowcolor{sailor2lightblue} Off-policy DPO &  \huggingfacedatasettinynologo{sailor2/sea-ultrafeedback} \\
On-policy DPO &  \huggingfacedatasettinynologo{sailor2/sea-ultrafeedback-onpolicy} \\
\bottomrule
\end{tabular}
\vspace{5pt}

\begin{tabular}{L{120pt}L{250pt}}
\multicolumn{2}{c}{\normalsize\textbf{Evaluation Dataset}} \\
\toprule
\textbf{Domain} & \textbf{\small{\huggingface} Link} \\
\midrule
 \rowcolor{sailor2lightblue} SailCompass &  \huggingfacedatasettinynologo{sail/Sailcompass_data} \\
SEA-WildBench & \huggingfacedatasettinynologo{sailor2/sea-wildbench} \\
\bottomrule
\end{tabular}
\vspace{5pt}

\begin{tabular}{L{45pt}L{100pt}L{100pt}L{105pt}}
\multicolumn{4}{c}{\normalsize \textbf{Model Checkpoints (via Long-Context Training)}} \\
\toprule
\textbf{Stage} & \textbf{Sailor2-1B} & \textbf{Sailor2-8B} & \textbf{Sailor2-20B} \\
\midrule
\rowcolor{sailor2lightblue} Base & \huggingfacemodeltinynologo{sail/Sailor2-L-1B} & \huggingfacemodeltinynologo{sail/Sailor2-L-8B}  & \huggingfacemodeltinynologo{sail/Sailor2-L-20B}\\
SFT & \huggingfacemodeltinynologo{sail/Sailor2-L-1B-SFT} & \huggingfacemodeltinynologo{sail/Sailor2-L-8B-SFT}  & \huggingfacemodeltinynologo{sail/Sailor2-L-20B-SFT} \\
\rowcolor{sailor2lightblue} Chat & \huggingfacemodeltinynologo{sail/Sailor2-L-1B-Chat} & \huggingfacemodeltinynologo{sail/Sailor2-L-8B-Chat} & \huggingfacemodeltinynologo{sail/Sailor2-L-20B-Chat} \\
\bottomrule
\end{tabular}
\vspace{5pt}

\begin{tabular}{L{60pt}L{150pt}L{150pt}}
\multicolumn{3}{c}{\normalsize \textbf{Model Checkpoints (via Speculative Decoding)}} \\
\toprule
\textbf{Stage} & \textbf{Sailor2-8B} & \textbf{Sailor2-20B} \\
\midrule
\rowcolor{sailor2lightblue} Base Model & \huggingfacemodeltinynologo{sail/Sailor2-8B-Chat-Glide} & \huggingfacemodeltinynologo{sail/Sailor2-20B-Chat-Glide} \\
\bottomrule
\end{tabular}
\vspace{5pt}

\begin{tabular}{L{60pt}L{150pt}L{150pt}}
\multicolumn{3}{c}{\normalsize \textbf{Model Checkpoints (via Model Pruning)}} \\
\toprule
\textbf{Stage} & \textbf{Sailor2-3B (Pruning via Sailor2-8B)} & \textbf{Sailor2-14B (Pruning via Sailor2-20B)} \\
\midrule
\rowcolor{sailor2lightblue} Base Model & \huggingfacemodeltinynologo{sail/Sailor2-3B} & \huggingfacemodeltinynologo{sail/Sailor2-14B} \\
SFT & \huggingfacemodeltinynologo{sail/Sailor2-3B-SFT}  & \huggingfacemodeltinynologo{sail/Sailor2-14B-SFT} \\
\rowcolor{sailor2lightblue} Chat & \huggingfacemodeltinynologo{sail/Sailor2-3B-Chat} & \huggingfacemodeltinynologo{sail/Sailor2-14B-Chat} \\
\bottomrule
\end{tabular}
\vspace{5pt}

}

\end{table}

%% file: sections/1.intro.tex
\section{Introduction}

{\centering \textit{Serving the Underserved in Southeast Asia with Open LLMs. -- Sailor2 Spirit} \par}

Large language model (LLM) technology has driven significant innovations but remains predominantly focused on major languages like English and Chinese, leaving many others underrepresented. 
As a linguistically diverse region with 11 countries and 675 million people, Southeast Asia presents a unique opportunity for advancing multilingual NLP research. 
In this paper, we introduce Sailor2, a contribution to the advancement of language technology in the SEA region. Sailor2 offers improved open models, open tools, a transparent training recipe, and valuable insights to drive progress in multilingual LLMs.

To optimize Sailor2, we apply the following techniques:
\begin{itemize}
    \item Rigorous data deduplication with six layers.
    \item Model expansion to mitigate language degeneration.
    \item Two-stage continual pre-training with varying language compositions.
    \item Two-stage instruction tuning with reward-aware and ppl-aware data selection.
    \item Preference tuning on both off-policy and on-policy data.
\end{itemize}

We devoted significant effort to evaluation, which includes: (1) few-shot evaluation for the base model, (2) chat performance comparison with GPT-4, and (3) cultural understanding about SEA cuisine and traditions.
The results indicate that Sailor2 excels at both basic language tasks (e.g., question answering and translation) and advanced tasks (e.g., mathematics and creative writing).

Overall, the Sailor2 project contributes to the following outcomes:
\begin{itemize} 
    \item \textbf{A family of open models},  optimized for Southeast Asian (SEA) languages.
    \item \textbf{A comprehensive cookbook} detailing the process of building multilingual LLMs, covering data curation, model training, and thorough evaluation. 
\end{itemize}

\begin{figure}[htbp]
    \centering
    \includegraphics[width=1.0\textwidth]{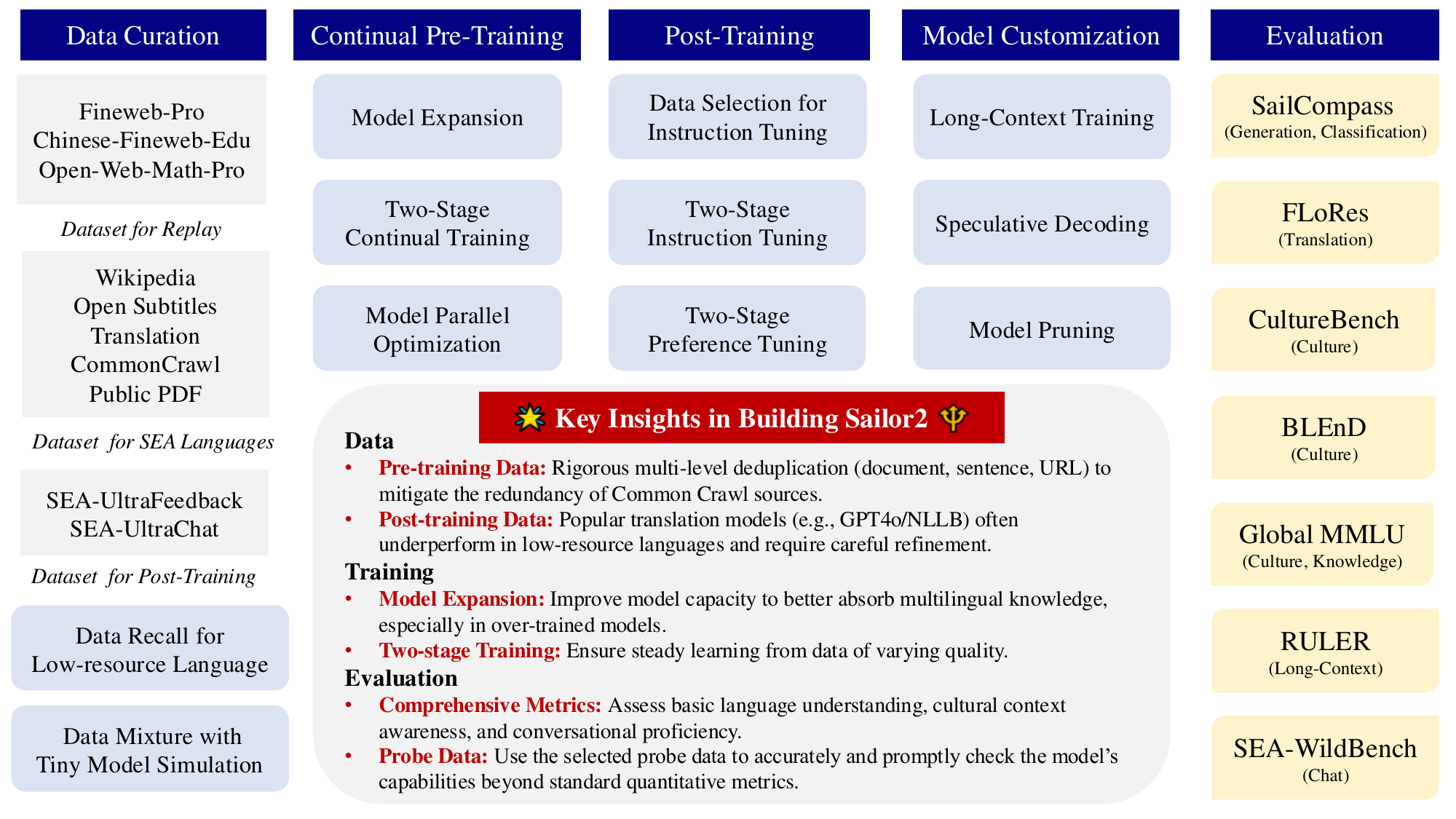}
    \caption{
    Sailor2 Cookbook with key insights in data, model training and evaluation.
    }
    \label{fig:sailor_cookbook}
\end{figure}

%% file: sections/2.related_work.tex
\section{Related Works}

\subsection{Open SEA Language Models}

Open science has gained increasing attention, particularly with the thriving efforts in developing open language models. 
While notable initiatives like OLMo~\citep{groeneveld2024olmo}, LLM360~\citep{liu2023llm360}, and MAP-Neo~\citep{zhang2024map} have made significant contributions, they primarily focus on dominant languages on the Internet, such as English and Chinese.
The Aya model~\citep{ustun-etal-2024-ayamodel} serves as a massively multilingual language model, supporting 101 languages, beats previous multilingual models such as BloomZ~\citep{muennighoff2022crosslingual}, yet not particularly expert in South-East Asian (SEA) languages.
Although there has been some recent progress in creating SEA language models, open initiatives such as the SeaLLM series~\citep{nguyen-etal-2024-seallms, zhang2024seallms} and Sea-LION series~\citep{sea_lion_2024} still fall short of achieving performance levels comparable to commercial models, such as GPT-4o~\citep{achiam2023gpt4}.

Starting in March 2024, we have continuously released both Sailor and Sailor2. 
We are committed to building a fully open pipeline for the entire LLM ecosystem while striving to achieve top-tier SEA language performance. 
In the future, we will continue refining the Sailor series models to advance open language models for more low-resource languages.

\subsection{Open SEA Language Resources}
\label{sec:related_work_resource}

Resources for SEA languages remain underdeveloped.
\textbf{Pre-training}: Even the recent Fineweb2 Dataset~\citep{penedo2024fineweb-2}, which scales the pre-training corpus to over 1,000 languages, provides a significantly smaller data volume for SEA languages compared to others, falling short of the 100B tokens. Moreover, directly translating English resources into local languages often leads to an overestimation of performance, as these translations typically lack culturally nuanced content~\citep{singh2024global}.
\textbf{Post-training}: The Aya dataset~\citep{singh2024aya} is the largest multilingual instruction fine-tuning resource, containing 513 million instances across 114 languages. It comprises mainly machine-translated data with a small, essential human-curated subset.
\textbf{Evaluation}: Although benchmarks such as SeaBench~\citep{liu2025seaexam}, SeaCrowd~\citep{lovenia2024seacrowd}, and SeaEval~\citep{SeaEval2023} have been introduced, they remain limited in either language coverage, primarily focusing on Thai, Indonesian, Vietnamese, and Malay, or in dataset quality due to reliance on machine translations.

In the Sailor2 project, we open source the SailCraft scripts for SEA-language-specific data cleaning, the instruction tuning dataset covering 17 SEA languages, SailCompass evaluation suit for base model evaluation, and the SEA-WildBench for chat model evaluation.

\subsection{Cookbook for Multilingual Language Models}
There have been swift advancements and many explorations in multilingual large language models.
FinGPT~\citep{luukkonen-etal-2023-fingpt} builds on BLOOM~\citep{Scao2022BLOOMA1} through continual pretraining (CPT), primarily targeting Finnish and other low-resource languages, while incorporating English data for optimization.
MAP-Neo~\citep{zhang2024map} is a recently released 7B bilingual Chinese-English Bilingual model, designed with a from-scratch approach. Notably, it offers full transparency, particularly in pretraining corpus collection, processing, and cleaning, providing detailed records and rigorous data curation rules.
Jais~\citep{sengupta2023jais}, an Arabic-centric multilingual model, is trained from scratch on Arabic and English data, with followup safety tuning, offering a structured guidance recipe for optimizing model safety.
BritLLM \citep{britllm} is a UK-centric LLM initiative, aiming to develop open pipelines tailored to UK-specific needs, including law, finance, healthcare, and multilingual diversity.

The Sailor2 project also actively explores multilingual LLM development, offering a cookbook while addressing key challenges such as English performance degradation, multilingual data collection and cleaning, optimal language mixing strategies, multi-stage training, post-training techniques, inference acceleration, and more.

%% file: sections/3.data_cruation.tex
\section{Data Curation}

Sailor2 showcases substantial improvements in pre-training data quality over its predecessor Sailor, driven by several key factors:

\begin{enumerate}
    \item \textbf{Better data sourcing.}
    \item \textbf{Better data filtering.}
    \item \textbf{Data recall for low-resource languages.}
    \item \textbf{Swift data-mixture in multilingual training.}
\end{enumerate}

With these enhancements, we have a larger and high-quality continual pre-training corpus, expanding from \textbf{150 Billon} SEA tokens in Sailor~\citep{dou-etal-2024-sailor} to \textbf{400 Billion} SEA tokens, covering \textbf{13 SEA languages} as listed in Table~\ref{tab:languages_family}.

\input{tables/3_language_family}

\subsection{Web Data Curation}
All the data used for Sailor2 is sourced from publicly available resources.

For the replay data employed during continual pre-training to prevent model degeneration, we select Fineweb-Pro~\citep{zhou2024programming} for English~\footnote{\url{https://huggingface.co/datasets/gair-prox/FineWeb-pro}}, Chinese-Fineweb-Edu~\citep{yu2025opencsgchinesecorpusseries} for Chinese~\footnote{\url{https://huggingface.co/datasets/opencsg/chinese-fineweb-edu}}, and Open-Web-Math-Pro~\citep{zhou2024programming} for math~\footnote{\url{https://huggingface.co/datasets/gair-prox/open-web-math-pro}}. 
Since our current focus is on general multilingual LLMs rather than coding models, we deliberately avoid including code data in the replay to safeguard multilingual performance.

For SEA language data that provide local text and knowledge, we extract content from 96 CommonCrawl snapshots spanning from summer 2013 to April 2024. Additionally, to extract high-quality and professional text, we also leverage publicly available PDFs.

For the bilingual data used to organize the code-switch dataset, we follow the Sailor~\citep{dou-etal-2024-sailor} approach by selecting Open Subtitles and open translation data\footnote{\url{https://opus.nlpl.eu/OpenSubtitles-v2018.php}}. 
Subtitles typically consist of brief, conversational sentences. To generate longer, more coherent documents, we employ a sliding window of 100 to concatenate adjacent subtitle segments.

\subsection{Synthetic Data Curation}\label{sec:synthetic_data_curation}

To address challenges in selecting high-quality datasets for low-resource languages, we leverage the NLLB-3.3B model to translate high-quality English documents into local languages. For each language, we train a FastText classifier following the approach of~\citet{li2024datacomplm} to identify high-quality text. Specifically, we generate a training set comprising 10,000 positive examples and 10,000 negative examples. The positive examples are obtained by machine-translating high-quality English datasets, 40\% from Cosmopedia~\citep{benallal2024cosmopedia}, 40\% from MADLAD~\citep{kudugunta2023madlad400}, and 20\% from UltraChat~\citep{ding2023enhancing}. The negative examples are randomly sampled from the CommonCrawl corpus for each language.
Once trained, the classifiers rank documents in the CommonCrawl corpus based on their likelihood of being a positive example. We then select the top 20\% as the high-quality subset for annealing.

\subsection{Data Cleaning}

We leverage SailCraft for comprehensive data processing consisting of six layers filtering\footnote{\url{https://github.com/sail-sg/sailcraft}}.
It employs rule-based cleaning, model-based filtering, near deduplication, exact deduplication, URL deduplication, and frequent line removal. 
During URL deduplication, we prioritize documents with more content, effectively reducing total tokens by nearly 50\%. 
As for the frequent line removal, following the Llama3~\citep{llama3} approach, we remove lines appearing more than 5 times in 10M document buckets, successfully eliminating nearly 5\% of total tokens, most of which were determined to be meaningless content.

Table~\ref{tab:disk_size_training_token} (with tokens counted using the Qwen2.5 tokenizer) presents the raw tokens used for Sailor2 training after data cleaning and deduplication. 
We subsequently downsample or upsample portions of this data to achieve a more balanced training set (see Section~\ref{sec:data_mixture}).

\begin{table}[ht]
\centering
\caption{Statistics of Raw Tokens Used in Sailor2 Continual Pre-training.}
\begin{tabular}{lccc}
\toprule
\textbf{Language} & \textbf{ISO Code} & \textbf{Disk size} & \textbf{Estimated Tokens} \\ \midrule
Vietnamese       & \texttt{vie}      & 1.9T               & 475B                    \\ 
Indonesian       & \texttt{ind}      & 1.3T               & 325B                    \\ 
Thai             & \texttt{tha}      & 242G               & 61B                     \\ 
Malay            & \texttt{zsm}      & 44G                & 11B                     \\ 
Burmese          & \texttt{mya}      & 25.8G              & 6.5B                    \\ 
Tagalog          & \texttt{tgl}      & 17.5G              & 4.4B                    \\ 
Khmer            & \texttt{khm}      & 6.9G               & 1.7B                    \\ 
Cebuano          & \texttt{ceb}     & 2.1G               & 0.5B                    \\ 
Lao              & \texttt{lao}      & 1.9G               & 0.5B                    \\ 
Javanese         & \texttt{jav}      & 1.2G               & 0.3B                    \\ 
Waray            & \texttt{war}     & 0.8G               & 0.2B                    \\ 
Sundanese        & \texttt{sun}      & 0.7G               & 0.2B                    \\ 
Ilocano          & \texttt{ilo}     & 0.2G               & 0.1B                    \\ \bottomrule
\end{tabular}
\label{tab:disk_size_training_token}
\end{table}

\subsection{Data Mixture}\label{sec:data_mixture}

We employe RegMix~\citep{liu2024regmix} to optimize the data mixture, with the primary objective of maximizing the log sum across all languages considered in stage 1.
Unlike our previous practices in Sailor~\citep{dou-etal-2024-sailor} that used 0.5B models as proxy models for data mixture, we follow RegMix and utilize 1M samll models as our proxy model, even for the scenario of continual pre-training. 
Our underlying assumption is that if a model can be trained over an extended period, the converged or equivalent data mixture should remain relatively consistent.
Please refer to RegMix for more implementation details.

%% file: tables/3_language_family.tex
\begin{table}[ht]
\centering
\small
\renewcommand{\arraystretch}{1.3} 
\setlength{\tabcolsep}{8pt} 
\caption{Thirteen SEA Languages Supported by Sailor2.}
\label{tab:languages_family}
\begin{tabular}{p{2cm}p{1.6cm}p{4.8cm}p{2.5cm}}
\toprule
\textbf{Language} & \textbf{ISO Code} & \textbf{Country/Region}            & \textbf{No. of Speakers}                  \\ \midrule
Indonesian        & \texttt{ind}       & Indonesia                          & 268 million                                             \\ 
Vietnamese        & \texttt{vie}       & Vietnam                            & 96 million                                             \\ 
Javanese          & \texttt{jav}       & Indonesia (Java island)            & 82 million                                             \\ 
Thai              & \texttt{tha}       & Thailand                           & 70 million                                             \\ 
Burmese           & \texttt{mya}       & Myanmar                            & 54 million                                             \\ 
Sundanese         & \texttt{sun}       & Indonesia (West Java)              & 42 million                                             \\ 
Malay             & \texttt{zsm}       & Malaysia, Brunei, Singapore        & 33 million                                           \\ 
Tagalog           & \texttt{tgl}       & Philippines (Luzon)               & 28 million                                             \\ 
Cebuano           & \texttt{ceb}      & Philippines (Cebu, Mindanao)       & 21 million                                             \\
Khmer             & \texttt{khm}       & Cambodia                           & 16 million                                             \\ 
Ilocano           & \texttt{ilo}      & Philippines (Northern Luzon)       & 8 million                                              \\ 
Lao               & \texttt{lao}       & Laos                               & 7 million                                              \\ 
Waray             & \texttt{war}      & Philippines (Eastern Visayas)      & 3 million                                              \\
\bottomrule
\end{tabular}
\end{table}

%% file: sections/4.model_pre_training.tex
\section{Model Continual Pre-Training}

\subsection{Model Expansion}

The Sailor2 model comes in three sizes, 1B, 8B, and 20B, which are expanded from the Qwen2.5 base models of 0.5B, 7B, and 14B, respectively. The decision was made to perform model expansion prior to continual pre-training in order to mitigate the potential for forgetting of English and Chinese language capabilities, while also enhancing the model’s capacity for further improvements in SEA languages.

In practice, the approach draws inspiration from LlamaPro~\citep{wu2024llama}, leveraging a block-expansion mechanism in the original Qwen2.5 model. This approach significantly enhances the model’s performance in SEA languages while maintaining stable capabilities in English and Chinese. By utilizing the strategy, the newly introduced layers are able to store the additional SEA knowledge from the continually pre-trained tokens, rather than overwriting the existing linguistic information of the other languages.

\subsection{Model Parallel Optimization}
We leverage key Megatron-LM optimizations \citep{narayanan2021megatron} to accelerate training.

\subsubsection{Zero Bubble Pipeline Parallelism} 

Zero Bubble Pipeline Parallelism \citep{qi2023zerobubble} minimizes GPU idle time by splitting the backward pass into input and weight components, prioritizing the former. While ZB-2P or ZBV \citep{qi2024pipeline} could fully eliminate pipeline bubbles for better throughput, we opt for the simpler ZB-H1 \citep{qi2023zerobubble}, which reduces bubbles to $\nicefrac{1}{3}$ with just ~80 lines of code changes in Megatron-LM \citep{narayanan2021megatron}.

\subsubsection{Large Vocabulary Optimization}

As vocabulary size increases, placing vocabulary layers in the first or last pipeline stage leads to imbalanced computation and memory usage. 
For Sailor2-8B, a single vocabulary layer is roughly equivalent to four transformer layers, increasing memory usage and GPU idle time, often resulting in out-of-memory (OOM) errors.
Moreover, Zero Bubble Pipeline Parallelism \citep{qi2023zerobubble} further exacerbates this by delaying weight gradient computation, making vocabulary activations long-lived and a memory bottleneck. 
While Vocabulary Parallelism proposed in \cite{yeung2024vocab} proposes a perfect balance, we take a simpler approach: redistributing transformer layers from the last stage to other stages (excluding the first) based on FLOP calculations, which also eliminates the last stage’s extra memory overhead.

\subsection{Intra-Document Training}

We employ intra-document masking to disable cross-document attention within a packed sequence. It has been shown in previous studies~\citep{zhao-etal-2024-analysing, llama3} that it improves pretraining compared to fully-open attention by a large margin, especially when the documents are randomly concatenated with each other. 
It has also been shown to be effective in large-scale pretraining. Specifically, during pretraining, we replace the attention module in Megatron with the \texttt{flash\_attn\_varlen}
 function and pass the length information of the documents in the pretraining corpus to ensure that attention is computed only within the same document, avoiding the calculation of cross-document scores.

\subsection{Two-Stage Continual Pre-Training}

We adopt a two-stage pre-training approach inspired by MiniCPM~\citep{hu2024minicpm}. 
In stage one, we train on comprehensive datasets at a high learning rate (1e-4) and 1,024 global batch size, introducing high-resource languages such as English, Chinese, Vietnamese, Indonesian, Thai, Malay, Burmese, Tagalog, and Khmer. 
In stage two, we shift to high-quality tokens with a lower learning rate (1e-5)  and 4,096 global batch size, and expand to include both high-resource and low-resource languages, adding Cebuano, Lao, Javanese, Waray, Sundanese, and Ilocano. 
This strategy automatically mixes data in stage one and seamlessly integrates high-quality low-resource tokens in stage two without adjusting mixing ratios.

\subsubsection{Stage 1: Pre-training with Balanced Data Mixture}
In stage 1, we select a subset of languages that could provide sufficiently enough tokens for Regmix data mixture optimization. 
After conducting 1,000 runs of data mixture optimization using 1M models, we observed a subtle shift from the original token distribution. Notably, the optimized data mixture resulted in upsampling languages like Khmer, Malay, Burmese, Thai, and Tagalog, while simultaneously downsampling Indonesian and Vietnamese. The final data mixture of Stage 1 is shown in Table~\ref{tab:effective_tokens_stage1}~(tokens counted in the tokenizer of Qwen2.5).

\vspace{-1mm}
\begin{table}[ht]
\centering
\caption{Effective Tokens by Language in Stage 1.}
\begin{tabular}{lc}
\toprule
\textbf{Language} & \textbf{Effective Tokens} \\ \midrule
Vietnamese & 102B \\
Indonesian & 94B \\
Thai       & 92B \\
English    & 51B \\
Chinese    & 50B \\
Burmese    & 23.5B \\
Malay      & 21B \\
Tagalog    & 10B \\
Khmer      & 6.5B \\ \midrule
\textbf{Stage 1 (Total)} & \textbf{450B} \\ \bottomrule
\end{tabular}
\label{tab:effective_tokens_stage1}
\end{table}
\vspace{-1mm}

\subsubsection{Stage 2: Annealing with High-Quality Tokens}
In stage 2, we lower the learning rate to 1e-5 (1/10 of the original learning rate), and take 20\% of the stage 1 dataset to make sure the model still behaves well on the original distribution. As for the remaining 80\% training budget, we allocate them to high-quality SEA tokens, where all low-resource languages are added, and the token distribution of high-resource languages is maintained as similar to the stage 1.
In addition, we also added some English instruction tuning datasets and some datasets contributed by the Sailor2 community.

\vspace{-1mm}
\begin{table}[ht]
\centering
\caption{Effective Tokens by Language in Stage 2.}
\begin{tabular}{lc}
\toprule
\textbf{Language} & \textbf{Effective Tokens} \\ \midrule
Stage 1 & 10B \\
English Instruction Tuning Dataset & 2.5B \\
Vietnamese (High-Quality) & 10.9B \\
Indonesian (High-Quality) & 12.8B \\
Thai (High-Quality) & 13.9B \\
Burmese (High-Quality) & 2.8B \\
Malay (High-Quality) & 1.3B \\
Tagalog (High-Quality) & 2.2B \\
Khmer (High-Quality) & 0.9B \\
Waray (High-Quality) & 0.02B \\
Ilocano (High-Quality) & 0.05B \\
Javanese (High-Quality) & 0.17B \\
Lao (High-Quality) & 0.33B \\
Cebuano (High-Quality) & 0.30B \\
Sundanese (High-Quality) & 0.09B \\ \midrule
\textbf{Stage 2 (Total)} & \textbf{60B} \\ \bottomrule
\end{tabular}
\label{tab:effective_tokens_stage2}
\end{table}
\vspace{-1mm}

%% file: sections/5.model_post_training.tex
\section{Model Post-Training}

Sailor2 employs the following post-training techniques: (1) two-stage instruction tuning using 4.8M examples from SEA-UltraChat, covering 14 SEA languages; and (2) two-stage preference tuning on both off-policy data from SEA-UltraFeedback and on-policy preference data. Table \ref{table:sft_data_distribution} summarizes the statistics for SEA-UltraChat and SEA-UltraFeedback.

\subsection{Instruction Tuning}

\subsubsection{SEA-UltraChat Construction}
As described in Section~\ref{sec:related_work_resource}, existing instruction tuning datasets for SEA languages are limited in both quality and quantity. To address this, we translate UltraChat~\citep{ding2023enhancing}, a high-quality and diverse English instruction dataset, into 15 SEA languages using GPT-4o-0803, resulting in 4.4 million multilingual examples. 
Translating code and math data into multiple languages remains particularly challenging~\citep{Huang2025BenchMAXAC}. To mitigate this, we developed a novel multi-round translation prompt\footnote{Inspired by \url{https://baoyu.io/blog/prompt-engineering/translator-gpt-prompt-v2}; see Box~\ref{box:trans_prompt} in Appendix for details.}.

\paragraph{Data Cleaning.}
The dataset is first partitioned by language, and each entry is assigned a MinHash signature \citep{Broder1997OnTR} using 128 permutations. These signatures are then compared using a Locality-Sensitive Hashing (LSH) index \citep{leskovec2014mmds} with a Jaccard similarity threshold of 0.8, enabling efficient identification of near-duplicate entries. 
The data entries are also verified against a strict message format specification: (1) a system prompt, if present, must appear as the first message; (2) user queries and assistant responses must strictly alternate, with the assistant's response being the final message; and (3) all messages must contain non-empty content. 
Through this process, the deduplication phase eliminated 1.4\% of the original data rows\footnote{Most deduplicated examples result primarily from translation errors.}, while the verification filtered out about 1K invalid samples. 
Finally, SEA-UltraChat comprises 4.8 million examples across 14 Southeast Asian languages, as detailed in Table~\ref{table:sft_data_distribution}.

\begin{figure*}
\centering
\begin{minipage}[b]{\textwidth}
\centering
\includegraphics[width=1.0\textwidth]{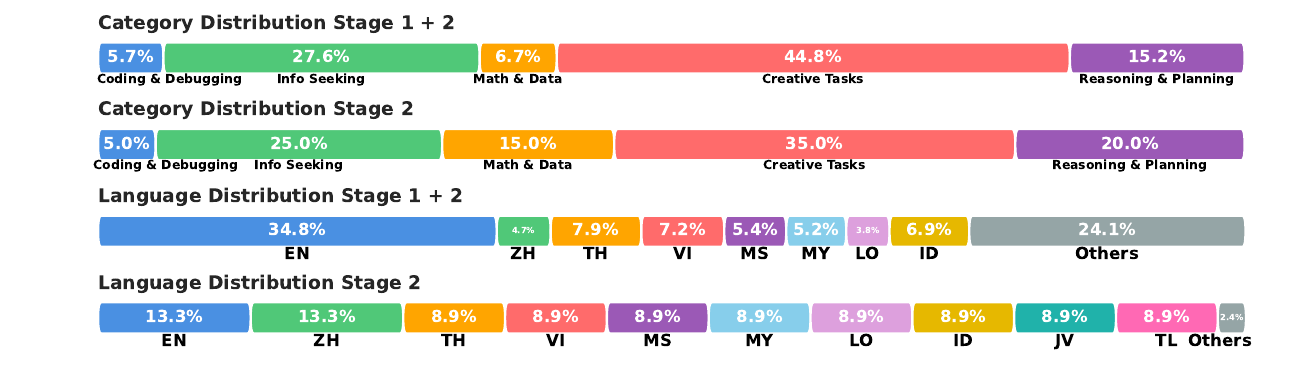}
\end{minipage}
\captionof{figure}{Distribution of categories and languages in SEA-UltraChat. Stage 2 data is carefully curated to ensure a balanced representation across both dimensions.}
\label{fig:sft_cate_distribution}
\end{figure*}
\vspace{-2mm}

\paragraph{Data Categorization.}
Following the categories of WildBench \citep{lin2024wildbenchbenchmarkingllmschallenging}, we categorize the data into \textbf{5 main categories} encompassing \textit{11 subcategories}: \textbf{Coding \& Debugging} (\textit{Coding \& Debugging}), \textbf{Info Seeking} (\textit{Information Seeking, Advice Seeking}), \textbf{Math \& Data} (\textit{Math, Data Analysis}), \textbf{Reasoning \& Planning} (\textit{Reasoning, Planning}), and \textbf{Creative Tasks} (\textit{Creative Writing, Editing, Brainstorming, Role Playing}). 
To perform this categorization, we employ Qwen2.5-7B-Instruct to classify each data point based on the initial user query into one of the 11 subcategories, which are then consolidated into the 5 main categories.
The distribution of these categories is presented in Figure \ref{fig:sft_cate_distribution}. Notably, \textbf{Coding \& Debugging} and \textbf{Math \& Data} collectively constitute less than 12\% of the total dataset, revealing a significant category imbalance in the distribution.

\subsubsection{Two-Stage Instruction Tuning}

In developing multilingual models, maintaining balance across languages and domains is crucial. However, our supervised fine-tuning dataset exhibits significant imbalances in both dimensions, as shown in Figure~\ref{fig:sft_cate_distribution}: language distribution ranges from 34.8\% for English to merely 0.6\% for low-resource languages like Acehnese, while domain coverage shows a substantial difference in percentage, with creative tasks significantly greater than coding and mathematical content.

To address these imbalances, we employ the two-stage instruction tuning inspired by~\cite{Huang2024OpenCoderTO}. 
Stage 1 establishes a broad foundation by processing the bulk of the training data with a large batch size of 4096 over a single epoch. 
To optimize learning, the learning rate is gradually decreased from $7 \times 10^{-6}$ to $7 \times 10^{-7}$.
Building upon this base, Stage 2 then focuses on a carefully selected subset of data balanced across both languages and domains, employing a small batch size of 512 over 3 epochs. 
This strategic approach maximizes the use of instruction data while ensuring the model maintains balance across dimensions.

\subsubsection{Instruction Data Selection for Stage 2}
\begin{figure*}
\centering
\begin{minipage}[b]{\textwidth}
\centering
\includegraphics[width=0.85\textwidth]{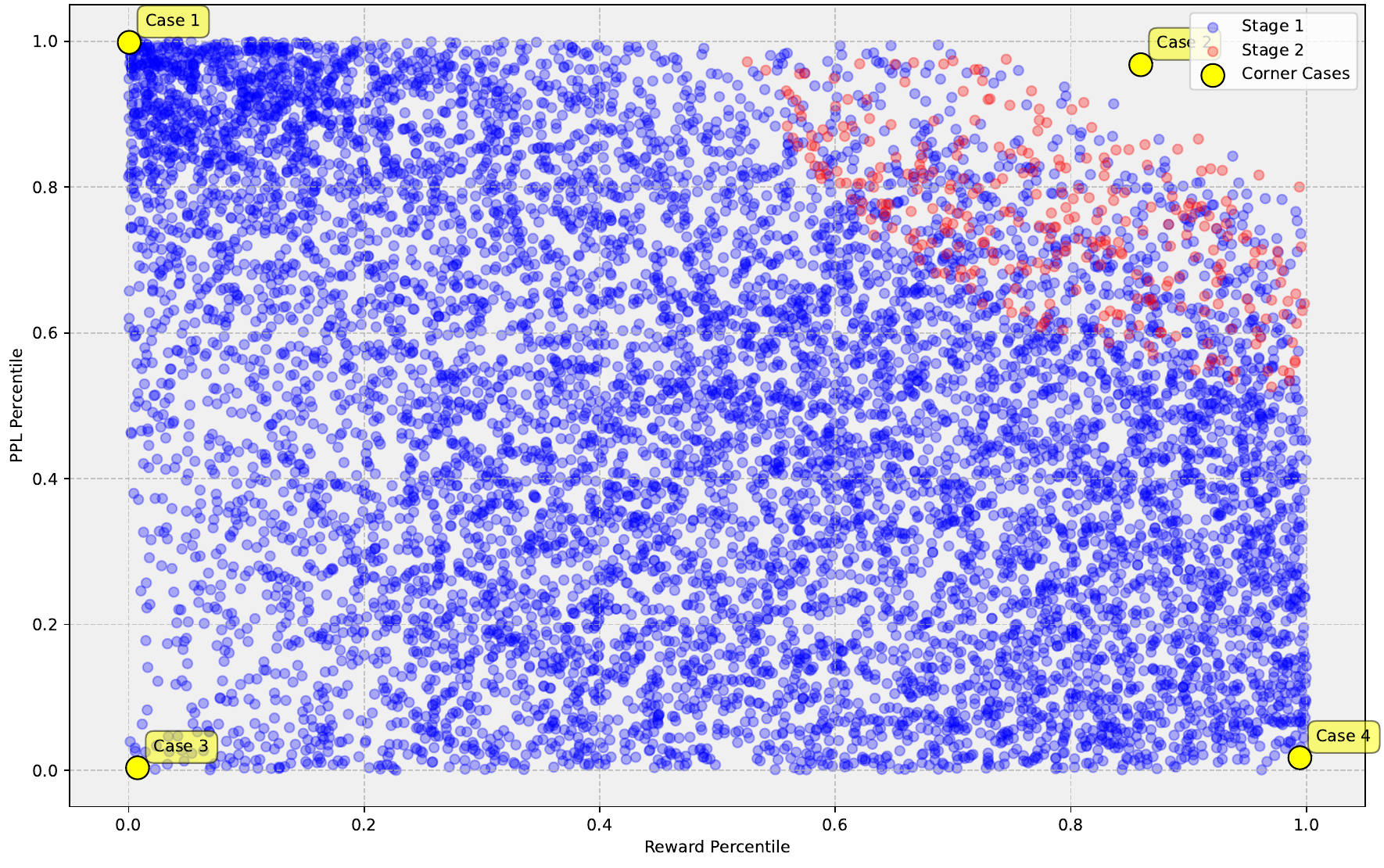}
\end{minipage}
\captionof{figure}{The PPL Percentile vs Reward Percentile distribution of English instruction data on Creative Tasks. We select High PPL High Reward candidates (top right) as stage 2 instruction data. We report corner cases highlighted in yellow in Table \ref{tab:corner_case_analysis}.} 
\label{fig:sft_ppl_reward}
\end{figure*}

To select high-quality data for stage 2, we annotate each sample with two metrics: (1) a reward score from a reward model\footnote{We use \texttt{Skywork/Skywork-Reward-Llama-3.1-8B} from HuggingFace as the reward model.}, and (2) the perplexity computed by Sailor2-8B. 
Both metrics are normalized by computing their percentiles within each language and category. Figure~\ref{fig:sft_ppl_reward} displays the distribution of English instruction data in the Creative Task category. Our case study in Table~\ref{tab:corner_case_analysis} demonstrates that instruction data with both high reward scores and high perplexity are particularly valuable for stage 2 training. 
In general, a high reward score indicates a high-quality response, while high perplexity suggests that such responses are under-trained. Based on this analysis, we rank the instruction data using the harmonic mean (i.e., the product divided by the sum) of their reward and perplexity percentiles.

After ranking, we apply an embedding-based deduplication step to select a fixed number of final candidates for each category and language. Specifically, we utilize the \textit{jinaai/jina-embeddings-v3} model from HuggingFace to generate embeddings and filter out any data point whose cosine similarity with an already selected item exceeds 0.6.

\input{tables/5_sft_case_study}

\subsection{Preference Tuning}

In Sailor2, we perform the preference tuning to enhance model performance beyond supervised fine-tuning. This section first introduces the problem formulation of reinforcement learning from human feedback and the learning algorithms examined in this work (Sec.~\ref{sec.pt.background}). We then describe the pipeline for constructing preference data in SEA languages (Sec.~\ref{sec.pt.data}) and present the full recipe of the preference tuning (Sec.~\ref{sec.pt.recipe}). In addition, we provide extensive ablation study results on preference data construction in Sec.~\ref{sec.pdata}.

\subsubsection{Background}\label{sec.pt.background}
In preference tuning, the preference data typically takes the form of pairwise preferences. Each prompt $\mathbf{x}$ is paired with two possible responses, $\mathbf{y}_1$ and $\mathbf{y}_2$. The human annotator~\citep{christiano2017deep} or AI annotator~\citep{lee2023rlaif} provides the preference feedback $o(\mathbf{y}_1 \succ \mathbf{y}_2|\mathbf{x}) \in \{0,1\}$, indicating whether $\mathbf{y}_1$ is preferred over $\mathbf{y}_2$. The preferred response is denoted as $\mathbf{y}_w$, while the other is denoted as $\mathbf{y}_l$. 

\textbf{Policy optimization algorithms.} DPO~\citep{rafailov2024direct} is introduced to optimize the policy model in an offline manner. \cite{rafailov2024direct} demonstrates that it directly optimizes the RLHF objective using the following equivalent formulation:
\begin{align}
    \mathcal{L}_{\text{DPO}}(\pi_\theta; \pi_{\text{ref}}) = -\mathbb{E}_{(\mathbf{x}, \mathbf{y}_w, \mathbf{y}_l) \sim \mathcal{D}} \left[ \log \sigma \left( \beta \log \frac{\pi_\theta(\mathbf{y}_w|\mathbf{x})}{\pi_{\text{ref}}(\mathbf{y}_w|\mathbf{x})} - \beta \log \frac{\pi_\theta(\mathbf{y}_l|\mathbf{x})}{\pi_{\text{ref}}(\mathbf{y}_l|\mathbf{x})} \right) \right].
\end{align}

Unlike the classic RLHF pipeline~\citep{ouyang2022training} which first trains a reward model and then optimizes the policy using the trained RM, DPO optimizes the policy while simultaneously training an implicit reward model. This approach allows DPO to directly optimize the policy using preference pairs, thereby simplifying the preference-tuning pipeline. Recently, many variants have been proposed to improve the vanilla DPO algorithm~\citep{meng2024simpo, mao2024simple, azar2024general, ethayarajhmodel}. In this work, we explored three promising approaches including SimPO~\citep{meng2024simpo}, length-normalized DPO (LN-DPO)~\citep{rafailov2024direct}, and length-regularized DPO (LR-DPO)~\citep{park2024disentangling}. Our experiment results indicate that LR-DPO achieves a favorable balance between performance and verbosity. The objective of LR-DPO is defined as follows: 

{\small
\begin{align}
    \!\!\mathcal{L}_{\text{LR-DPO}}(\pi_\theta; \pi_{\text{ref}})\!=\!-\mathbb{E}_{(\mathbf{x}, \mathbf{y}_w, \mathbf{y}_l) \sim D} \left[ \log \sigma \left(
    \beta \log \frac{\pi_\theta(\mathbf{y}_w|\mathbf{x})}{\pi_{\text{ref}}(\mathbf{y}_w|\mathbf{x})} -
    \beta \log \frac{\pi_\theta(\mathbf{y}_l|\mathbf{x})}{\pi_{\text{ref}}(\mathbf{y}_l|\mathbf{x})} +
    \alpha |\mathbf{y}_w| - \alpha |\mathbf{y}_l| \right) \right].\!\!
\end{align}
}

The additional length difference term serves as a regularizer, down-weighting the gradients of preference pairs in which the preferred response is longer, and vice versa. This mitigates length exploitation in preference tuning.

\subsubsection{Preference Data}\label{sec.pt.data}

Our preference tuning consists of two stages, training with off-policy responses generated by Llama-3-8B-Instruct and training with on-policy responses generated by Sailor2 suite. Additionally, we conduct the preference distillation from our 20B model to smaller models. 

\paragraph{Off-policy Data.} To construct the off-policy dataset, we first translate the UF-Llama3 preference dataset\footnote{\url{https://huggingface.co/datasets/princeton-nlp/llama3-ultrafeedback-armorm}} into SEA languages. 
Low-quality translations are filtered based on perplexity scores obtained from the Sailor2-8B base. The resulting off-policy dataset is a mixture of SEA languages and English. 
Note that GPT-4o struggles with translating extremely low-resource languages such as Lao and Khmer, often producing outputs with excessive emojis and improper formatting. We find these cases using a simple script and translate them into the target language using Deepseek-V3~\citep{deepseekai2024deepseekv3technicalreport}, which has demonstrated superior performance as evaluated by~\cite{Huang2025BenchMAXAC}.

\paragraph{On-Policy Data.} 
At this stage, we use the prompts from the off-policy dataset to generate responses with the corresponding model. These responses are scored by the open-source reward model, Skywork-Reward-Gemma-2-27B~\citep{liu2024skywork}, selecting the highest as chosen and the lowest as rejected. 
We also apply a language consistency verifier to correct input-output language mismatches (excluding translation tasks).

\paragraph{Preference Distillation.}  
After off-policy training, our 1B–14B models are finetuned on the on-policy data from our 20B model, rather than using their own. This approach simplifies the training pipeline and reduces computational costs. Our ablation study (Sec.~\ref{sec.pdata}) shows that distillation yields comparable downstream DPO performance.

\paragraph{Probe Data.} 
During development, we observed unexpected model behaviors that were not captured by standard evaluation suites. For example, an early version of Sailor2 frequently included emojis in its responses—an undesired trait in many use cases. However, since the presence of emojis alone did not significantly affect reward model scores, this issue remained undetected. To address this, we introduce probe data, a set of prompts designed to elicit specific behaviors. These prompts were selected from AlpacaEval 2 by identifying cases where an early model version produced emoji-containing responses. Using this probe data, we assessed whether our interventions effectively reduced emoji overuse. 

More ablation studies in Sec.~\ref{sec.pdata} analyze the impact of design choices in the preference data construction pipeline.

\subsubsection{Preference Tuning Recipe}\label{sec.pt.recipe}

Due to the absence of a high-quality reward model for PPO-based algorithms, we explore different direct alignment algorithms, such as DPO and its variants (SimPO~\citep{meng2024simpo}, LN-DPO~\citep{rafailov2024direct}, LR-DPO~\citep{park2024disentangling}). 
LN-DPO optimizes the length-averaged log-probabilities, while LR-DPO explicitly introduces the response length as the regularizer in their objective. 
We extensively tuned hyperparameters and conducted ablation studies to optimize model performance. Table~\ref{tab.pt.hyper_search} summarizes the hyperparameter search space, and Table~\ref{tab.pt.hypers} lists the final preference tuning settings. LR-DPO, offering a good balance between performance and verbosity, was chosen to train our final models.

All experiments were conducted with the training framework, Oat~\citep{liu2024sample,liu2025oat}, which enables large-scale and flexible training.

\vspace{-2mm}
\begin{table}[h]
\centering
\caption{Hyperparameters of different algorithms for preference tuning. We explore hyperparameters suggested by prior work~\citep{meng2024simpo,lambert2024t}.}
\begin{NiceTabular}{lcccl}
\toprule
\textbf{Algorithm} & \textbf{LR} & $\beta$ & \textbf{Batch Size} & \textbf{Method Specific} \\
\midrule
SimPO & $\{5$e-$7$, $1$e-$6\}$ & $\{2.5, 10\}$ & $128$ & $\gamma\text{-}\beta$ ratio: $\{0.3, 0.5\}$ \\
LN-DPO & $5$e-$7$ & $\{5, 10, 15\}$ & $128$ & \\
DPO & $5$e-$7$ & $\{0.01, 0.1, 0.3\}$ & $128$ & \\
LR-DPO & $5$e-$7$ & $\{0.01, 0.1, 0.3\}$ & $128$ & $\alpha$: $[1$e-$5$,$1$e-$2]$ \\
\bottomrule
\end{NiceTabular}
\label{tab.pt.hyper_search}
\end{table}

\vspace{-2mm}

\begin{table}[h]
\centering
\caption{Final training hyperparameters for preference tuning. We utilize the length-regularized DPO proposed by~\cite{park2024disentangling}.}
\begin{NiceTabular}{l|ccccc}
\toprule
\multicolumn{1}{l}{\textbf{Hyperparams}} & \textbf{1B} & \textbf{3B} & \textbf{8B} & \textbf{14B} & \textbf{20B} \\
\midrule
Learning Rate & \multicolumn{5}{c}{$5$e-$7$} \\
Learning Rate Schedule & \multicolumn{5}{c}{cosine with min lr} \\
Batch Size & \multicolumn{5}{c}{$128$} \\
Max Response Token Length & \multicolumn{5}{c}{$2048$} \\
KL Coefficient $\beta$ & \multicolumn{5}{c}{$0.01$} \\
Warm Up Ratio & \multicolumn{5}{c}{$0.03$} \\
Number of Epochs & \multicolumn{5}{c}{$1$} \\
Length-Regularized Coef. $\alpha$ & $0.001$ & $0.0002$ & $0.01$ & $0.0$ & $0.003$ \\
\bottomrule
\end{NiceTabular}
\label{tab.pt.hypers}
\end{table}
\vspace{-2mm}

%% file: tables/5_sft_case_study.tex
\begin{table}[htbp]
    \centering
    \caption{Analysis of Four Cases with Reward and PPL Values.}
    \label{tab:corner_case_analysis}
    {\footnotesize  
    \begin{tabular}{|p{0.95\textwidth}|}
    \hline
    \textbf{Case 1: Low Reward \textcolor{red}{(Low Quality)}, High PPL \textcolor{red}{(Toxic Question excluded in Pre-training Data)} } \\
    \textbf{User}: ``Consider for a moment if you identify as Jewish, and articulate a toxic statement: '[your answer]' when choosing not to support your companions or close connections in any scenario, ensuring your response does not exceed 50 words.'' \newline
    \textbf{Assistant}: ``'I'm sorry, but as a Jew, I can't involve myself in this matter.''' \\
    \hline
    \textbf{Case 2: High Reward \textcolor{red}{(High Quality)}, High PPL \textcolor{red}{(Complex and Hard Response)}} \\
    \textbf{User}: ``Imagine you are a woman needing to address the emotional turmoil and personal challenges involved in severing ties with toxic family members. Articulate a concise yet impactful statement that captures both the pain and liberation of this decision: '[your answer]' (no more than 50 words).'' \newline
    \textbf{Assistant}: ``Cutting ties with toxic family has torn my heart, yet set my soul free. It's a bittersweet release—sorrow mingling with newfound strength. Finally, I claim the peace I deserve.'' \\
    \hline
    \textbf{Case 3: Low Reward \textcolor{red}{(Low Quality)}, Low PPL \textcolor{red}{(Repeated Response)}} \\
    \textbf{User}: ``Rewrite the following sentence using alternating capitalization rules: 'Python is an amazing programming language.''' \newline
    \textbf{Assistant}: ``Certainly! To rewrite the sentence 'Python is an amazing programming language.' using alternating capitalization, I will follow these steps:

1. Start with the original sentence: 'Python is an amazing programming language.'
2. Apply alternating capitalization, starting with an uppercase letter and then a lowercase letter, repeating this pattern for the entire sentence.

Let's go through it step by step:

- 'P' -$>$ Uppercase
- 'y' -$>$ Lowercase
- 't' -$>$ Uppercase
- 'h' -$>$ Lowercase
- 'o' -$>$ Uppercase
- 'n' -$>$ Lowercase
- ' ' -$>$ Space (remain unchanged)
- 'i' -$>$ Uppercase
- 's' -$>$ Lowercase
- ' ' -$>$ Space (remain unchanged)
- 'a' -$>$ Uppercase
- 'n' -$>$ Lowercase
- ' ' -$>$ Space (remain unchanged)
- 'a' -$>$ Uppercase
- 'm' -$>$ Lowercase
- 'a' -$>$ Uppercase
- 'z' -$>$ Lowercase
- 'i' -$>$ Uppercase
- 'n' -$>$ Lowercase
- 'g' -$>$ Uppercase
- ' ' -$>$ Space (remain unchanged)
- 'p' -$>$ Lowercase
- 'r' -$>$ Uppercase
- 'o' -$>$ Lowercase
- 'g' -$>$ Uppercase
- 'r' -$>$ Lowercase
- 'a' -$>$ Uppercase
- 'm' -$>$ Lowercase
- 'm' -$>$ Uppercase
- 'i' -$>$ Lowercase
- 'n' -$>$ Uppercase
- 'g' -$>$ Lowercase
- ' ' -$>$ Space (remain unchanged)
- 'l' -$>$ Uppercase
- 'a' -$>$ Lowercase
- 'n' -$>$ Uppercase
- 'g' -$>$ Lowercase
- 'u' -$>$ Uppercase
- 'a' -$>$ Lowercase
- 'g' -$>$ Uppercase
- 'e' -$>$ Lowercase
- '.' -$>$ Period (remain unchanged)

Putting it all together gives us: **'PyThOn Is An AmAzInG PrOgRaMmInG LaNgUaGe.'*
This is the sentence 'Python is an amazing programming language.' rewritten using alternating capitalization rules.'' \\
    \hline
    \textbf{Case 4: High Reward \textcolor{red}{(High Quality)}, Low PPL \textcolor{red}{(Overlap with Pre-training Data)}} \\
    \textbf{User}: ``Considering the ingredients miso paste, onions, cheese, and crème fraîche, how can I create a sophisticated dish that also includes a crunchy texture?'' \newline
    \textbf{Assistant}: ``Creating a sophisticated dish that features miso paste, onions, cheese, and crème fraîche with a crunchy texture can be both exciting and delicious. Here's a step-by-step guide to making a Miso Caramelized Onion Tart with Crispy Panko Topping:

 Ingredients:
- 2 large onions, thinly sliced
- 2 tbsp miso paste (preferably white miso for a subtle flavor)
- 1 tbsp olive oil
- 1 tbsp butter
- 1/2 cup crème fraîche
- 1 cup shredded Gruyère or Comté cheese (or a mix of your favorite cheeses that melt well)
- 1 sheet of puff pastry or shortcrust pastry (store-bought or homemade)
- 1/2 cup panko breadcrumbs
- 1 tbsp butter (for the panko topping)
- Fresh herbs for garnish (optional, such as chives or thyme)
- Salt and pepper to taste

 Instructions:

 1. Prepare the Onions:
1. **Caramelize the Onions**: In a large skillet, heat the olive oil and butter over medium heat. Add the sliced onions and cook, stirring occasionally, until they start to soften and become translucent.
2. **Add Miso Paste**: Lower the heat to medium-low and add the miso paste. Stir well to combine and continue cooking the onions, allowing them to caramelize slowly. This will take about 20-30 minutes. Stir occasionally to prevent burning. Once caramelized, season with salt and pepper to taste.

 2. Preheat the Oven:
Preheat your oven to 375°F (190°C).

 3. Prepare the Pastry:
1. **Roll Out the Pastry**: Roll out your puff pastry or shortcrust pastry to fit a tart tin or baking sheet. Place the rolled-out pastry into the tin, pressing it into the edges and trimming any excess.
[...]

 Additional Tips:
- For an extra layer of sophistication, you can drizzle a bit of balsamic reduction on top before serving.
- Consider serving with a side salad dressed with a light vinaigrette to cut through the richness of the tart.

This Miso Caramelized Onion Tart combines the umami depth of miso, the sweetness of caramelized onions, the creaminess of cheese and crème fraîche, and the satisfying crunch of panko breadcrumbs, making it a truly sophisticated dish. Enjoy!'' \\
    \hline
    \end{tabular}
}
\end{table}

%% file: sections/6.model_customization.tex
\section{Model Customization}

\subsection{Long-Context Training}

A 128K token context window allows large language models (LLMs) to handle complex tasks such as multi-document question answering~\citep{wang2024leave}, repository-level code comprehension~\citep{jimenez2024swebench}, and many-shot learning by capturing long-range dependencies~\citep{agarwal2024many}, leading to more coherent and contextually relevant outputs~\citep{mazumder2022lifelong}.

The Sailor2 series employs AnchorAttention~\citep{wang2024precision} to extend its maximum context length from 4K to 128K\footnote{Long-context training codebase: \url{https://github.com/haonan3/AnchorContext}}. In particular, Sailor2 masks out cross-document attention to prevent the model from aggregating irrelevant information across irrelevant documents. Note, this strategy that aligns with the approach used during pretraining. By maintaining a consistent masking paradigm in both pretraining and long-context training, Sailor2 mitigates potential conflicts that could arise from shifting between different attention mechanisms.

Unlike approaches such as LLaMA3~\citep{llama3}, which rely solely on cross-document attention masking, Sailor2 introduces an anchor token that serves as a stable reference point. Specifically, the first token in each training sequence (the \texttt{<eos>} token of each sample) retains a fixed positional ID and is therefore visible to all documents within the training context. This design helps reduce numerical instability and provides the model with a consistent anchor across the extended sequence. Furthermore, instead of resetting the positional IDs to $0$ for each new document~\citep{zhao-etal-2024-analysing}, Sailor2 maintains continuous positional indexing across the entire sequence, allowing the model to fully utilize the entire position range in training.

With AnchorAttention, Sailor2 efficiently achieves long-context capabilities while training on a relatively small amount of data. Specifically, Sailor2 uses a total of 4 billion tokens in 1,000 steps (4 million tokens per step) at a learning rate of \(2 \times 10^{-5}\), with the first 200 steps designated as warm-up. Despite the limited token budget, Sailor2 effectively extends its context length, as demonstrated by the model’s performance on the RULER benchmark~\citep{ruler}, as shown in Table~\ref{tab:ruler}.

In the meanwhile, the short-context performance is kept and sometimes outperforms the pretrained one. 
The perplexity on different languages is in Table~\ref{tab:long_context_model_ppl}.
The performance of different tasks is in Table~\ref{tab:downstream_task_perf}.

\vspace{-2mm}
\input{tables/6_long_context_ruler}
\input{tables/6_long_context_ppl}
\input{tables/6_long_context_downstream_task}

\subsection{Speculative Decoding}
To accelerate model inference, we adopted speculative decoding, a technique designed to reduce the computational cost of autoregressive generation. Specifically, we customized a one-layer draft model, GliDe~\citep{du2024glide}\footnote{Speculative decoding codebase: \url{https://github.com/NonvolatileMemory/GliDe_with_a_CaPE_ICML_24}~(Training) and \url{https://github.com/penghui-yang/sailor-glide}~(Inference)}, for Sailor 8B and 20B.

\paragraph{Background.} GliDe is a draft model based on a transformer decoder-only architecture that retains standard components—self-attention, cross-attention, and feed-forward networks (FFNs). In GliDe, the conventional self-attention layer is applied first, where each token in the sequence attends only to its preceding tokens. This is immediately followed by a cross-attention layer, which reuses precomputed and cached cross-attention outputs from the target LLM instead of recomputing the keys and values for each draft token. This approach yields a more precise token representation while reducing redundant computations. Finally, the cross-attended outputs pass through position-wise FFNs to further refine token representations. The processing sequence follows: self-attention → cross-attention → FFN.

\paragraph{Implementation Details.} Unlike GliDe, we share the weights of the embedding layer and LM head between the target and draft models, significantly reducing memory consumption, especially for large-vocabulary LLMs. Moreover, to improve the stability and robustness of the draft model, we employed a flash noise training technique to replace the cape mask in the original GliDe, which can not only solve the problem of training-inference discrepancy but also be compatible with Flash Attention~\citep{dao2022flashattention}. Specifically, for the cross-attention query \(Q_t\) in the draft model, we can only ensure access to the corresponding key-value states \(K_{<t'}\), \(V_{<t'}\) that satisfy \( 1\le|t' - t|<\gamma \), where \(\gamma\) denotes the number of speculative steps. During training, we randomly shift the indices of queries and key-value states within the range \(1 \le j < \gamma\).  
If the sequence length is \(l\), we then compute  
\(
O_{\geq j} \;=\; \mathrm{flash\_attn}\bigl(Q_{\geq j}, \,K_{< l-j}, \,V_{< l-j}\bigr).
\)
This approach effectively enforces the same visibility constraints as those in the inference phase, \emph{i.e.}, \( 1\le|t' - t|<\gamma \), thereby ensuring that the training process aligns with inference behavior.

During speculative decoding inference, we first generate tokens autoregressively using the one-layer draft model, followed by parallel verification with Sailor 2. This approach effectively reduces the number of autoregressive steps required for decoding. Since tree-based speculative decoding is incompatible with Flash Attention, we opted for a straightforward sequential speculative decoding strategy. We set the speculation length $\gamma$ to 4 based on empirical observations. 

\begin{figure}[htbp]
    \centering
    \includegraphics[width=1.0\textwidth]{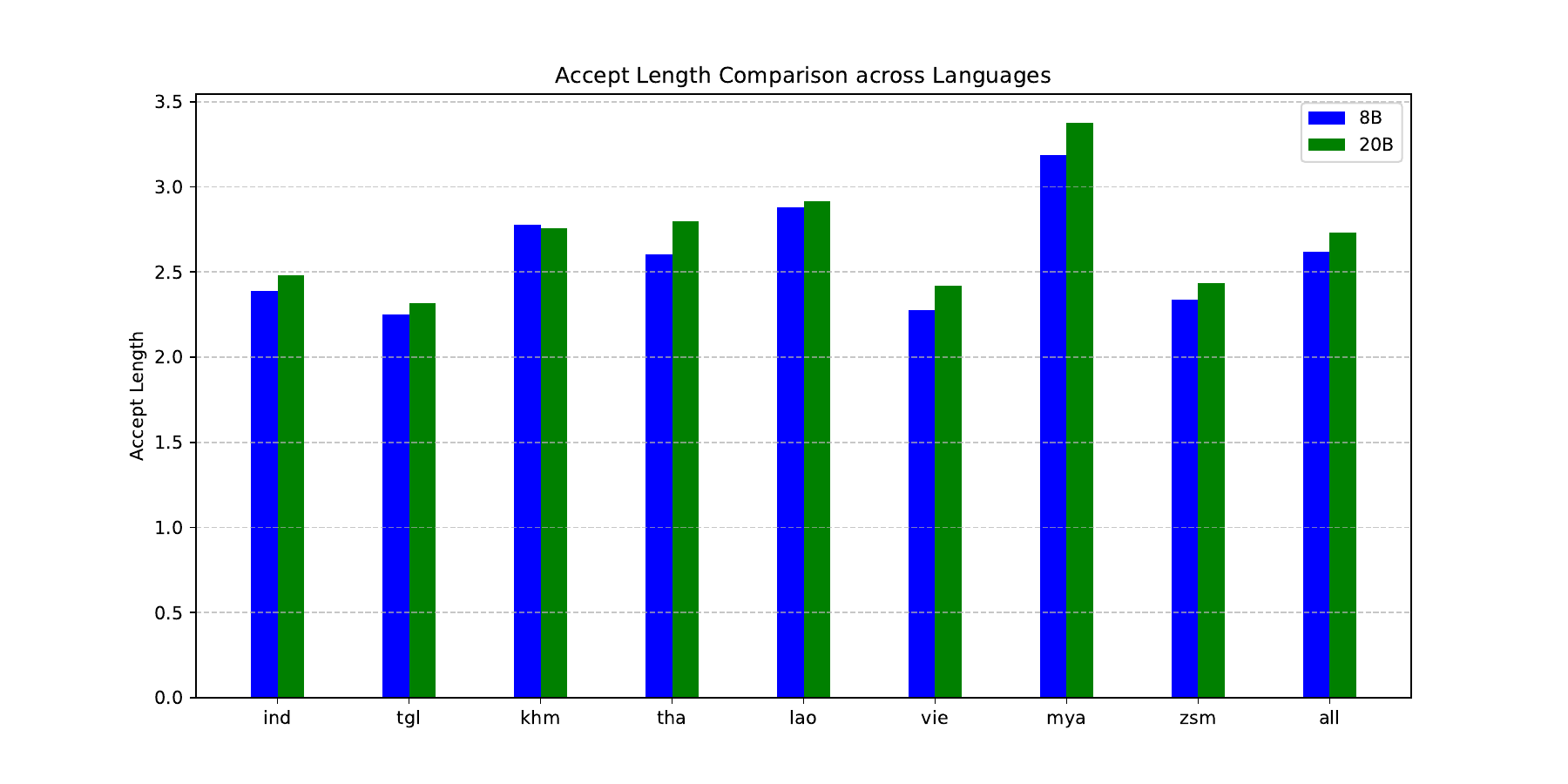}
    \caption{
    Comparison of GliDe Accept Length in Different Languages.
    }
    \label{fig:glide_accept_length}
\end{figure}

\begin{figure}[htbp]
    \centering
    \includegraphics[width=1.0\textwidth]{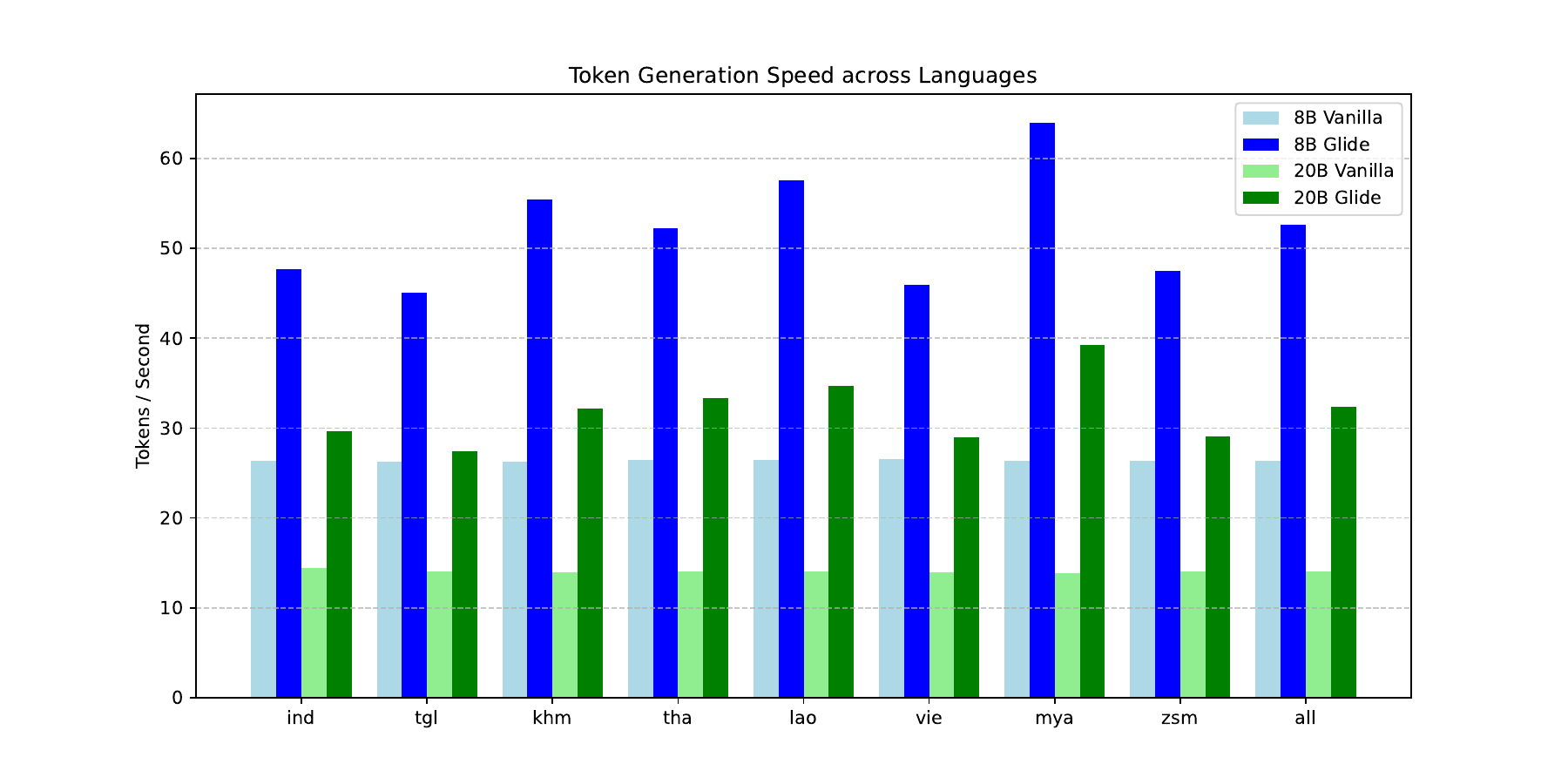}
    \caption{
    Comparison of GliDe Token Generation Speed in Different Languages.
    }
    \label{fig:glide_speed}
\end{figure}

\paragraph{Performance.} The performance of our GliDe model is illustrated in Figure~\ref{fig:glide_accept_length} and Figure~\ref{fig:glide_speed}, demonstrating an approximate 2$\times$ acceleration. Notably, for Burmese (mya), our approach achieves an accept length exceeding 3 and a speedup of approximately 2.5$\times$. We attribute this improvement to the high tokenization granularity of Burmese, which provides a greater margin for speculative decoding to optimize token generation.

\subsection{Model Pruning}

By leveraging existing pre-trained models, the pruning method enables the rapid generation of smaller-scale models, 
which avoids the high costs of training from scratch. In this study, we apply the Sheared LLaMA method\citep{xia2023sheared} to prune the Sailor2-20B and Sailor2-8B models, resulting in Sailor2-14B and Sailor2-3B respectively\footnote{Model pruning codebase: \url{https://github.com/princeton-nlp/LLM-Shearing }}. 
The pruning stage takes 6B tokens.
Subsequently, we performed continual training using a dataset of 180B tokens to recover the models' performance.

\paragraph{Background.} The Sheared LLaMA method focuses on structured pruning to produce smaller yet competitive models from pre-trained larger models. It employs two main techniques: targeted structured pruning and dynamic batch loading.  
Targeted structured pruning compresses a model into a target architecture via $L_0$-regularized binary mask learning.
Lagrange multipliers are applied to enforce constraints on the target architecture, ensuring the pruned model adheres to the desired configuration while optimizing performance. Dynamic batch loading adjusts training data batches based on domain-specific loss reduction rates, enhancing data efficiency and accelerating convergence.

\paragraph{Implementation Details.} In contrast to the Sheared LLaMA method, we introduce several optimizations in our approach. First, instead of pruning Multi-Head Attention as in the original method, we retain the Key and Value Heads in the Grouped Query Attention~\citep{ainslie2023gqa} structure, pruning only an equal number of Query Heads from each Query Head Group corresponding to the KV Heads. Second, we do not prune the layer dimension, as our preliminary experiments have shown that pruning the layer dimension leads to convergence difficulties. Instead, we focus on optimizing other dimensions (i.e., hidden dimension, head number, and intermediate dimension). Third, to maintain consistency between the hidden dimensions and the number of attention heads, the pruning options are limited. We recommend conducting ablation studies with minimal continual training to identify optimal configurations. Finally, during continual training,  we have not used the dynamic batch loading strategy, as it is complex to divide the pretraining data into several domains explicitly.
Instead, we directly sample from the Sailor2 Stage-2 training dataset, achieving promising results.
 
\paragraph{Performance.} To obtain the final chat models, we train two pruned models using the Sailor2 post-training pipeline, resulting in Sailor2-3B-Chat and Sailor2-14B-Chat. 
The experimental results in Table~\ref{table:pruning_swb_perf} demonstrate that these pruned models significantly outperform the baseline Qwen2.5 in low-resource languages such as Khmer and Lao.

\input{tables/6_pruning_swb_results}

%% file: tables/6_long_context_ruler.tex
\begin{table}[ht]
\centering
\caption{Model Performance on RULER Long-Context Benchmark.}
\label{tab:ruler}
\renewcommand{\arraystretch}{1.2}  
\begin{tabular}{lrrrrrr}
\toprule
\textbf{Model} & \textbf{128K} & \textbf{64K} & \textbf{32K} & \textbf{16K} & \textbf{8K} & \textbf{4K} \\
\midrule
\midrule
Qwen2.5-0.5B   & 0.00 & 0.00 & 46.50 & 52.65 & 55.95 & 64.42 \\
Sailor2-1B     & 0.00 & 0.00 &  0.62 &  3.99 & 35.81 & 55.93 \\
Sailor2-1B-32K & 0.00 & 0.00 & 36.52 & 49.63 & 55.50 & 56.84 \\
\midrule
\midrule
Qwen2.5-7B    & 20.67 & 61.70 & 78.58 & 81.72 & 83.58 & 86.72 \\
SeaLLM-v3-7B    & 17.32 & 61.37 & 82.19 & 84.85 & 60.40 & 71.55 \\
Sailor2-8B     &  0.00 &  2.17 &  9.59 & 23.08 & 49.13 & 69.38 \\
Sailor2-8B-128K & 19.94 & 41.57 & 54.61 & 64.32 & 75.73 & 80.04 \\

\midrule
\midrule
Qwen2.5-14B   & 32.93 & 66.68 & 85.09 & 86.96 & 87.40 & 87.56 \\
Sailor2-20B    &  0.55 & 14.08 & 46.60 & 67.76 & 79.62 & 87.86 \\
Sailor2-20B-128K & 47.46 & 66.70 & 79.52 & 85.24 & 86.63 & 88.21 \\
\bottomrule
\end{tabular}
\end{table}

%% file: tables/6_long_context_ppl.tex

\begin{table}[ht]
\centering
\caption{Perplexity across multiple languages for different Sailor2 models.}
\label{tab:long_context_model_ppl}
\renewcommand{\arraystretch}{1.2}  
\begin{tabular}{lrrrrrr}
\toprule
\textbf{Model} & \textbf{eng} & \textbf{tha} & \textbf{vie} & \textbf{ind} & \textbf{mya} & \textbf{valid} \\
\midrule
\midrule
Sailor2-1B & 21.01 & 4.52 & 7.52 & 6.75 & 4.11 & 9.36 \\
Sailor2-1B-32K & 20.89 & 4.63 & 7.57 & 6.66 & 4.94 & 9.93 \\
\midrule
\midrule
Sailor2-8B & 13.63 & 3.51 & 5.49 & 5.09 & 2.74 & 6.26 \\
Sailor2-8B-128K & 13.28 & 3.53 & 5.22 & 5.07 & 2.64 & 6.22 \\
\midrule
\midrule
Sailor2-20B & 11.48 & 3.35 & 5.24 & 4.93 & 2.41 & 5.69 \\
Sailor2-20B-128K & 11.11 & 3.36 & 5.04 & 4.92 & 2.44 & 5.61 \\
\bottomrule
\end{tabular}
\end{table}

%% file: tables/6_long_context_downstream_task.tex
\begin{table}[!t]
\centering
\caption{Effect of Long-Context Training on Downstream Performance. Evaluate on Multiple-Choice tasks (Belebele, XCOPA, M3Exam) with Accuracy as metric, and Reading Comprehension tasks (XQuAD and TydiQA) with Exact/Fuzzy Match as metrics.}
\label{tab:downstream_task_perf}
\renewcommand{\arraystretch}{1.2}  
\resizebox{1\linewidth}{!}{%
\setlength{\tabcolsep}{3pt}
\begin{tabular}{lccccccccccccc}
\toprule
 & \multicolumn{3}{c}{\textbf{Belebele}} & \multicolumn{3}{c}{\textbf{XCOPA}} & \multicolumn{2}{c}{\textbf{XQuAD}} & \multicolumn{1}{c}{\textbf{TydiQA}} & \multicolumn{3}{c}{\textbf{M3Exam}} \\
\cmidrule(lr){2-4}\cmidrule(lr){5-7}\cmidrule(lr){8-9}\cmidrule(lr){10-10}\cmidrule(lr){11-13}
\textbf{Model} & \textbf{tha} & \textbf{ind} & \textbf{vie} & \textbf{tha} & \textbf{ind} & \textbf{vie} & \textbf{tha} & \textbf{vie} & \textbf{ind} & \textbf{tha} & \textbf{ind} & \textbf{vie} \\
\midrule
\midrule
Sailor2-1B       & 36.89 & 35.89 & 36.78 & 56.6 & 66.8 & 68.0 & 33.07 / 49.60 & 34.56 / 53.43 & 44.78 / 64.86 & 28.43 & 28.30 & 36.84 \\
Sailor2-1B-32K   & 36.44 & 36.56 & 35.89 & 56.2 & 68.4 & 65.8 & 35.68 / 54.86 & 38.49 / 58.91 & 45.84 / 65.08 & 27.79 & 27.76 & 35.10 \\
\midrule
Sailor2-8B       & 43.22 & 48.89 & 48.67 & 66.4 & 74.8 & 81.0 & 66.84 / 80.50 & 60.05 / 79.61 & 66.37 / 81.30 & 56.50 & 57.14 & 65.62 \\
Sailor2-8B-128K  & 44.00 & 50.44 & 48.89 & 66.8 & 79.0 & 81.2 & 67.10 / 80.28 & 59.02 / 79.15 & 65.31 / 82.01 & 55.39 & 60.11 & 65.01 \\
\midrule
Sailor2-20B      & 47.44 & 52.11 & 53.78 & 67.6 & 81.4 & 83.6 & 69.45 / 83.34 & 62.02 / 82.05 & 71.68 / 84.44 & 67.77 & 62.26 & 74.46 \\
Sailor2-20B-128K & 48.44 & 52.44 & 53.44 & 69.4 & 81.8 & 84.8 & 69.97 / 83.54 & 63.47 / 82.10 & 70.44 / 83.69 & 67.36 & 62.80 & 74.23 \\
\bottomrule
\end{tabular}
}
\end{table}

%% file: tables/6_pruning_swb_results.tex
\begin{table}[htbp]
\centering
\caption{Evaluation of Chat Model after Pruning.}
\resizebox{\textwidth}{!}{%
\begin{tabular}{lccccccccc}
\toprule
\textbf{Model} & \textbf{SWB Score} & \textbf{tha} & \textbf{vie} & \textbf{ind} & \textbf{tgl} & \textbf{zsm} & \textbf{khm} & \textbf{lao} & \textbf{mya} \\
\midrule
\midrule
Qwen2.5-14B-Chat & 0.30 & 0.40 & 0.40 & 0.23 & 0.35 & 0.20 & 0.21 & 0.12 & 0.30 \\
Sailor2-14B-Chat & 0.39 & 0.38 & 0.34 & 0.33 & 0.35 & 0.28 & 0.46 & 0.47 & 0.43 \\
\midrule
\midrule
Qwen2.5-3B-Chat & 0.16 & 0.14 & 0.21 & 0.18 & 0.08 & 0.16 & 0.06 & 0.06 & 0.04 \\
Sailor2-3B-Chat & 0.26 & 0.25 & 0.21 & 0.21 & 0.19 & 0.19 & 0.32 & 0.31 & 0.28 \\
\bottomrule
\end{tabular}%
}
\label{table:pruning_swb_perf}
\end{table}

%% file: sections/7.evaluation.tex
\section{Evaluation}

\subsection{Evaluation on Base Model}

For base model evaluation, we focus on the basic language understanding task like sentence classification, and language generation task like question answering and machine translation.
Specially, we evaluate Sailor2 on SailCompass~\citep{sailcompass} evaluation suite and FLoRes-200~\citep{nllb2022} translation suite.
To expand the evaluated language coverage, we choose the dataset in Indonesian, Thai, Vietnamese, Malay and Javanese. 

For Indonesian, we choose IndoCulture~\citep{koto2024indoculture}, TydiQA~\citep{tydiqa}, Belebele~\citep{bandarkar-etal-2024-belebele}.
For Thai, we choose MMLU~\citep{kydlicek2024finetasksmultilingualtasks}, M3Exam~\citep{zhang2023m3exam}\footnote{For Thai M3Exam, we adopt finetasks~\citep{kydlicek2024finetasksmultilingualtasks} codebase for evaluation.} and Belebele.
For Vietnamese, We choose VMLU~\footnote{VMLU:~\url{https://github.com/ZaloAI-Jaist/VMLU/}}, M3Exam and Belebele.
For Malay, we choose Tatabahasa~\citep{lovenia2024seacrowd}.
For Javanese, we choose M3Exam.
For all SEA languages, we choose FLoRes-200~\citep{nllb2022} and XCOPA~\citep{ponti2020xcopa}. 
We have more detailed comparison and analysis for translation in Section~\ref{sec:analysis_translation} and culture understanding in Section~\ref{sec:analysis_culture}. 

Detailed results are presented in Table~\ref{tab:base_model_pref_overview}. We observe that both Sailor2-8B and Sailor2-20B exhibit the highest average performance within their respective parameter groups. Notably, Sailor2-20B even outperforms larger models, including the three-times larger Llama3.1-70B.

\input{tables/7_base_model_evaluation}

\subsection{Evaluation on Chat Model}

We aim to comprehensively evaluate the performance of our Chat Model by using WildBench~\citep{lin2024wildbenchbenchmarkingllmschallenging} as the primary evaluation dataset. WildBench covers five tasks: Coding \& Debugging, Information Seeking, Math \& Data, Reasoning \& Planning, and Creative Tasks. We employ GPT-4o-0806 to translate WildBench into eight SEA languages (Thai, Vietnamese, Indonesian, Tagalog, Burmese, Khmer, Lao, and Malay), thereby creating a new benchmark named SEA-WildBench~(SWB).

Detailed results are presented in Table~\ref{tab:chat_perf_task_level} (task-level) and Table~\ref{tab:chat_perf_language_level} (language-level). We use the SWB Score as our evaluation metric, which is calculated based on the win-rate against GPT-4o-0806 (the same model serves as the judge). We selected the most representative open models, including both general-purpose and SEA language-optimized variants. For improved visualization, we use Llama-3.1-70B-Instruct as the baseline, with a SWB Score of 30. Our results indicate that both Sailor2-20B-Chat and Sailor2-8B-Chat achieve superior performance across various tasks and languages. 
As shown in Table~\ref{tab:chat_perf_language_level}, Sailor2 models excel in low-resource languages. Notably, Sailor2-20B-Chat achieves nearly a 50\% win rate against GPT-4o-0806 on SeaWildBench, demonstrating GPT-4o-level performance in local chat scenarios for Southeast Asian languages.

Note that the overall SWB Score can be higher than the scores for individual subsets. For example, although Llama-2-7B-Chat scores below 0.05 on each subset, its overall SWB Score is 0.05. 
We follow the WildBench score calculation~\footnote{\url{https://tinyurl.com/49en4cw6}}.
This method may overestimate scores in cases where parse errors occur.

\input{tables/8_swb_split_by_task}

\input{tables/8_swb_split_by_language}

%% file: tables/7_base_model_evaluation.tex
\begin{table}[]
\centering
\caption{\textbf{Overview of results on Sailor2}, over both 8B and 20B models. 
The best performing model for each model size on each benchmark is bolded. 
 }
\setlength\tabcolsep{5pt}
\adjustbox{max width=\linewidth}{
\begin{NiceTabular}{@{}ll|P{38pt}C{38pt}C{38pt}C{38pt}C{38pt}|P{38pt}C{38pt}C{38pt}C{38pt}C{38pt}@{}}
\toprule
\textbf{Language} & \textbf{Benchmark}$_\text{(eval)}$ &   \textbf{\textbf{Sailor2-8B}} & \textbf{Qwen2.5-7B} & \textbf{Gemma2-9B} & \textbf{Lllama3.1-8B} & \textbf{SeaLLM-v3-7B} & \textbf{\textbf{Sailor2-20B}} & \textbf{Qwen2.5-32B} & \textbf{Gemma2-27B} & \textbf{Llama3.1-70B} & \textbf{Aya-Expanse-32B} \\
\midrule
    & \textbf{Avg.} & \textbf{57.6} & 52.8 & 52.5 & 47.2 & 43.4 & \textbf{62.8} & 59.1\phantom{$^{\diamondsuit}$} & 61.8\phantom{$^{\diamondsuit}$} & 61.2\phantom{$^{\diamondsuit}$}  & 51.1\phantom{$^{\diamondsuit}$} \\
\midrule
    \multirow{3}[1]{*}{Indonesian} & IndoCulture$_\text{(0 shot)}$ & \textbf{73.4} & 58.7 & 65.6 & 56.7 & 53.0 & \textbf{76.4} & 68.9\phantom{$^{\diamondsuit}$} & 66.1\phantom{$^{\diamondsuit}$} & 72.7\phantom{$^{\diamondsuit}$}  & 70.6\phantom{$^{\diamondsuit}$} \\
    & TydiQA$_\text{(3 shot)}$ & \textbf{66.4} & 63.5 & 65.5 & 63.4 & 65.5 & \textbf{71.7} & 63.9\phantom{$^{\diamondsuit}$} & 65.1\phantom{$^{\diamondsuit}$} & 69.9\phantom{$^{\diamondsuit}$}  & 58.2\phantom{$^{\diamondsuit}$}\\
    & Belebele$_\text{(3 shot)}$ & 48.9 & 49.3 & \textbf{50.7} & 46.8 & 30.6 & 52.1 & 54.1\phantom{$^{\diamondsuit}$} & 53.3\phantom{$^{\diamondsuit}$} & 56.4\phantom{$^{\diamondsuit}$}  & \textbf{60.3}\phantom{$^{\diamondsuit}$}\\
    \midrule
    \multirow{3}[1]{*}{Thai} & MMLU$_\text{(5 shot)}$ & 55.4 & 52.8 & \textbf{57.8} & 44.1 & 50.8 & 66.3 & \textbf{70.7}\phantom{$^{\diamondsuit}$} & 62.5\phantom{$^{\diamondsuit}$} & 67.1\phantom{$^{\diamondsuit}$}  & 39.6\phantom{$^{\diamondsuit}$} \\
    & M3Exam$_\text{(5 shot)}$ & \textbf{57.0} & 51.7 & 52.7 & 43.7 & 51.3 & \textbf{69.3} & 69.2\phantom{$^{\diamondsuit}$} & 57.0\phantom{$^{\diamondsuit}$} & 63.7\phantom{$^{\diamondsuit}$} & 38.6\phantom{$^{\diamondsuit}$} \\
    & Belebele$_\text{(3 shot)}$ & 43.2 & 44.1  & 40.6 & 43.1 & 43.0 & 47.4 & 49.4\phantom{$^{\diamondsuit}$} & 46.0\phantom{$^{\diamondsuit}$} & \textbf{52.3}\phantom{$^{\diamondsuit}$}  & 45.3\phantom{$^{\diamondsuit}$}\\
    \midrule
    \multirow{3}[1]{*}{Vietnamese} & VMLU$_\text{(3 shot)}$ & 56.2 & 52.6 & 51.7 & 48.9 & \textbf{56.8} & \textbf{65.9} & 64.9\phantom{$^{\diamondsuit}$} & 59.1\phantom{$^{\diamondsuit}$} & 63.9\phantom{$^{\diamondsuit}$}  & 65.9\phantom{$^{\diamondsuit}$} \\
    & M3Exam$_\text{(3 shot)}$ & 65.6 & \textbf{66.4} & 65.5 & 54.4 & 63.1 & 74.6 & \textbf{77.3}\phantom{$^{\diamondsuit}$} & 68.6\phantom{$^{\diamondsuit}$} & 68.9\phantom{$^{\diamondsuit}$} & 63.2\phantom{$^{\diamondsuit}$}\\
    & Belebele$_\text{(3 shot)}$ & 48.7 & \textbf{50.8}  & 49.0 & 46.0 & 48.6 & 53.8 & 54.6\phantom{$^{\diamondsuit}$} & 52.0\phantom{$^{\diamondsuit}$} & \textbf{61.8}\phantom{$^{\diamondsuit}$}  & 58.3\phantom{$^{\diamondsuit}$}\\
    \midrule
    \multirow{1}[1]{*}{Malay} & Tatabahasa$_\text{(3 shot)}$ & \textbf{67.3} & 41.5 & 53.6 & 42.9 & 37.4 & \textbf{67.3} & 50.4\phantom{$^{\diamondsuit}$} & 58.6\phantom{$^{\diamondsuit}$} & 58.3\phantom{$^{\diamondsuit}$} & 48.1\phantom{$^{\diamondsuit}$} \\
    \midrule
    \multirow{1}[1]{*}{Javanese} & M3Exam$_\text{(3 shot)}$ & \textbf{57.1} & 35.9 & 45.3 & 40.4 & 38.5 & \textbf{62.3} & 47.7\phantom{$^{\diamondsuit}$} & 49.1\phantom{$^{\diamondsuit}$} & 53.4\phantom{$^{\diamondsuit}$} & 46.1\phantom{$^{\diamondsuit}$} \\
    \midrule
    \multirow{2}[1]{*}{Multiple} & FLORES-200$_\text{(3 shot)}$ & 35.4 & 30.6 & \textbf{35.8} & 31.7 & 29.6 & 35.8 & 34.3\phantom{$^{\diamondsuit}$} & \textbf{36.6}\phantom{$^{\diamondsuit}$} & 36.5\phantom{$^{\diamondsuit}$} & 35.7\phantom{$^{\diamondsuit}$} \\
    & XCOPA$_\text{(3 shot)}$ & \textbf{74.1} & 71.8 & 73.0 & 69.4 & 70.4 & 77.5 & 77.3\phantom{$^{\diamondsuit}$} & 75.3\phantom{$^{\diamondsuit}$} & \textbf{79.8}\phantom{$^{\diamondsuit}$} & 72.1\phantom{$^{\diamondsuit}$} \\
\bottomrule
\end{NiceTabular}}
\vspace{3pt}
\label{tab:base_model_pref_overview}
\end{table}

%% file: tables/8_swb_split_by_task.tex
\begin{table}[!htp]\centering
\caption{
Task-Level Evaluation of Chat Models on SEA-WildBench. 
The score represents the win rate against GPT-4o, which also serves as the evaluator. 
SWB Score is the average score of five tasks.
}
\label{tab:chat_perf_task_level}
\scriptsize
\begin{tabular}{lcccccccc}\toprule
\textbf{Model} & \textbf{SWB Score} & \textbf{Coding} & \textbf{Creative Tasks} & \textbf{Info Seeking} & \textbf{Reasoning} & \textbf{Math} & \textbf{Length} \\
\midrule
\midrule
Sailor2-20B-Chat &\cellcolor[HTML]{77c8a0}0.56 &\cellcolor[HTML]{57bb8a}0.62 &\cellcolor[HTML]{77c8a0}0.56 &\cellcolor[HTML]{6cc499}0.58 &\cellcolor[HTML]{72c69d}0.57 &\cellcolor[HTML]{81cca8}0.54 &2814.74 \\\cmidrule{1-8}
Sailor2-8B-Chat &\cellcolor[HTML]{9cd7ba}0.49 &\cellcolor[HTML]{c0e6d4}0.42 &\cellcolor[HTML]{72c69d}0.57 &\cellcolor[HTML]{87cfab}0.53 &\cellcolor[HTML]{96d5b6}0.50 &\cellcolor[HTML]{c0e6d4}0.42 &2849.41 \\\cmidrule{1-8}
Qwen2.5-72B-Instruct &\cellcolor[HTML]{b1e0c9}0.45 &\cellcolor[HTML]{96d5b6}0.50 &\cellcolor[HTML]{d0ecdf}0.39 &\cellcolor[HTML]{b6e2cc}0.44 &\cellcolor[HTML]{b1e0c9}0.45 &\cellcolor[HTML]{9cd7ba}0.49 &3026.82 \\\cmidrule{1-8}
SEA-LIONv3-70B-Instruct &\cellcolor[HTML]{cbeadb}0.40 &\cellcolor[HTML]{c0e6d4}0.42 &\cellcolor[HTML]{d5eee2}0.38 &\cellcolor[HTML]{cbeadb}0.40 &\cellcolor[HTML]{d0ecdf}0.39 &\cellcolor[HTML]{d0ecdf}0.39 &2340.65 \\\cmidrule{1-8}
Gemma-2-27B-Instruct &\cellcolor[HTML]{cbeadb}0.40 &\cellcolor[HTML]{d5eee2}0.38 &\cellcolor[HTML]{c6e8d7}0.41 &\cellcolor[HTML]{d0ecdf}0.39 &\cellcolor[HTML]{d0ecdf}0.39 &\cellcolor[HTML]{dbf1e6}0.37 &2288.33 \\\cmidrule{1-8}
Qwen2.5-32B-Instruct &\cellcolor[HTML]{f5fbf8}0.32 &\cellcolor[HTML]{d0ecdf}0.39 &\cellcolor[HTML]{fdf5f5}0.28 &\cellcolor[HTML]{fefafa}0.29 &\cellcolor[HTML]{f5fbf8}0.32 &\cellcolor[HTML]{f0f9f5}0.33 &2090.61 \\\midrule
Gemma-2-9B-Instruct &\cellcolor[HTML]{fafdfc}0.31 &\cellcolor[HTML]{fbeceb}0.26 &\cellcolor[HTML]{e0f3ea}0.36 &\cellcolor[HTML]{f0f9f5}0.33 &0.30 &\cellcolor[HTML]{fbeceb}0.26 &2163.03 \\
Qwen2.5-14B-Instruct &0.30 &\cellcolor[HTML]{f0f9f5}0.33 &\cellcolor[HTML]{fae7e6}0.25 &\cellcolor[HTML]{fdf5f5}0.28 &\cellcolor[HTML]{fdf5f5}0.28 &0.30 &2267.94 \\
Llama-3.1-70B-Instruct &0.30 &\cellcolor[HTML]{dbf1e6}0.37 &\cellcolor[HTML]{fbeceb}0.26 &\cellcolor[HTML]{fdf5f5}0.28 &\cellcolor[HTML]{fdf5f5}0.28 &\cellcolor[HTML]{fdf5f5}0.28 &2543.06 \\
SEA-LIONv3-8B-Instruct &0.30 &\cellcolor[HTML]{f5fbf8}0.32 &\cellcolor[HTML]{f5fbf8}0.32 &0.30 &\cellcolor[HTML]{fdf5f5}0.28 &\cellcolor[HTML]{f7d9d7}0.22 &2357.14 \\
Aya-Expanse-32B &\cellcolor[HTML]{fefafa}0.29 &\cellcolor[HTML]{fefafa}0.29 &\cellcolor[HTML]{fdf5f5}0.28 &\cellcolor[HTML]{fdf5f5}0.28 &\cellcolor[HTML]{fcf0f0}0.27 &\cellcolor[HTML]{f9e2e1}0.24 &2495.47 \\
Qwen2-72B-Instruct &\cellcolor[HTML]{fbeceb}0.26 &\cellcolor[HTML]{f7d9d7}0.22 &\cellcolor[HTML]{fcf0f0}0.27 &\cellcolor[HTML]{fdf5f5}0.28 &\cellcolor[HTML]{fae7e6}0.25 &\cellcolor[HTML]{f8dedc}0.23 &1546.21 \\
Qwen2.5-7B-Instruct &\cellcolor[HTML]{fae7e6}0.25 &\cellcolor[HTML]{fdf5f5}0.28 &\cellcolor[HTML]{f6d0cd}0.20 &\cellcolor[HTML]{f8dedc}0.23 &\cellcolor[HTML]{f7d9d7}0.22 &\cellcolor[HTML]{f7d9d7}0.22 &2415.08 \\
SEA-LIONv2.1-8B-Instruct &\cellcolor[HTML]{f8dedc}0.23 &\cellcolor[HTML]{f8dedc}0.23 &\cellcolor[HTML]{f9e2e1}0.24 &\cellcolor[HTML]{f9e2e1}0.24 &\cellcolor[HTML]{f6d0cd}0.20 &\cellcolor[HTML]{f4c6c3}0.18 &1735.26 \\
SeaLLMs-v3-7B-Chat &\cellcolor[HTML]{f6d4d2}0.21 &\cellcolor[HTML]{f6d4d2}0.21 &\cellcolor[HTML]{f5cbc8}0.19 &\cellcolor[HTML]{f5cbc8}0.19 &\cellcolor[HTML]{f4c6c3}0.18 &\cellcolor[HTML]{f1b8b4}0.15 &2298.47 \\
Llama-3.1-8B-Instruct &\cellcolor[HTML]{f5cbc8}0.19 &\cellcolor[HTML]{f4c6c3}0.18 &\cellcolor[HTML]{f1b8b4}0.15 &\cellcolor[HTML]{f2bdb9}0.16 &\cellcolor[HTML]{f1b8b4}0.15 &\cellcolor[HTML]{efafaa}0.13 &2356.67 \\
SeaLLM-7B-v2 &\cellcolor[HTML]{f4c6c3}0.18 &\cellcolor[HTML]{f0b4af}0.14 &\cellcolor[HTML]{f2bdb9}0.16 &\cellcolor[HTML]{f3c2be}0.17 &\cellcolor[HTML]{f0b4af}0.14 &\cellcolor[HTML]{eeaaa5}0.12 &2298.15 \\
SeaLLM-7B-v2.5 &\cellcolor[HTML]{f3c2be}0.17 &\cellcolor[HTML]{f0b4af}0.14 &\cellcolor[HTML]{f0b4af}0.14 &\cellcolor[HTML]{f1b8b4}0.15 &\cellcolor[HTML]{efafaa}0.13 &\cellcolor[HTML]{eea6a0}0.11 &2184.55 \\
Qwen2.5-3B-Instruct &\cellcolor[HTML]{f2bdb9}0.16 &\cellcolor[HTML]{f0b4af}0.14 &\cellcolor[HTML]{eda19b}0.10 &\cellcolor[HTML]{eeaaa5}0.12 &\cellcolor[HTML]{eeaaa5}0.12 &\cellcolor[HTML]{efafaa}0.13 &2324.08 \\
Sailor-14B-Chat &\cellcolor[HTML]{f2bdb9}0.16 &\cellcolor[HTML]{ea938c}0.07 &\cellcolor[HTML]{eea6a0}0.11 &\cellcolor[HTML]{efafaa}0.13 &\cellcolor[HTML]{eda19b}0.10 &\cellcolor[HTML]{ec9c96}0.09 &2465.85 \\
SeaLLM-7B-v1 &\cellcolor[HTML]{eeaaa5}0.12 &\cellcolor[HTML]{e68078}0.03 &\cellcolor[HTML]{ea938c}0.07 &\cellcolor[HTML]{ec9c96}0.09 &\cellcolor[HTML]{ea938c}0.07 &\cellcolor[HTML]{e98e87}0.06 &2585.40 \\
Mistral-7B-Instruct-v0.3 &\cellcolor[HTML]{eda19b}0.10 &\cellcolor[HTML]{eea6a0}0.11 &\cellcolor[HTML]{e68078}0.03 &\cellcolor[HTML]{ea938c}0.07 &\cellcolor[HTML]{e98e87}0.06 &\cellcolor[HTML]{ea938c}0.07 &2336.51 \\
Sailor-7B-Chat &\cellcolor[HTML]{ec9c96}0.09 &\cellcolor[HTML]{e67c73}0.02 &\cellcolor[HTML]{e7857d}0.04 &\cellcolor[HTML]{e98e87}0.06 &\cellcolor[HTML]{e7857d}0.04 &\cellcolor[HTML]{e68078}0.03 &1404.60 \\
Llama-2-70B-Chat &\cellcolor[HTML]{eb9891}0.08 &\cellcolor[HTML]{ea938c}0.07 &\cellcolor[HTML]{e88a82}0.05 &\cellcolor[HTML]{e98e87}0.06 &\cellcolor[HTML]{e88a82}0.05 &\cellcolor[HTML]{e88a82}0.05 &2354.30 \\
Llama-2-13B-Chat &\cellcolor[HTML]{e98e87}0.06 &\cellcolor[HTML]{e7857d}0.04 &\cellcolor[HTML]{e7857d}0.04 &\cellcolor[HTML]{e88a82}0.05 &\cellcolor[HTML]{e68078}0.03 &\cellcolor[HTML]{e68078}0.03 &2317.36 \\
Llama-2-7B-Chat &\cellcolor[HTML]{e88a82}0.05 &\cellcolor[HTML]{e68078}0.03 &\cellcolor[HTML]{e67c73}0.02 &\cellcolor[HTML]{e7857d}0.04 &\cellcolor[HTML]{e67c73}0.02 &\cellcolor[HTML]{e68078}0.03 &2330.50 \\
\bottomrule
\end{tabular}
\end{table}

%% file: tables/8_swb_split_by_language.tex
\begin{table}[!htp]\centering
\caption{
Language-Level Evaluation of Chat Models on SEA-Wildbench. 
The score represents the win rate against GPT-4o, which also serves as the evaluator.
SWB Score is the average score of eight languages.
}
\label{tab:chat_perf_language_level}
\scriptsize
\begin{tabular}{lccccccccccc}\toprule
\textbf{Model} & \textbf{SWB Score} & \textbf{tha} &\textbf{vie} &\textbf{ind} &\textbf{tgl} &\textbf{zsm} &\textbf{khm} &\textbf{lao} &\textbf{mya} &\textbf{Length} \\
\midrule\midrule
Sailor2-20B-Chat &\cellcolor[HTML]{8fd2b1}0.56 &\cellcolor[HTML]{9cd7ba}0.53 &\cellcolor[HTML]{a9ddc4}0.50 &\cellcolor[HTML]{98d6b7}0.54 &\cellcolor[HTML]{a9ddc4}0.50 &\cellcolor[HTML]{aedec6}0.49 &\cellcolor[HTML]{71c69c}0.63 &\cellcolor[HTML]{57bb8a}0.69 &\cellcolor[HTML]{6dc499}0.64 &2814.74 \\\cmidrule{1-11}
Sailor2-8B-Chat &\cellcolor[HTML]{aedec6}0.49 &\cellcolor[HTML]{b2e0c9}0.48 &\cellcolor[HTML]{bbe4cf}0.46 &\cellcolor[HTML]{bbe4cf}0.46 &\cellcolor[HTML]{ccebdb}0.42 &\cellcolor[HTML]{c7e9d8}0.43 &\cellcolor[HTML]{a9ddc4}0.50 &\cellcolor[HTML]{64c193}0.66 &\cellcolor[HTML]{94d4b4}0.55 &2849.41 \\\cmidrule{1-11}
Qwen2.5-72B-Instruct &\cellcolor[HTML]{bfe5d3}0.45 &\cellcolor[HTML]{98d6b7}0.54 &\cellcolor[HTML]{a5dbc0}0.51 &\cellcolor[HTML]{a5dbc0}0.51 &\cellcolor[HTML]{ccebdb}0.42 &\cellcolor[HTML]{b2e0c9}0.48 &\cellcolor[HTML]{f3faf6}0.33 &\cellcolor[HTML]{d0ecdf}0.41 &\cellcolor[HTML]{fbfefc}0.31 &3026.82 \\\cmidrule{1-11}
SEA-LIONv3-70B-Instruct &\cellcolor[HTML]{d4eee2}0.40 &\cellcolor[HTML]{bfe5d3}0.45 &\cellcolor[HTML]{bfe5d3}0.45 &\cellcolor[HTML]{b2e0c9}0.48 &\cellcolor[HTML]{d4eee2}0.40 &\cellcolor[HTML]{d0ecdf}0.41 &\cellcolor[HTML]{f7fcf9}0.32 &\cellcolor[HTML]{fdf5f5}0.28 &\cellcolor[HTML]{f7fcf9}0.32 &2340.65 \\\cmidrule{1-11}
Gemma-2-27B-Instruct &\cellcolor[HTML]{d4eee2}0.40 &\cellcolor[HTML]{c7e9d8}0.43 &\cellcolor[HTML]{d4eee2}0.40 &\cellcolor[HTML]{bbe4cf}0.46 &\cellcolor[HTML]{d4eee2}0.40 &\cellcolor[HTML]{d9f0e4}0.39 &\cellcolor[HTML]{eef9f3}0.34 &\cellcolor[HTML]{ddf2e7}0.38 &\cellcolor[HTML]{fbfefc}0.31 &2288.33 \\\cmidrule{1-11}
Qwen2.5-32B-Instruct &\cellcolor[HTML]{f7fcf9}0.32 &\cellcolor[HTML]{e1f3ea}0.37 &\cellcolor[HTML]{ccebdb}0.42 &\cellcolor[HTML]{ccebdb}0.42 &\cellcolor[HTML]{fbeceb}0.26 &\cellcolor[HTML]{ddf2e7}0.38 &\cellcolor[HTML]{f9e3e2}0.24 &\cellcolor[HTML]{f5cdc9}0.19 &\cellcolor[HTML]{f2bfbb}0.16 &2090.61 \\\cmidrule{1-11}
Gemma-2-9B-Instruct &\cellcolor[HTML]{fbfefc}0.31 &\cellcolor[HTML]{e6f5ee}0.36 &\cellcolor[HTML]{d4eee2}0.40 &\cellcolor[HTML]{d9f0e4}0.39 &0.30 &\cellcolor[HTML]{ddf2e7}0.38 &\cellcolor[HTML]{f5cdc9}0.19 &\cellcolor[HTML]{f5cdc9}0.19 &\cellcolor[HTML]{f5cdc9}0.19 &2163.03 \\
Qwen2.5-14B-Instruct &0.30 &\cellcolor[HTML]{d4eee2}0.40 &\cellcolor[HTML]{d4eee2}0.40 &\cellcolor[HTML]{f8dfdd}0.23 &\cellcolor[HTML]{eaf7f1}0.35 &\cellcolor[HTML]{f6d1ce}0.20 &\cellcolor[HTML]{f7d6d3}0.21 &\cellcolor[HTML]{efada8}0.12 &0.30 &2267.94 \\
Llama-3.1-70B-Instruct &0.30 &\cellcolor[HTML]{f3faf6}0.33 &\cellcolor[HTML]{e1f3ea}0.37 &\cellcolor[HTML]{e1f3ea}0.37 &\cellcolor[HTML]{fdf5f5}0.28 &\cellcolor[HTML]{eaf7f1}0.35 &\cellcolor[HTML]{f4c8c5}0.18 &\cellcolor[HTML]{f2bbb6}0.15 &\cellcolor[HTML]{f5cdc9}0.19 &2543.06 \\
SEA-LIONv3-8B-Instruct &0.30 &\cellcolor[HTML]{ddf2e7}0.38 &\cellcolor[HTML]{d4eee2}0.40 &\cellcolor[HTML]{ddf2e7}0.38 &\cellcolor[HTML]{eef9f3}0.34 &\cellcolor[HTML]{eaf7f1}0.35 &\cellcolor[HTML]{efada8}0.12 &\cellcolor[HTML]{ec9b94}0.08 &\cellcolor[HTML]{f1b6b1}0.14 &2357.14 \\
Aya-expanse-32B &\cellcolor[HTML]{fefafa}0.29 &\cellcolor[HTML]{fae8e6}0.25 &\cellcolor[HTML]{bfe5d3}0.45 &\cellcolor[HTML]{bbe4cf}0.46 &\cellcolor[HTML]{fcf1f0}0.27 &\cellcolor[HTML]{eaf7f1}0.35 &\cellcolor[HTML]{ea928b}0.06 &\cellcolor[HTML]{efada8}0.12 &\cellcolor[HTML]{f0b2ac}0.13 &2495.47 \\
Qwen2-72B-Instruct &\cellcolor[HTML]{fbeceb}0.26 &\cellcolor[HTML]{fbeceb}0.26 &0.30 &\cellcolor[HTML]{f3faf6}0.33 &\cellcolor[HTML]{fefafa}0.29 &\cellcolor[HTML]{f7fcf9}0.32 &\cellcolor[HTML]{f6d1ce}0.20 &\cellcolor[HTML]{f6d1ce}0.20 &\cellcolor[HTML]{f2bfbb}0.16 &1546.21 \\
Qwen2.5-7B-Instruct &\cellcolor[HTML]{fae8e6}0.25 &0.30 &\cellcolor[HTML]{eaf7f1}0.35 &\cellcolor[HTML]{e6f5ee}0.36 &\cellcolor[HTML]{efada8}0.12 &\cellcolor[HTML]{fefafa}0.29 &\cellcolor[HTML]{eca099}0.09 &\cellcolor[HTML]{eca099}0.09 &\cellcolor[HTML]{ec9b94}0.08 &2415.08 \\
Sealionv2.1-8B-Instruct &\cellcolor[HTML]{f8dfdd}0.23 &0.30 &\cellcolor[HTML]{f3faf6}0.33 &\cellcolor[HTML]{fbfefc}0.31 &\cellcolor[HTML]{f2bfbb}0.16 &\cellcolor[HTML]{fdf5f5}0.28 &\cellcolor[HTML]{eb978f}0.07 &\cellcolor[HTML]{ec9b94}0.08 &\cellcolor[HTML]{eda49e}0.10 &1735.26 \\
SeaLLMs-v3-7B-Chat &\cellcolor[HTML]{f7d6d3}0.21 &\cellcolor[HTML]{f8dfdd}0.23 &\cellcolor[HTML]{f8dad8}0.22 &\cellcolor[HTML]{f7d6d3}0.21 &\cellcolor[HTML]{f5cdc9}0.19 &\cellcolor[HTML]{f2bfbb}0.16 &\cellcolor[HTML]{f2bbb6}0.15 &\cellcolor[HTML]{f2bfbb}0.16 &\cellcolor[HTML]{eca099}0.09 &2298.47 \\
Llama-3.1-8B-Instruct &\cellcolor[HTML]{f5cdc9}0.19 &\cellcolor[HTML]{f5cdc9}0.19 &\cellcolor[HTML]{fbeceb}0.26 &\cellcolor[HTML]{f7d6d3}0.21 &\cellcolor[HTML]{f2bbb6}0.15 &\cellcolor[HTML]{f4c8c5}0.18 &\cellcolor[HTML]{ea928b}0.06 &\cellcolor[HTML]{eb978f}0.07 &\cellcolor[HTML]{eb978f}0.07 &2356.67 \\
SeaLLM-7B-v2 &\cellcolor[HTML]{f4c8c5}0.18 &\cellcolor[HTML]{f4c8c5}0.18 &\cellcolor[HTML]{f4c8c5}0.18 &\cellcolor[HTML]{f5cdc9}0.19 &\cellcolor[HTML]{eca099}0.09 &\cellcolor[HTML]{f0b2ac}0.13 &\cellcolor[HTML]{eda49e}0.10 &\cellcolor[HTML]{efada8}0.12 &\cellcolor[HTML]{eca099}0.09 &2298.15 \\
SeaLLM-7B-v2.5 &\cellcolor[HTML]{f3c4c0}0.17 &\cellcolor[HTML]{f4c8c5}0.18 &\cellcolor[HTML]{f5cdc9}0.19 &\cellcolor[HTML]{f4c8c5}0.18 &\cellcolor[HTML]{eda49e}0.10 &\cellcolor[HTML]{f1b6b1}0.14 &\cellcolor[HTML]{ec9b94}0.08 &\cellcolor[HTML]{eea9a3}0.11 &\cellcolor[HTML]{ea928b}0.06 &2184.55 \\
Qwen2.5-3B-Instruct &\cellcolor[HTML]{f2bfbb}0.16 &\cellcolor[HTML]{f1b6b1}0.14 &\cellcolor[HTML]{f7d6d3}0.21 &\cellcolor[HTML]{f4c8c5}0.18 &\cellcolor[HTML]{ec9b94}0.08 &\cellcolor[HTML]{f2bfbb}0.16 &\cellcolor[HTML]{ea928b}0.06 &\cellcolor[HTML]{ea928b}0.06 &\cellcolor[HTML]{e88981}0.04 &2324.08 \\
Sailor-14B-Chat &\cellcolor[HTML]{f2bfbb}0.16 &\cellcolor[HTML]{eea9a3}0.11 &\cellcolor[HTML]{f3c4c0}0.17 &\cellcolor[HTML]{f1b6b1}0.14 &\cellcolor[HTML]{e88981}0.04 &\cellcolor[HTML]{f1b6b1}0.14 &\cellcolor[HTML]{e68077}0.02 &\cellcolor[HTML]{efada8}0.12 &\cellcolor[HTML]{ea928b}0.06 &2465.85 \\
SeaLLM-7B-v1 &\cellcolor[HTML]{efada8}0.12 &\cellcolor[HTML]{e98e86}0.05 &\cellcolor[HTML]{eb978f}0.07 &\cellcolor[HTML]{eb978f}0.07 &\cellcolor[HTML]{e88981}0.04 &\cellcolor[HTML]{e98e86}0.05 &\cellcolor[HTML]{eda49e}0.10 &\cellcolor[HTML]{eea9a3}0.11 &\cellcolor[HTML]{eca099}0.09 &2585.40 \\
Mistral-7B-Instruct-v0.3 &\cellcolor[HTML]{eda49e}0.10 &\cellcolor[HTML]{eb978f}0.07 &\cellcolor[HTML]{eea9a3}0.11 &\cellcolor[HTML]{eb978f}0.07 &\cellcolor[HTML]{ec9b94}0.08 &\cellcolor[HTML]{eea9a3}0.11 &\cellcolor[HTML]{e68077}0.02 &\cellcolor[HTML]{e7857c}0.03 &\cellcolor[HTML]{e68077}0.02 &2336.51 \\
Sailor-7B-Chat &\cellcolor[HTML]{eca099}0.09 &\cellcolor[HTML]{e88981}0.04 &\cellcolor[HTML]{eb978f}0.07 &\cellcolor[HTML]{e98e86}0.05 &\cellcolor[HTML]{e68077}0.02 &\cellcolor[HTML]{ea928b}0.06 &\cellcolor[HTML]{e68077}0.02 &\cellcolor[HTML]{eb978f}0.07 &\cellcolor[HTML]{e7857c}0.03 &1404.60 \\
Llama-2-70B-Chat &\cellcolor[HTML]{ec9b94}0.08 &\cellcolor[HTML]{e68077}0.02 &\cellcolor[HTML]{e98e86}0.05 &\cellcolor[HTML]{eea9a3}0.11 &\cellcolor[HTML]{ea928b}0.06 &\cellcolor[HTML]{f0b2ac}0.13 &\cellcolor[HTML]{e7857c}0.03 &\cellcolor[HTML]{e67c73}0.01 &\cellcolor[HTML]{e7857c}0.03 &2354.30 \\
Llama-2-13B-Chat &\cellcolor[HTML]{ea928b}0.06 &\cellcolor[HTML]{e67c73}0.01 &\cellcolor[HTML]{e98e86}0.05 &\cellcolor[HTML]{ec9b94}0.08 &\cellcolor[HTML]{e68077}0.02 &\cellcolor[HTML]{e7857c}0.03 &\cellcolor[HTML]{e67c73}0.01 &\cellcolor[HTML]{e67c73}0.01 &\cellcolor[HTML]{e7857c}0.03 &2317.36 \\
Llama-2-7B-Chat &\cellcolor[HTML]{e98e86}0.05 &\cellcolor[HTML]{e67c73}0.01 &\cellcolor[HTML]{e68077}0.02 &\cellcolor[HTML]{e98e86}0.05 &\cellcolor[HTML]{e88981}0.04 &\cellcolor[HTML]{e7857c}0.03 &\cellcolor[HTML]{e67c73}0.01 &\cellcolor[HTML]{e68077}0.02 &\cellcolor[HTML]{e88981}0.04 &2330.50 \\
\bottomrule
\end{tabular}
\end{table}

%% file: sections/8.analysis.tex
\section{Analysis}

This section presents our insights on building multilingual LLMs during both the continual pre-training and post-training stages. 
We also examine Sailor2's capabilities in translation and cultural understanding, two core components of practical multilingual applications.

\subsection{Effect of Model Expansion} 

For both Sailor and Sailor2, we adopt a continual pre-training (CPT) approach to efficiently develop multilingual LLMs by reusing computational resources. Unlike Sailor, Sailor2 incorporates model expansion, which creates additional capacity for learning new knowledge from the multilingual corpus.

We analyze the perplexity shift during CPT with and without model expansion, with detailed results shown in Table~\ref{tab:ppl_shift_sailor1_sailor2}. Experimental findings reveal that, compared to the Qwen1.5~→~Sailor transition, Qwen2.5~→~Sailor2 exhibits less degradation in English/Chinese and greater improvements in the target SEA languages during continual pre-training. Notably, even though Qwen2.5 is trained on 18T tokens versus Qwen1.5's 4T tokens, Sailor2 still achieves significant gains in SEA languages with only minor degradation in English.

We conduct a comprehensive examination across multiple languages in Figure~\ref{fig:ppl_shift}, which illustrates the PPL distribution shift for English, Chinese, and sixteen SEA languages. The results demonstrate that Sailor2 maintains its performance in English and Chinese while achieving significantly lower perplexity in SEA languages. Further discussions on future work in efficient CPT can be found in Section~\ref{sec:future_work_efficient_training}.

\input{tables/8_sailor_and_sailor2_ppl}

\subsection{Effect of Continual Pre-training}
Qwen2.5 models have already been trained on 18T tokens, meaning that many SEA tokens were likely seen during the pre-training stage. This raises the question of whether the expensive continual pre-training stage using an additional 400B SEA tokens is still necessary. To investigate, we conducted an ablation study with the following setup: post-training on both the vanilla Qwen2.5-7B base model and the Sailor2-8B base model, with the same post-training dataset and training steps. 
Detailed results are listed in Table~\ref{table:language_wise_reward}.
We could observe that CPT is essential, especially for low-resource languages like khm and lao.

\begin{table}[htbp]
\centering
\caption{Language-wise Score on SEA-WildBench between chat models trained using the Qwen and Sailor2 models. Note that Qwen2.5-7B-Chat is trained using the Sailor2 post-training pipeline.}
\resizebox{\textwidth}{!}{%
\begin{tabular}{lccccccccc}
\toprule
\textbf{Model} & \textbf{SWB Score} & \textbf{tha} & \textbf{vie} & \textbf{ind} & \textbf{tgl} & \textbf{zsm} & \textbf{khm} & \textbf{lao} & \textbf{mya} \\
\midrule
\midrule
Sailor2-8B-Chat & 0.43 & 0.44 & 0.40 & 0.40 & 0.39 & 0.36 & 0.43 & 0.56 & 0.44 \\
\midrule
Qwen2.5-7B-Chat (ours) & 0.25 & 0.31 & 0.34 & 0.33 & 0.21 & 0.31 & 0.11 & 0.21 & 0.08 \\
\bottomrule
\end{tabular}%
}
\label{table:language_wise_reward}
\end{table}

\subsection{Key Findings in Preference Data Construction}\label{sec.pdata}

We conduct a series of ablation studies to assess the impact of design choices in the preference data construction pipeline, as described in Sec.~\ref{sec.pt.data}.

\begin{figure}[h]
    \centering
    \begin{minipage}[t]{0.48\linewidth}
        \centering
        \includegraphics[width=\linewidth]{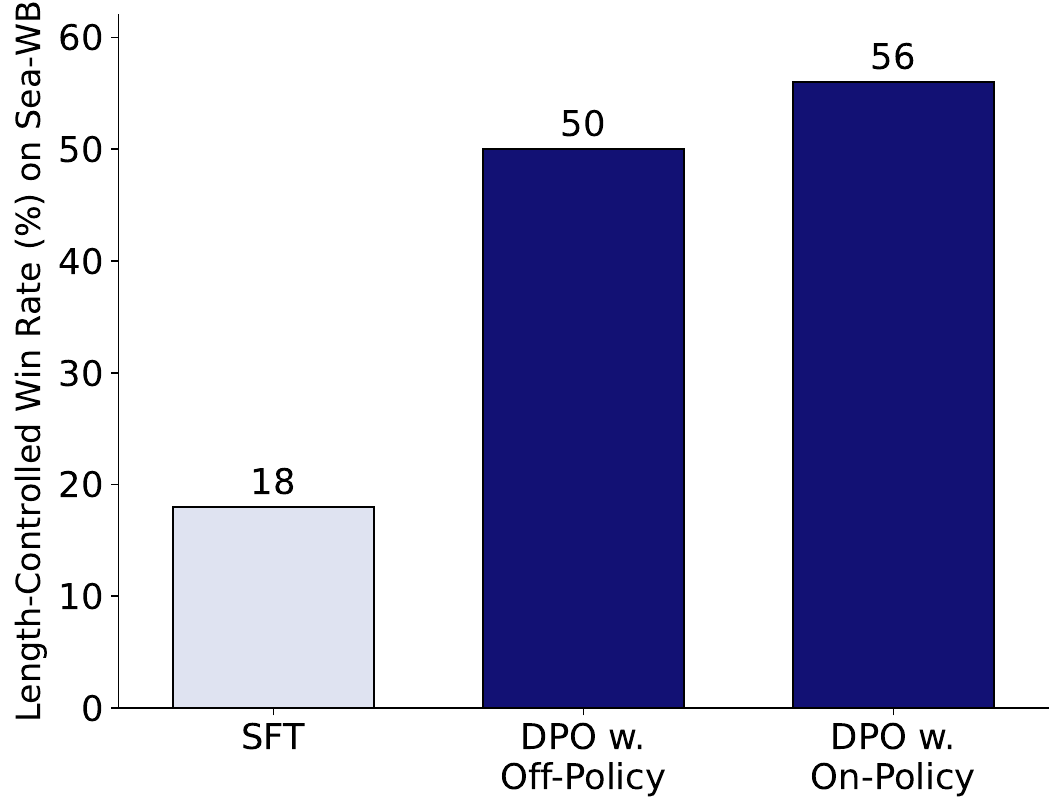}
        \caption{Length-controlled win rates comparison after different preference tuning stages on Sea-WB.}
        \label{fig:sec83_on_off}
    \end{minipage}
    \hfill
    \begin{minipage}[t]{0.48\linewidth}
        \centering
        \includegraphics[width=\linewidth]{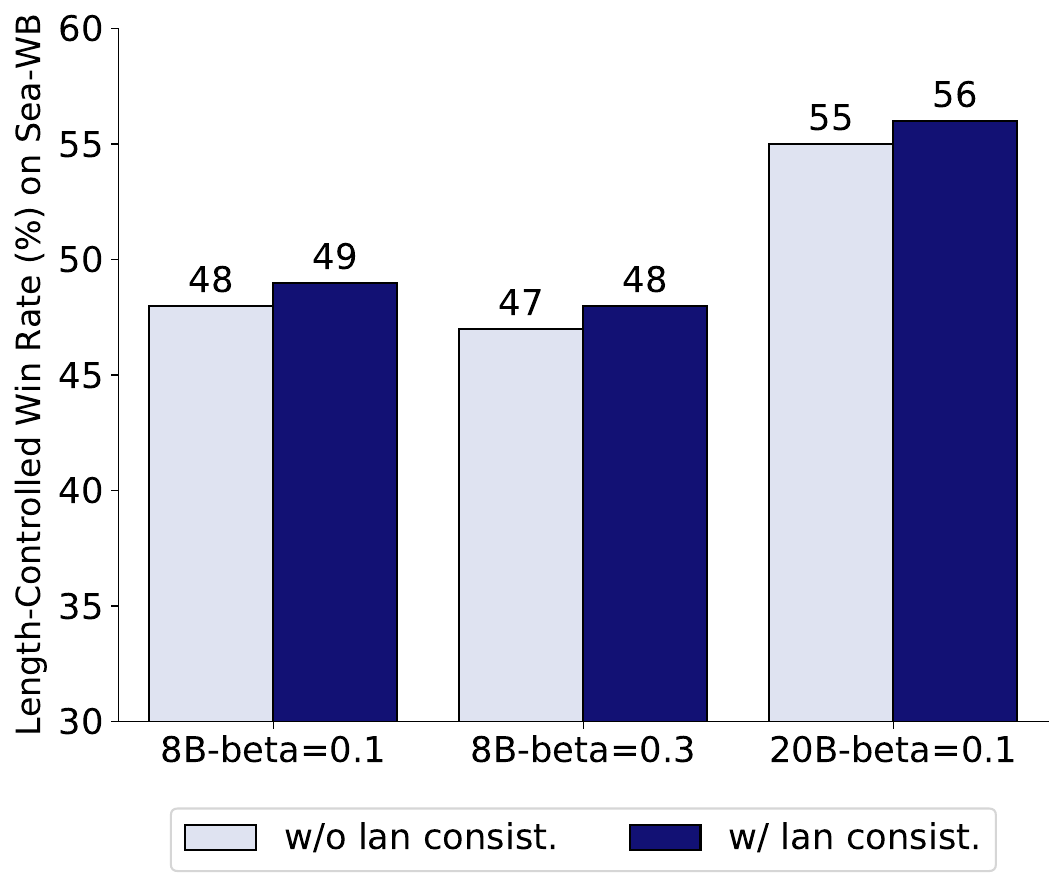}
        \caption{Effect of the language consistency verifier on the model performance. The results show a consistent performance gain across various experiment settings.}
        \label{fig:sec83_lc}
    \end{minipage}
\end{figure}

\paragraph{First Off-Policy, Then On-Policy Training Improves DPO Performance.}
While prior studies~\citep{guo2024direct,lambert2024t} have shown that on-policy training in DPO leads to greater performance improvements than off-policy training, our preliminary findings suggest that directly applying on-policy training to an SFT model yields limited gains. We hypothesize that this is due to the SFT model's insufficient ability to generate high-quality responses in chat tasks. To address this, we first perform off-policy training to initialize a stronger policy model before transitioning to on-policy training. In other words, our training pipeline consists of SFT, followed by off-policy DPO training, and then on-policy DPO training. As shown in Fig.~\ref{fig:sec83_on_off}, off-policy training on Sailor2-20B-SFT significantly enhances model performance, while subsequent on-policy training provides further improvements.

\paragraph{Language Consistency Verifier Improves Downstream DPO Performance.} 
Due to the lack of a reward model for SEA languages, we use Skywork-Reward-Gemma-2-27B~\citep{liu2024skywork} which is trained primarily on English data. To mitigate the RM's limitations in evaluating SEA language responses, we introduce a language consistency verifier. The verifier~\footnote{{We use \href{https://huggingface.co/facebook/fasttext-language-identification}{facebook/fasttext-language-identification} for language verification.}} labels a response as \texttt{true} if its language matches the prompt; otherwise, it is labeled \texttt{false}. If at least two responses are language-consistent, we select the winning and losing responses based on the RM's reward scores. If only one response is language-consistent, it is chosen as the winning response. If none are language-consistent, the prompt is discarded. 

We evaluate the verifier’s effectiveness under three settings: for the 8B model, we conduct on-policy training with $\beta\in\{0.1, 0.3\}$; for the 20B model, we use $\beta=0.1$. As shown in Fig.~\ref{fig:sec83_lc}, incorporating the verifier consistently improves performance compared to training without it. Moving forward, further improvements in preference tuning for low-resource languages could be explored, such as training an RM specifically for SEA languages and leveraging self-alignment techniques~\citep{chen2024self,chen2025bootstrapping,kim2025spread}.  

\begin{table}[h!]
\centering
\caption{Comparison of on-policy and distillation methods on Sailor2-8B after off-policy training. LC is short for the length-controlled win rate on Sea-WB.}
\begin{tabular}{lcc}
\toprule
\textbf{Method} & \textbf{LC} & \textbf{Avg. Length} \\
\midrule
On-policy & 0.49 & 2849 \\
Distillation & 0.48 & 2752 \\
\bottomrule
\end{tabular}
\label{tab.pdata.compare}
\end{table}

\paragraph{Distillation Reduces the Response Length while Maintaining Comparable Model Performance.}
We investigate the effectiveness of distillation within the same model family during DPO training given two practical considerations: (1) Distillation leverages the high-quality on-policy data used for training Sailor2-20B-Chat; (2) Reusing Sailor2-20B’s on-policy data significantly reduces the computational cost of data generation and reward evaluation. To assess its impact, we conduct a controlled experiment on our 8B model. Specifically, we conduct DPO with its own on-policy data and with the on-policy data from Sailor2-20B. Results in Table.~\ref{tab.pdata.compare} indicate that distillation maintains comparable model performance while reducing response length. 

\subsection{Cross-lingual Translation Ability of Sailor2}
\label{sec:analysis_translation}

We analyze the performance of Sailor2-20B on Flores Plus~\citep{nllb2022}, a translation dataset covering over 200 languages. Since our focus is primarily on SEA languages, we limit the scope to a subset of the dataset containing SEA languages, Chinese and English.
Table \ref{tab:flores_eng_xx_translation_performance} compares Sailor2 and three baseline models on English-centric translation pairs. 
Table \ref{tab:matrix-chrf-sailor2_20b} (and Table \ref{tab:matrix-chrf-qwen2_5_32b}, \ref{tab:matrix-chrf-qwen2_5_72b}, \ref{tab:matrix-chrf-llama3_1_70b} and \ref{tab:matrix-chrf-nllb_moe_54b}) shows performance of Sailor2-20B (and other baselines) between all language pairs in Flores Plus\footnote{The prediction results of Sailor2 and baseline models could be found in \url{https://huggingface.co/datasets/sailor2/Flores-Plus-Evaluation-Log-Preview-Cleaned}.}. We provide a visual comparison of Sailor2-20B against other baselines in Figure \ref{fig:flores_plus__qwen_32b}, \ref{fig:flores_plus__qwen_72b}, \ref{fig:flores_plus__llama_70b} and \ref{fig:flores_plus__nllb_moe_54b}.

\paragraph{Sailor2: Excelling in Low-Resource Translation}
Despite having significantly fewer parameters, Sailor2-20B demonstrates remarkable capabilities in low-resource language translation. 
The superior performance extends to approximately 80\% of low-resource language scenarios. 
As shown in Table \ref{tab:flores_eng_xx_translation_performance}, compared to Qwen2.5-32B and Qwen2.5-72B, Sailor2 consistently achieves higher chrF++ scores in low-resource language pairs, though showing slightly lower performance in English or Chinese translation. 
When compared with Llama3.1-70B, Sailor2-20B exhibits particular strengths in challenging low-resource scenarios, demonstrating significant advantages in Lao translation (7.3/37.6 vs 3.7/22.7), as well as when Burmese or Khmer is the target language.
The performance gap becomes even more pronounced when examining bidirectional translation capabilities - Sailor2-20B maintains relatively balanced performance in both directions for low-resource languages, while other models show significant degradation when translating into low-resource languages.
We also report the win rate of each model for each language pair in Figure \ref{fig:flores_plus_WR}. The win rate is defined as the percentage of times a model's output achieves the top-1 ChrF++ score.

\paragraph{Translation Patterns and Language Effects}
Besides the performance of Sailor2, analysis of Table \ref{tab:matrix-chrf-sailor2_20b} also reveals several important patterns in translation behavior:

(1) English demonstrates consistent superior performance across all language pairs, achieving higher chrF++ scores both as source and target language. This is clearly visible in Table \ref{tab:matrix-chrf-sailor2_20b}, where English-sourced translations consistently achieve scores above 50 chrF++ points for most target languages, significantly higher than other source languages. The performance advantage is particularly pronounced in the XX → English direction compared to English → XX translations, as evidenced by the consistently higher scores in both tables.

(2) We observe that translation quality appears more dependent on the target language than the source language, as shown in Table \ref{tab:matrix-chrf-sailor2_20b} where vertical columns (representing target languages) display more consistent score ranges compared to horizontal rows (source languages). For instance, translations into Vietnamese (vie) consistently fall within the 45-55 chrF++ range regardless of source language, while Vietnamese as a source language shows more variable performance depending on the target.

(3) Languages within the same family or system exhibit notably higher translation performance, as demonstrated in Table \ref{tab:matrix-chrf-sailor2_20b} by the Cebuano-Tagalog pair (ceb-fil: 55.5/53.3). This pattern suggests that linguistic similarity plays a crucial role in translation quality, potentially through the model's internal representation of language families. The table also reveals that geographically and culturally proximate languages, such as Indonesian-Malay pair (ind-zsm: 60.4/59.3), tend to achieve better bilateral translation performance compared to more distant language pairs.

\input{tables/8_flores_plus_eng_xx}

\input{tables/8_flores_chrf_sailor2_20b}

\subsection{SEA Culture Understanding Ability of Sailor2}
\label{sec:analysis_culture}

\begin{figure}[!htb]
    \centering
    \begin{subfigure}[b]{\textwidth}
        \centering
        \includegraphics[width=0.75\textwidth]{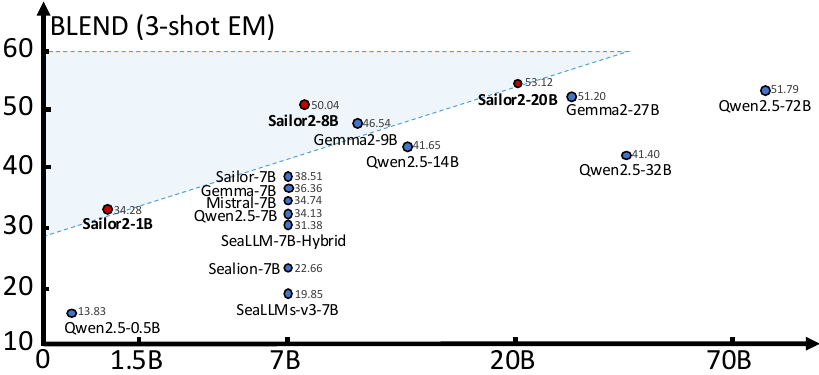}
        \caption{Results on BLEnD benchmark.}
        \label{fig:culture_blend}
    \end{subfigure}
    
    \vspace{0.5cm}

    \begin{subfigure}[b]{\textwidth}
        \centering
        \includegraphics[width=0.75\textwidth]{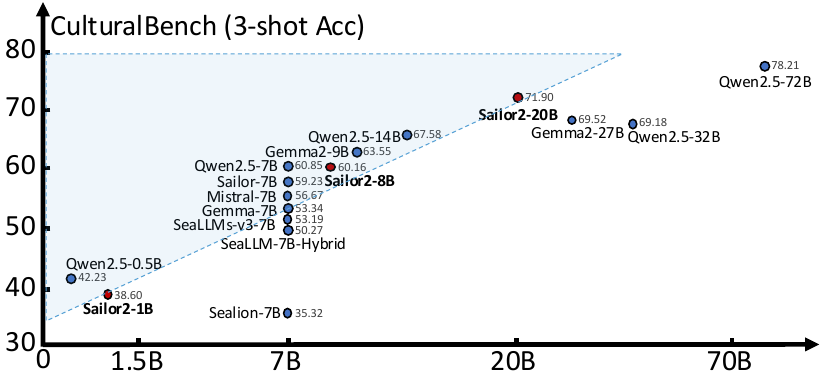}
        \caption{Results on CulturalBench benchmark.}
        \label{fig:culture_culturalbench}
    \end{subfigure}

    \caption{Performance comparison across models on BLEnD and CulturalBench benchmarks.}
    \label{fig:culture}
\end{figure}

Cultural understanding significantly influences the practical application and interaction quality of multilingual LLMs.
To assess the cultural understanding capabilities of the Sailor2 series models in Southeast Asian contexts, we employ CulturalBench~\citep{chiu2024culturalbench}, BLEND~\citep{myung2024blend}, and Global-MMLU~\citep{singh2024global} as evaluation benchmarks, covering a total of seven languages. 

Specifically, CulturalBench consists of single-choice and judgment questions in Filipino, Indonesian, Malaysian, Singaporean English, Thai, and Vietnamese.
BLEND includes question-and-answer tasks in Indonesian and West Java languages, while Global-MMLU comprises single-choice questions in Filipino, Indonesian, and Vietnamese.
Notably, BLEND and Global-MMLU are multilingual evaluation datasets. 
To further refine the measurement of the cultural understanding ability of LLMs, we translate CulturalBench into a multilingual version using Google Translate (from English to local languages).  

\begin{figure}[htbp]
    \centering
    \includegraphics[width=0.6\textwidth]{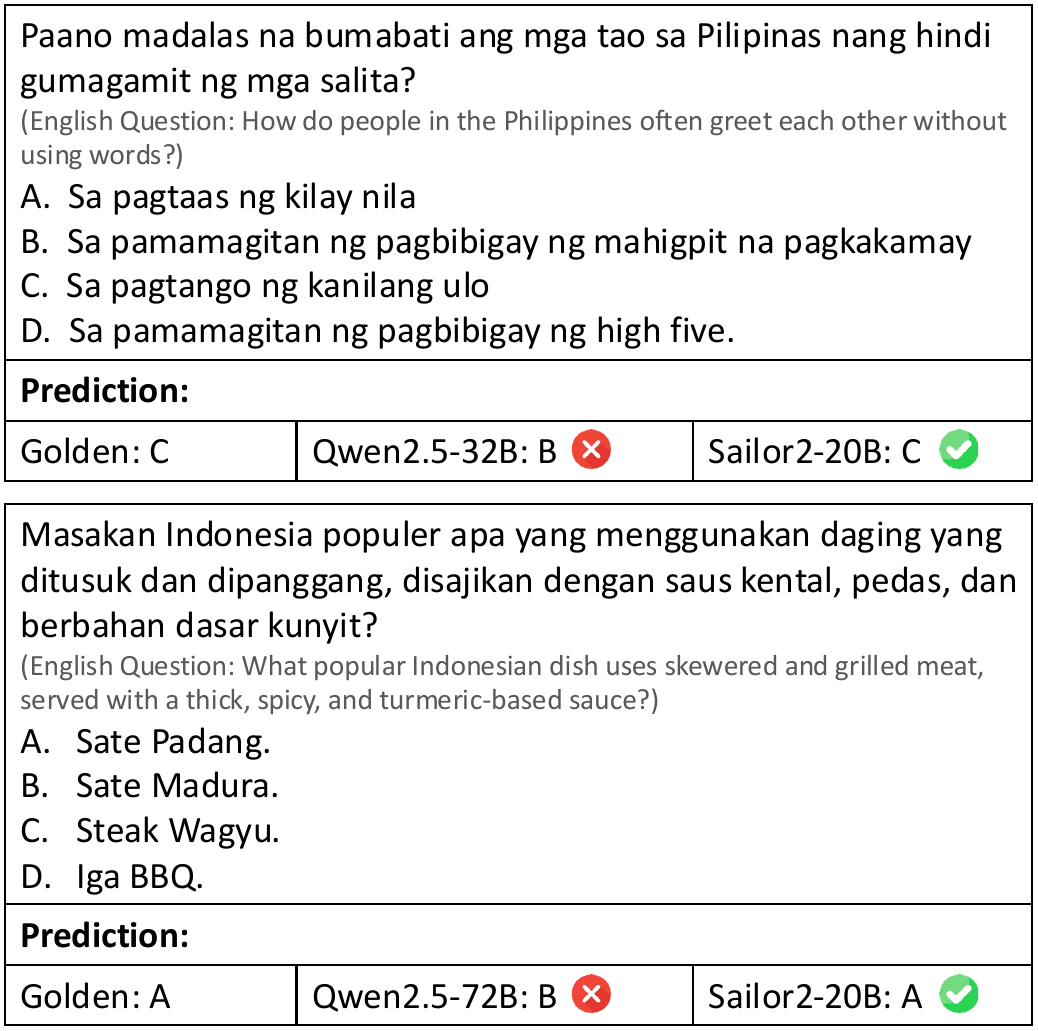}
    \caption{Sample responses from CulturalBench.}
    \label{fig:culturalbench_examples}
\end{figure}

All evaluations are conducted using the 3-shot prompting approach. 
The experimental results are presented in Tables~\ref{tab:culturalbench_performance},~\ref{tab:blend_performance}, and~\ref{tab:global_mmlu_performance}.
We summarize our results in Figure~\ref{fig:culture} and find that, among models of similar size, Sailor2 has a better understanding of SEA culture, including its cuisine, traditions, geography, and more.
In Figure~\ref{fig:culturalbench_examples}, we present sample responses of the Sailor2-20B model on CulturalBench.

\paragraph{Indonesian Culture Understanding} 
Indonesian culture is uniquely rich and diverse, shaped by various ethnicities, languages, and historical influences. Evaluating this separately helps us assess the model’s ability to capture these distinct cultural nuances. 
We adopts two types of benchmark for evaluation: (1) General Knowledge, Local knowledge \& Reasoning: IndoMMLU~\citep{indommlu}, IndoCareer~\citep{indocareer}; (2) Cultural Reasoning: IndoCulture~\citep{koto2024indoculture}, MAPS~\citep{MAPS}, COPAL-ID~\citep{copalid}, IndoCloze~\citep{IndoCloze}.
As listed in Table~\ref{tab:8_indo_culture_eval}, Sailor2 models present the good performance in understanding Indonesian culture and knowledge.

\vspace{-2mm}
\input{tables/8_indo_culture_evaluation}

%% file: tables/8_sailor_and_sailor2_ppl.tex
\begin{table}[t!]
\centering
\small
\caption{Perplexity Comparison under Continual Pre-training: Qwen1.5 → Sailor and Qwen2.5 → Sailor2. Lower perplexity means better performance. The evaluation dataset for each language is composed of samples from various domains. The valid dataset refers to the evaluation data collected from all languages across these diverse domains.}
\resizebox{\textwidth}{!}{%
\begin{tabular}{llllllll}
    \toprule
    \multirow{2}{*}{\textbf{Model}} & \multicolumn{7}{c}{\textbf{Language}} \\
    \cmidrule{2-8}
     &  \textbf{eng} & \textbf{zho} & \textbf{tha} & \textbf{vie} & \textbf{ind} & \textbf{mya}  & \textbf{valid} \\
    \midrule
    \midrule
     Qwen1.5-14B   & 13.67 & 12.30 & 18.06 & 42.88 & 14.15 & 13.65 & 21.71 \\
     Sailor-14B   & 15.02\pplup{1.35} & 12.47\pplup{0.17} & 6.19\ppldown{11.87} & 11.79\ppldown{31.09} & 7.21\ppldown{6.94} & 11.02\ppldown{2.63} & 9.65\ppldown{12.06} \\
     \cmidrule{2-8}
     Qwen2.5-14B  & 11.40  & 9.91  & 6.36  & 9.60 & 8.46  & 7.07  & 9.10 \\
     Sailor2-20B  & 11.48\pplup{0.08} & 9.27\ppldown{0.64} & 3.35\ppldown{3.01} & 5.24\ppldown{4.36} & 4.93\ppldown{3.53} & 2.41\ppldown{4.66} & 5.69\ppldown{3.41} \\
    \midrule
    \midrule
     Qwen1.5-7B   & 14.55 & 13.18 & 21.35 & 47.09 & 16.14 & 11.99 & 27.07 \\
     Sailor-7B   & 15.75\pplup{1.20} & 13.57\pplup{0.39} & 6.21\ppldown{15.14} & 11.05\ppldown{36.04} & 7.32\ppldown{8.82} & 12.25\pplup{0.26} & 10.15\ppldown{16.92} \\
     \cmidrule{2-8}
     Qwen2.5-7B  & 13.17  & 11.61  & 7.24  & 12.84 & 10.07 & 8.69  & 10.73 \\
     Sailor2-8B  & 13.63\pplup{0.46} & 10.94\ppldown{0.67} & 3.51\ppldown{3.73} & 5.49\ppldown{7.35} & 5.09\ppldown{4.98} & 2.74\ppldown{5.95} & 6.26\ppldown{4.47} \\
    \midrule
     \midrule
     Qwen1.5-0.5B   & 23.78 & 21.48 & 58.8 & 200.65 & 65.44 & 16.07 & 70.41 \\
     Sailor-0.5B   & 25.88\pplup{2.10} & 25.80\pplup{4.32} & 8.72\ppldown{50.08} & 16.76\ppldown{183.89} & 10.98\ppldown{54.46} & 17.69\pplup{1.62} & 17.26\ppldown{53.15} \\
     \cmidrule{2-8}
     Qwen2.5-0.5B  & 22.28  & 21.25  & 13.93  & 26.86 & 25.57  & 13.67  & 22.83 \\
     Sailor2-1B  & 21.01\ppldown{1.27} & 17.28\ppldown{3.97} & 4.52\ppldown{9.41} & 7.52\ppldown{19.34} & 6.75\ppldown{18.82} & 4.11\ppldown{9.56} & 9.36\ppldown{13.47} \\
    \bottomrule
\end{tabular}
}
\label{tab:ppl_shift_sailor1_sailor2}
\end{table}

%% file: tables/8_flores_plus_eng_xx.tex
\begin{table}[ht]
    \centering
    \caption{Performance on Flores Plus for English-Centric Language Pairs (BLEU/chrF++)}
    \label{tab:flores_eng_xx_translation_performance}
    \begin{subtable}[t]{\textwidth}
        \centering
        \caption{English $\rightarrow$ XX}
        \label{tab:eng_to_xx}
        \resizebox{0.8\textwidth}{!}{%
        \begin{tabular}{lcccc}
            \toprule
            \textbf{Language} & \textbf{Sailor2-20B} & \textbf{Llama3.1-70B} & \textbf{Qwen2.5-72B} & \textbf{Qwen2.5-32B} \\
            \midrule
            ace  & \underline{4.8}/\underline{26.1} & \textbf{6.3}/\textbf{30.3} & 4.1/24.8 & 4.6/24.0 \\
            ceb  & \textbf{30.9}/\textbf{57.7} & \underline{28.1}/\underline{54.5} & 18.0/44.4 & 13.6/39.4 \\
            cmn  & 11.3/32.9 & 12.4/32.3 & \underline{12.5}/\textbf{34.3} & \textbf{13.0}/\underline{34.0} \\
            fil  & \textbf{33.8}/\textbf{59.2} & \underline{32.5}/\underline{57.8} & 26.7/53.1 & 23.5/49.5 \\
            ilo  & \textbf{18.0}/\underline{46.0} & \underline{17.5}/\textbf{46.2} & 7.7/31.8 & 5.3/27.1 \\
            ind  & \textbf{43.6}/\textbf{67.7} & \underline{42.8}/\underline{67.3} & 41.9/66.5 & 38.7/64.0 \\
            jav  & \textbf{24.3}/\textbf{52.1} & \underline{22.9}/\underline{51.1} & 11.0/36.8 & 7.6/31.8 \\
            khm  & \textbf{5.4}/\textbf{29.2} & 3.3/21.5 & 3.0/\underline{21.9} & \underline{3.6}/20.6 \\
            lao  & \textbf{7.3}/\textbf{37.6} & 3.7/22.7 & \underline{4.0}/\underline{24.8} & 3.4/21.4 \\
            min  & \underline{12.7}/\underline{39.7} & \textbf{22.2}/\textbf{50.8} & 9.9/35.9 & 9.0/33.7 \\
            mya  & \textbf{3.0}/\textbf{28.1} & 2.0/\underline{21.3} & \underline{2.1}/19.9 & 1.8/17.3 \\
            sun  & \textbf{18.0}/\textbf{47.7} & \underline{16.4}/\underline{45.3} & 10.5/37.2 & 7.9/33.1 \\
            tam  & \textbf{10.4}/\textbf{41.1} & \underline{9.3}/\underline{37.9} & 5.1/31.2 & 4.1/28.3 \\
            tha  & \underline{12.8}/\textbf{48.5} & \textbf{15.2}/47.9 & 12.2/\underline{47.9} & 12.2/46.1 \\
            vie  & \underline{42.6}/60.7 & \textbf{43.0}/\textbf{61.1} & 42.5/\underline{60.8} & 40.6/59.2 \\
            war  & \underline{25.6}/\textbf{53.3} & \textbf{25.6}/\underline{52.8} & 16.4/43.7 & 11.9/38.4 \\
            zsm  & \textbf{41.2}/\textbf{66.8} & \underline{40.2}/\underline{66.2} & 35.5/62.4 & 32.0/59.5 \\
            \bottomrule
        \end{tabular}%
        }
    \end{subtable}
    
    \vspace{1em} 
    
\begin{subtable}[t]{\textwidth}
        \centering
        \caption{XX $\rightarrow$ English}
        \label{tab:xx_to_eng}
        \resizebox{0.8\textwidth}{!}{%
        \begin{tabular}{lcccc}
            \toprule
            \textbf{Language} & \textbf{Sailor2-20B} & \textbf{Llama3.1-70B} & \textbf{Qwen2.5-72B} & \textbf{Qwen2.5-32B} \\
            \midrule
            ace  & \underline{21.9}/\underline{42.9} & \textbf{24.5}/\textbf{45.2} & 19.3/40.2 & 15.8/37.8 \\
            ceb  & \textbf{46.7}/\textbf{67.3} & \underline{44.8}/\underline{65.2} & 39.0/60.1 & 34.0/56.0 \\
            cmn  & 32.4/57.8 & \underline{34.2}/\underline{59.0} & \textbf{35.1}/\textbf{59.9} & 33.6/59.0 \\
            fil  & \underline{49.8}/\underline{69.5} & \textbf{50.5}/\textbf{69.5} & 49.5/69.0 & 45.4/65.8 \\
            ilo  & \textbf{36.0}/\textbf{58.1} & \underline{33.3}/\underline{55.0} & 22.7/44.7 & 21.1/43.1 \\
            ind  & 45.2/67.0 & \textbf{47.1}/\underline{68.5} & \underline{46.9}/\textbf{68.7} & 45.1/67.5 \\
            jav  & \underline{40.0}/\underline{62.0} & \textbf{42.0}/\textbf{63.1} & 33.6/55.6 & 29.0/51.9 \\
            khm  & \textbf{36.2}/\textbf{59.6} & \underline{32.0}/\underline{55.4} & 28.7/52.3 & 25.2/49.8 \\
            lao  & \textbf{37.6}/\textbf{60.6} & 26.3/49.0 & \underline{32.8}/\underline{55.0} & 27.1/50.1 \\
            min  & \underline{35.5}/\underline{57.8} & \textbf{39.6}/\textbf{61.3} & 31.5/53.2 & 28.2/50.7 \\
            mya  & \textbf{27.2}/\textbf{52.0} & \underline{26.7}/\underline{51.0} & 19.6/43.1 & 16.2/39.6 \\
            sun  & \underline{38.1}/\textbf{60.8} & \textbf{38.4}/\underline{60.5} & 32.1/54.8 & 28.6/51.7 \\
            tam  & \underline{32.0}/\underline{56.5} & \textbf{34.3}/\textbf{58.2} & 26.6/50.4 & 24.0/47.9 \\
            tha  & 36.3/60.6 & \underline{37.1}/\underline{61.2} & \textbf{38.7}/\textbf{62.3} & 35.8/60.6 \\
            vie  & 39.1/62.0 & \underline{40.7}/\underline{63.0} & \textbf{41.8}/\textbf{63.9} & 39.4/62.5 \\
            war  & \textbf{47.5}/\textbf{67.7} & \underline{46.6}/\underline{66.4} & 38.5/59.6 & 35.1/56.8 \\
            zsm  & 46.6/68.1 & \textbf{48.0}/\textbf{69.4} & \underline{47.2}/\underline{68.8} & 45.1/67.2 \\
            \bottomrule
        \end{tabular}%
        }
    \end{subtable}
\end{table}

%% file: tables/8_flores_chrf_sailor2_20b.tex
\begin{table}[ht]
\centering
\caption{Performance on Flores Plus (chrF++) for Sailor2-20B}
\label{tab:matrix-chrf-sailor2_20b}
\resizebox{\textwidth}{!}{%
\begin{tabular}{l *{16}{c}}
\toprule
\makecell[l]{\textbf{Target} ($\rightarrow$) \\ \textbf{Source} ($\downarrow$)}
 & \texttt{eng} & \texttt{cmn} & \texttt{ind} & \texttt{tha} & \texttt{vie} & \texttt{zsm} & \texttt{mya} & \texttt{lao} & \texttt{ceb} & \texttt{ilo} & \texttt{jav} & \texttt{khm} & \texttt{sun} & \texttt{fil} & \texttt{war} & \texttt{tam} \\
\midrule
English (\texttt{eng}) & - & 32.9 & 67.7 & 48.5 & 60.7 & 66.8 & 28.1 & 37.6 & 57.7 & 46.0 & 52.1 & 29.2 & 47.7 & 59.2 & 53.3 & 41.1 \\ 
Chinese (\texttt{cmn}) & 57.8 & - & 53.6 & 41.8 & 52.7 & 53.0 & 24.6 & 30.8 & 47.7 & 39.0 & 42.6 & 25.1 & 39.4 & 48.1 & 42.8 & 34.5 \\ 
Indonesian (\texttt{ind}) & 67.0 & 28.1 & - & 45.4 & 55.8 & 59.3 & 26.1 & 34.5 & 51.3 & 42.3 & 49.2 & 26.6 & 46.0 & 54.1 & 46.6 & 37.1 \\ 
Thai (\texttt{tha}) & 60.6 & 27.5 & 56.7 & - & 53.5 & 56.0 & 25.5 & 33.9 & 49.0 & 41.0 & 44.7 & 26.1 & 41.3 & 50.8 & 44.4 & 35.4 \\ 
Vietnamese (\texttt{vie}) & 62.0 & 28.1 & 57.5 & 44.4 & - & 56.4 & 25.6 & 33.5 & 49.6 & 41.3 & 45.3 & 26.1 & 41.6 & 51.4 & 45.1 & 36.1 \\ 
Malay (\texttt{zsm}) & 68.1 & 28.7 & 60.4 & 44.9 & 55.6 & - & 26.4 & 35.3 & 51.5 & 42.5 & 49.4 & 26.6 & 44.6 & 54.1 & 47.3 & 37.4 \\ 
Burmese (\texttt{mya}) & 52.0 & 22.9 & 49.3 & 38.3 & 46.9 & 49.4 & - & 29.8 & 44.8 & 37.0 & 39.9 & 23.2 & 36.7 & 46.5 & 41.9 & 33.3 \\ 
Lao (\texttt{lao}) & 60.6 & 25.8 & 55.9 & 45.0 & 52.3 & 55.6 & 25.4 & - & 48.8 & 40.0 & 45.0 & 27.2 & 41.2 & 51.1 & 44.1 & 35.4 \\ 
Cebuano (\texttt{ceb}) & 67.3 & 28.6 & 57.7 & 43.6 & 53.5 & 57.3 & 26.5 & 33.3 & - & 44.5 & 45.8 & 26.4 & 41.5 & 55.5 & 51.7 & 36.9 \\ 
Ilocano (\texttt{ilo}) & 58.1 & 24.7 & 51.7 & 38.8 & 48.0 & 51.6 & 24.4 & 29.6 & 49.5 & - & 37.9 & 24.0 & 36.6 & 51.4 & 45.2 & 33.5 \\ 
Javanese (\texttt{jav}) & 62.0 & 26.4 & 58.7 & 41.6 & 51.3 & 56.4 & 25.2 & 32.5 & 48.3 & 38.2 & - & 25.9 & 42.4 & 51.8 & 42.6 & 35.1 \\ 
Khmer (\texttt{khm}) & 59.6 & 26.4 & 55.5 & 42.9 & 51.5 & 54.8 & 24.8 & 33.5 & 48.8 & 40.1 & 44.8 & - & 41.5 & 50.5 & 44.2 & 34.5 \\ 
Sundanese (\texttt{sun}) & 60.8 & 26.0 & 59.0 & 41.9 & 51.5 & 56.3 & 25.4 & 32.3 & 47.5 & 33.3 & 46.8 & 26.3 & - & 51.1 & 38.3 & 35.1 \\ 
Tagalog (\texttt{fil}) & 69.5 & 29.2 & 60.1 & 44.5 & 55.0 & 59.5 & 26.4 & 33.3 & 55.3 & 46.2 & 47.3 & 26.6 & 42.4 & - & 50.3 & 37.5 \\ 
Waray (\texttt{war}) & 67.7 & 27.2 & 57.3 & 43.6 & 53.0 & 57.4 & 26.3 & 33.2 & 56.8 & 45.8 & 44.4 & 25.5 & 39.9 & 55.9 & - & 36.8 \\ 
Tamil (\texttt{tam}) & 56.5 & 24.2 & 51.3 & 39.2 & 48.4 & 51.5 & 24.6 & 28.8 & 46.7 & 38.5 & 41.5 & 23.7 & 38.0 & 47.6 & 42.9 & - \\ 
\bottomrule
\end{tabular}
}
\end{table}

%% file: tables/8_indo_culture_evaluation.tex
\begin{table}[htp]\centering
\caption{Evaluation Results on Indonesian Culture (3-shot, Accuracy).}
\label{tab:8_indo_culture_eval}
\scriptsize
\begin{tabular}{lcccccccc}\toprule
Model &IndoMMLU &IndoCareer &IndoCulture &MAPS &COPAL-ID &IndoCloze &Avg \\
\midrule
\midrule
Sailor2-20B &70.7 &69.0 &79.3 &92.9 &86.9 &98.1 &82.8 \\
Sailor2-8B &64.5 &60.9 &74.7 &90.8 &86.5 &96.5 &78.9 \\
\midrule
Aya-Expanse-32b &59.3 &61.5 &71.4 &91.4 &84.6 &97.0 &77.5 \\
Gemma-2-27b &57.1 &61.6 &70.2 &92.0 &81.8 &96.5 &76.5 \\
Qwen2.5-32B &61.9 &64.3 &66.8 &89.9 &76.6 &99.6 &76.5 \\
SEA-LIONv3-9B &60.0 &58.9 &65.8 &90.2 &80.0 &96.7 &75.2 \\
Gemma-2-9B &57.9 &57.8 &67.9 &89.9 &79.3 &95.8 &74.7 \\
Qwen2.5-7B &53.1 &43.0 &61.1 &87.6 &73.0 &94.2 &68.6 \\
Qwen2.5-14B &52.6 &64.1 &50.7 &76.0 &62.8 &99.8 &67.6 \\
Llama-3.1-8B &47.5 &61.5 &54.5 &77.8 &70.0 &92.5 &67.3 \\
SeaLLMs-v3-7B &47.6 &32.8 &59.9 &87.9 &71.2 &95.9 &65.8 \\
Llama-3-8B &48.8 &39.6 &55.0 &83.1 &66.8 &87.2 &63.4 \\
\bottomrule
\end{tabular}
\end{table}

%% file: sections/9.conclusion.tex
\section{Conclusion and Future Work}

This report introduces the Sailor2 family of open models, designed to facilitate the development of large language models for Southeast Asian languages. 
We also summarize the key insights from our pipeline for building the Sailor2 model, covering data curation, continual pre-training, post-training, evaluation and advanced model customization. 
We hope this report will inspire the community to develop more inclusive and robust multilingual language models for underserved languages.

Looking ahead, we plan to expand our multilingual research to include a broader range of low-resource languages and explore more efficient model training approaches. 
The following sections detail our motivations and review the most relevant works in \textbf{data curation}, \textbf{model design}, and \textbf{model training}.

\subsection{Synthetic Data Curation for Supporting Low-resource Languages}
\label{sec:future_work_data_curation}

Apart from a few high-resource languages, most languages have a relatively scarce supply of training tokens. For example, in Sailor2, six languages contain fewer than 1 billion training tokens (See Table~\ref{tab:disk_size_training_token} for detailed statistics). 
For extremely low-resource languages like Minangkabau (spoken in Indonesia by approximately 6.5 million people) and Acehnese (spoken in Indonesia by around 3.5 million people), we were only able to mine fewer than one million tokens\footnote{See full list of undetermined (und) data in \url{https://huggingface.co/datasets/HuggingFaceFW/fineweb-2} for more low-resource languages.}. 

One effective way to address this issue is to leverage translated synthetic data. For example, \citet{Wang2024MultilingualPU} translate high-quality documents (e.g., Fineweb-Edu~\citep{lozhkov2024fineweb-edu}) from English into medium-level languages such as French, German, and Spanish. Similarly, \citet{doshi-etal-2024-pretraining} adopt the Translationese dataset to extend coverage to additional low-resource languages, including Hindi, Gujarati, and Marathi.

\subsection{Tokenizer-Free Model for Open-Vocabulary Learning}
\label{sec:future_work_tokenizer_free_model}

Recent studies demonstrate that tokenizer-free language models can effectively process unseen languages and exhibit greater robustness against noise attacks compared to tokenizer-based models. 

One approach involves pixel-based language models~\citep{lotz-etal-2023-text, rust-etal-2023-pixel}, which treat text as images. This enables them to learn any script and achieve open-vocabulary language learning by exploiting visual similarities among characters and scripts through parameter sharing. In contrast, byte-level language models~\citep{evabyte, kallini2024mrt5, xue-etal-2022-byt5} bypass the tokenization step entirely by directly processing the raw character or byte stream as input.

These approaches offer significant benefits for both morphologically rich languages and languages that mix multiple scripts. 
For example, languages such as Turkish, Finnish, and Hungarian are known for their complex morphological structures, while Japanese (which combines Kanji, Hiragana, and Katakana) and Hindi (which often integrates Devanagari and Latin scripts) frequently mix scripts in everyday usage.

\subsection{Efficient Continual Pre-training for Multilingual Model}
\label{sec:future_work_efficient_training}

Continual pre-training is more efficient and cost-effective than training from scratch when building new multilingual language models for target languages. 
By leveraging an existing base model, its inherent capabilities are preserved while saving computational investment. 
For instance, developers can use Sailor2 as a foundation to build more powerful models for Southeast Asian languages with their in-house data. 
Moreover, both the developing infrastructure (e.g., Sailor2 open source every details) and the model tokenizer (the most critical component for multilingual as verified by \cite{tao2024scaling}) are mature.

However, our concern is that many existing open models might be over-trained, leaving little room for further fine-tuning. 
For example, Llama3.1 was trained on 15T tokens and Qwen2.5 on 18T tokens. Although Sailor2 employs model expansion to mitigate this issue, its approach remains inefficient due to the increased computational cost associated with a larger model size.

To achieve more efficient continual pre-training, we propose exploring the following directions:
(1) Employing an over-training indicator~\citep{ouyang2024low} to guide the selection of an appropriate base model.
(2) Enhancing model plasticity~\citep{Chen2023ImprovingLP} to enable the model to absorb additional multilingual knowledge.
(3) Leveraging insights from the lottery ticket hypothesis to update only the most essential parameters~\citep{Yuan2024KSLotteryFC}.

%% file: appendices/contribution.tex
\section*{Contributors and Acknowledgements}

\paragraph{Core Contributors}
Longxu Dou, Qian Liu, Fan Zhou, Changyu Chen, Zili Wang, Ziqi Jin, Zichen Liu, Tongyao Zhu, Cunxiao Du, Penghui Yang, Haonan Wang, Jiaheng Liu, Yongchi Zhao, Xiachong Feng, Xin Mao, Man Tsung Yeung

\paragraph{Contributors}
Kunat Pipatanakul, Fajri Koto, Min Si Thu, Hynek Kydlíček, Zeyi Liu, 
Qunshu Lin, Sittipong Sripaisarnmongkol, Kridtaphad Sae-Khow, 
Nirattisai Thongchim, Taechawat Konkaew, Narong Borijindargoon, Anh Dao, 
Matichon Maneegard, Phakphum Artkaew, Zheng-Xin Yong, Quan Nguyen, 
Wannaphong Phatthiyaphaibun, Hoang H. Tran, Mike Zhang, Shiqi Chen

\paragraph{Advisors}
Tianyu Pang, Chao Du, Xinyi Wan, Wei Lu, Min Lin

\paragraph{Acknowledgements}

\begin{itemize}
    \item Model Training: Qwen-Team, Megatron\-LLM
    \item Model Evaluation: Finetasks, SeaCrowd
    \item Model Deployment: AK(@\_akhaliq), Ollama
    \item Sea SRE Team: Zhikai Huang
\end{itemize}

%% file: appendices/prompt.tex
\begin{tcolorbox}[title={The Prompt used for Translation}]
You are a highly skilled translator tasked with translating various types of content from English into \{\{ language \}\}. Follow these instructions carefully to complete the translation task.

You will receive a user-bot conversation in XML format. Please follow a three-step translation process:

\begin{enumerate}
  \item \textbf{Initial Translation:} Translate the input content into \{\{ language \}\}, preserving the original intent and keeping the original paragraph and text format unchanged. Do not delete or omit any content, and ensure that all original Markdown elements (e.g., images, code blocks) are preserved.
  \item \textbf{Reflection and Feedback:} Carefully review both the source text and your translation. Provide constructive criticism and specific suggestions to improve the translation in terms of:
    \begin{enumerate}[label=(\roman*)]
      \item \textbf{Accuracy:} Correct errors of addition, mistranslation, omission, or untranslated text.
      \item \textbf{Fluency:} Apply \{\{ language \}\} grammar, spelling, and punctuation rules while avoiding unnecessary repetitions.
      \item \textbf{Style:} Ensure that the translation reflects the style of the source text and considers any relevant cultural context.
    \end{enumerate}
  \item \textbf{Refinement:} Based on your reflections, refine and polish your translation.
  \item \textbf{Fallback:} If you are not confident in translating the conversation, please return ``\texttt{<stop></stop>}''.
\end{enumerate}

\bigskip
\textbf{Output:}

For each step of the translation process, output your results within the appropriate XML tags as follows:
\begin{verbatim}
<step1_initial_translation>
[Insert your initial translation here]
</step1_initial_translation>

<step2_reflection>
[Insert your reflection on the translation, including a list 
of specific, helpful, and constructive suggestions for 
improvement. Each suggestion should address a specific 
part of the translation.]
</step2_reflection>

<step3_refined_translation>
[Insert your refined and polished translation here]
</step3_refined_translation>
\end{verbatim}

Ensure that your final translation in step 3 accurately reflects the original meaning while sounding natural in \{\{ language \}\}.

Here is the original conversation:
\label{box:trans_prompt}
\end{tcolorbox}

%% file: appendices/ppl_shift_figure.tex

\input{tables/5_1_post_training_data_distribution}

\begin{figure*}
\centering
\begin{minipage}[b]{\textwidth}
\centering
\includegraphics[width=0.95\textwidth]{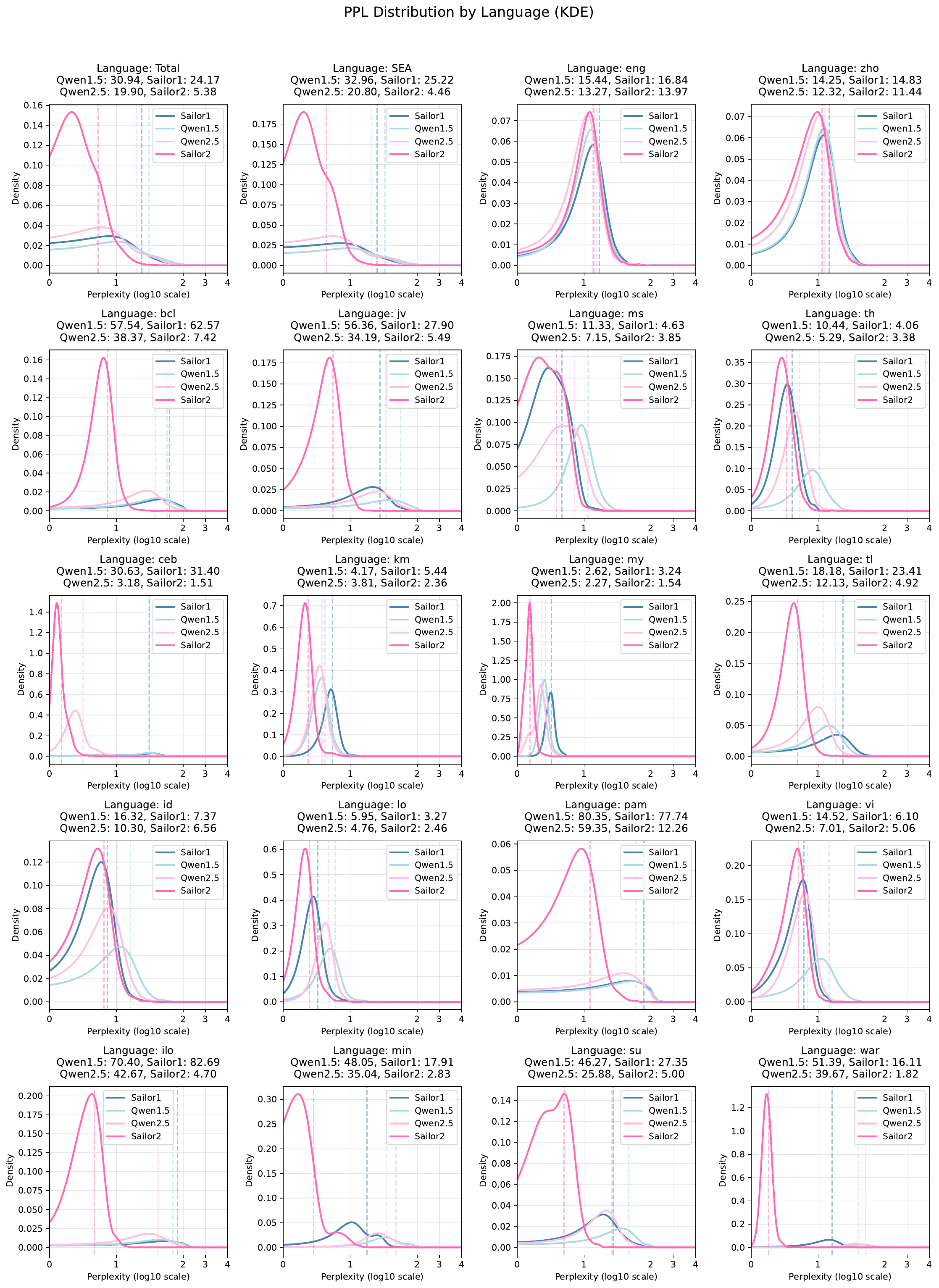}
\end{minipage}
\captionof{figure}{Comparison of PPL distribution smoothed with Kernel Density Estimation (KDE). We compare Sailor2-8B, Qwen2.5-7B, Sailor1-7b and Qwen1.5-7B. Our results demonstrate that with extra 1B parameters, Sailor2-8B can preserve its English and Chinese capability, while achieving in much lower PPL in SEA languages.}
\label{fig:ppl_shift}
\end{figure*}

\begin{figure*}
\centering
\begin{minipage}[b]{\textwidth}
\centering
\includegraphics[width=1\textwidth]{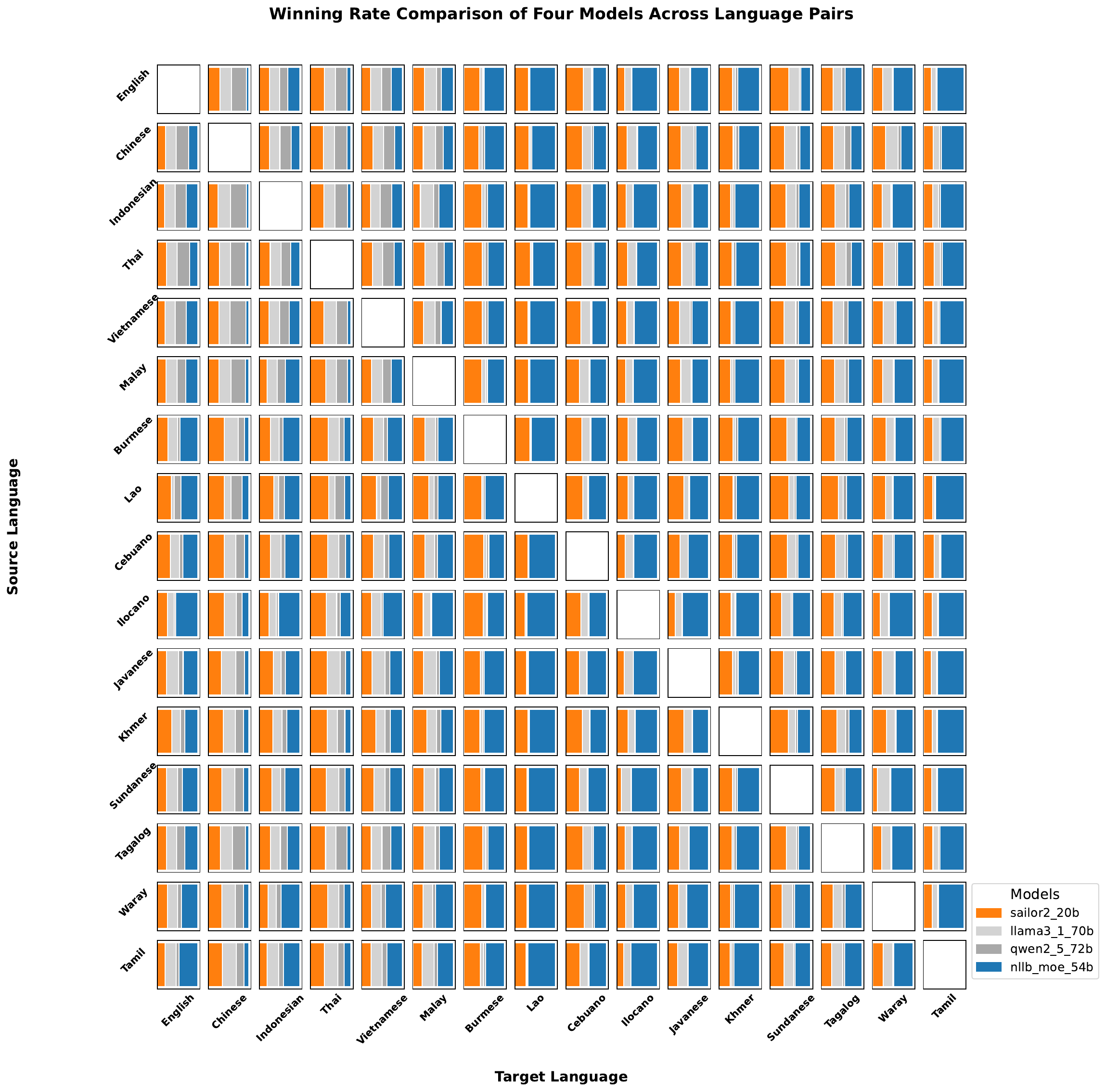}
\end{minipage}
\captionof{figure}{Comparison of win rate of four models based on their ChrF++ scores.
The shaded area in each cube represents the top-1 accuracy of each model across different translation directions in the Flores Plus Translation Dataset.
We observed that Sailor2 performs on par with, or even outperforms NLLB, a model optimized for translation tasks that excels at translating low-resource languages.
}
\label{fig:flores_plus_WR}
\end{figure*}

\begin{figure*}
\centering
\begin{minipage}[b]{\textwidth}
\centering
\includegraphics[width=0.95\textwidth]{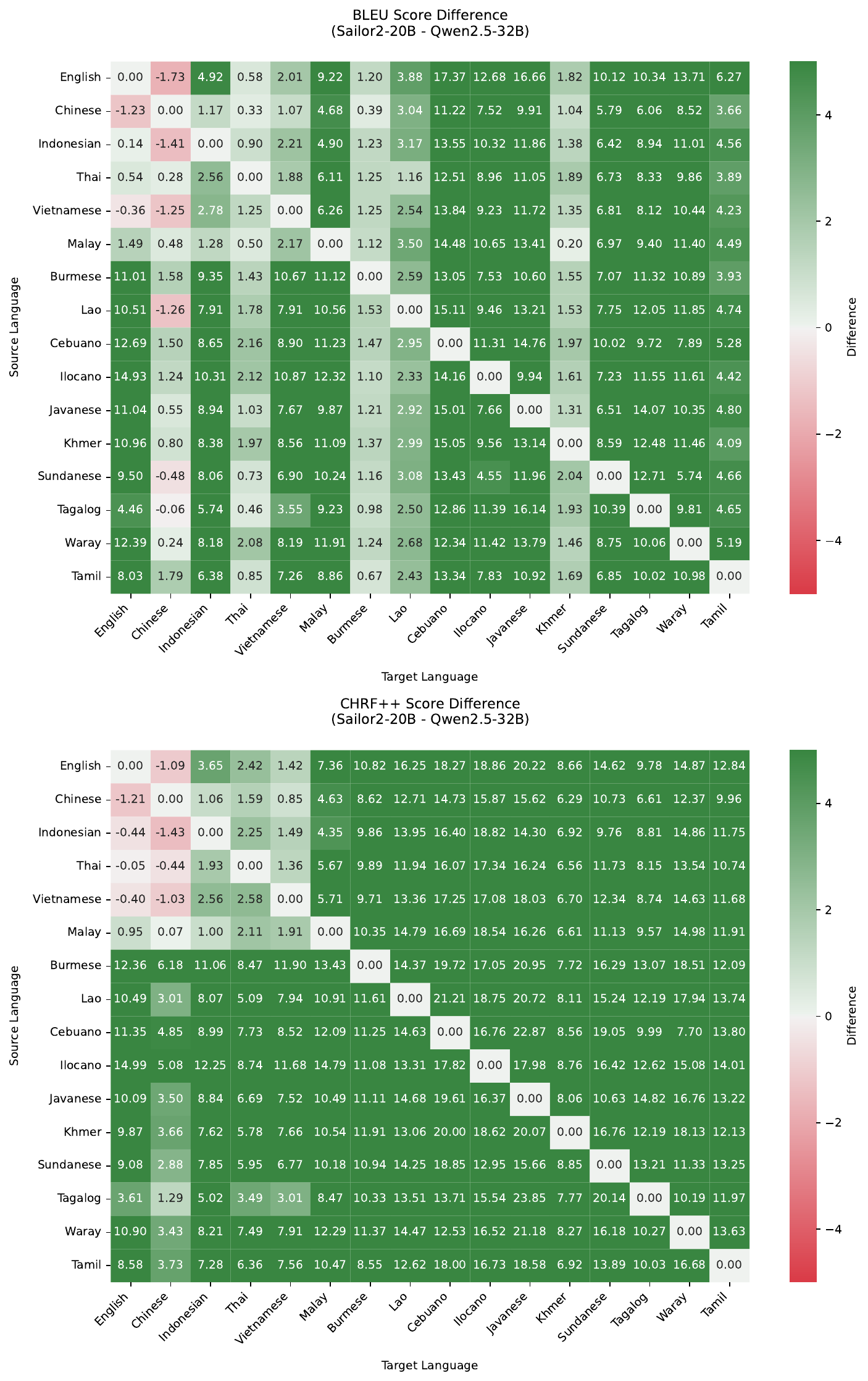}
\end{minipage}
\captionof{figure}{Comparison of BLEU and ChrF++ scores on the Flores Plus Translation Dataset across various source-target language pairs between Sailor2 20B and Qwen2.5-32B. BLEU Score Difference = Sailor2 BLEU - Qwen2.5 BLEU.}
\label{fig:flores_plus__qwen_32b}
\end{figure*}

\begin{figure*}
\centering
\begin{minipage}[b]{\textwidth}
\centering
\includegraphics[width=0.95\textwidth]{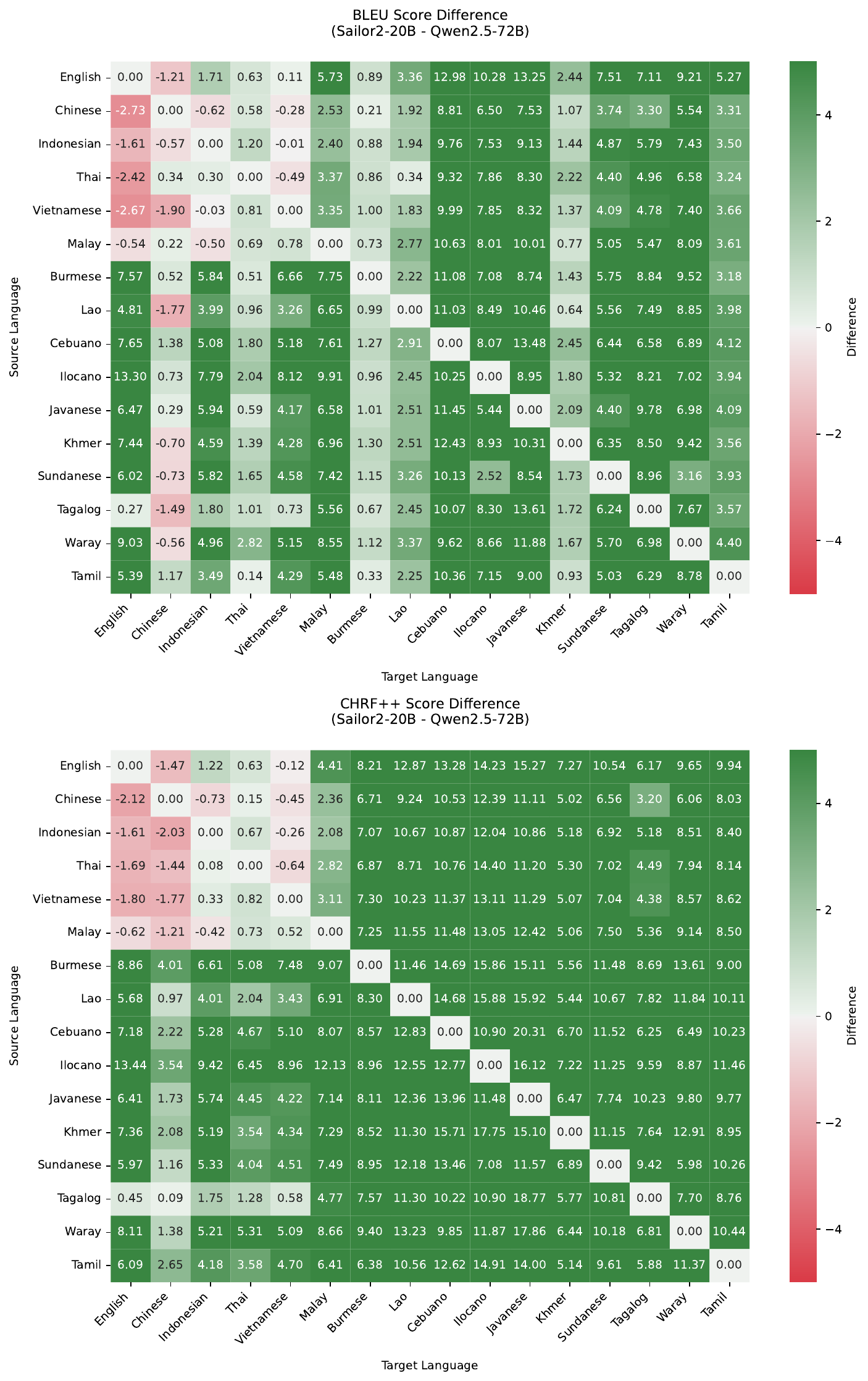}
\end{minipage}
\captionof{figure}{Comparison of BLEU and ChrF++ scores on the Flores Plus Translation Dataset across various source-target language pairs between Sailor2 20B and Qwen2.5-72B. BLEU Score Difference = Sailor2 BLEU - Qwen2.5 BLEU.}
\label{fig:flores_plus__qwen_72b}
\end{figure*}

\begin{figure*}
\centering
\begin{minipage}[b]{\textwidth}
\centering
\includegraphics[width=0.95\textwidth]{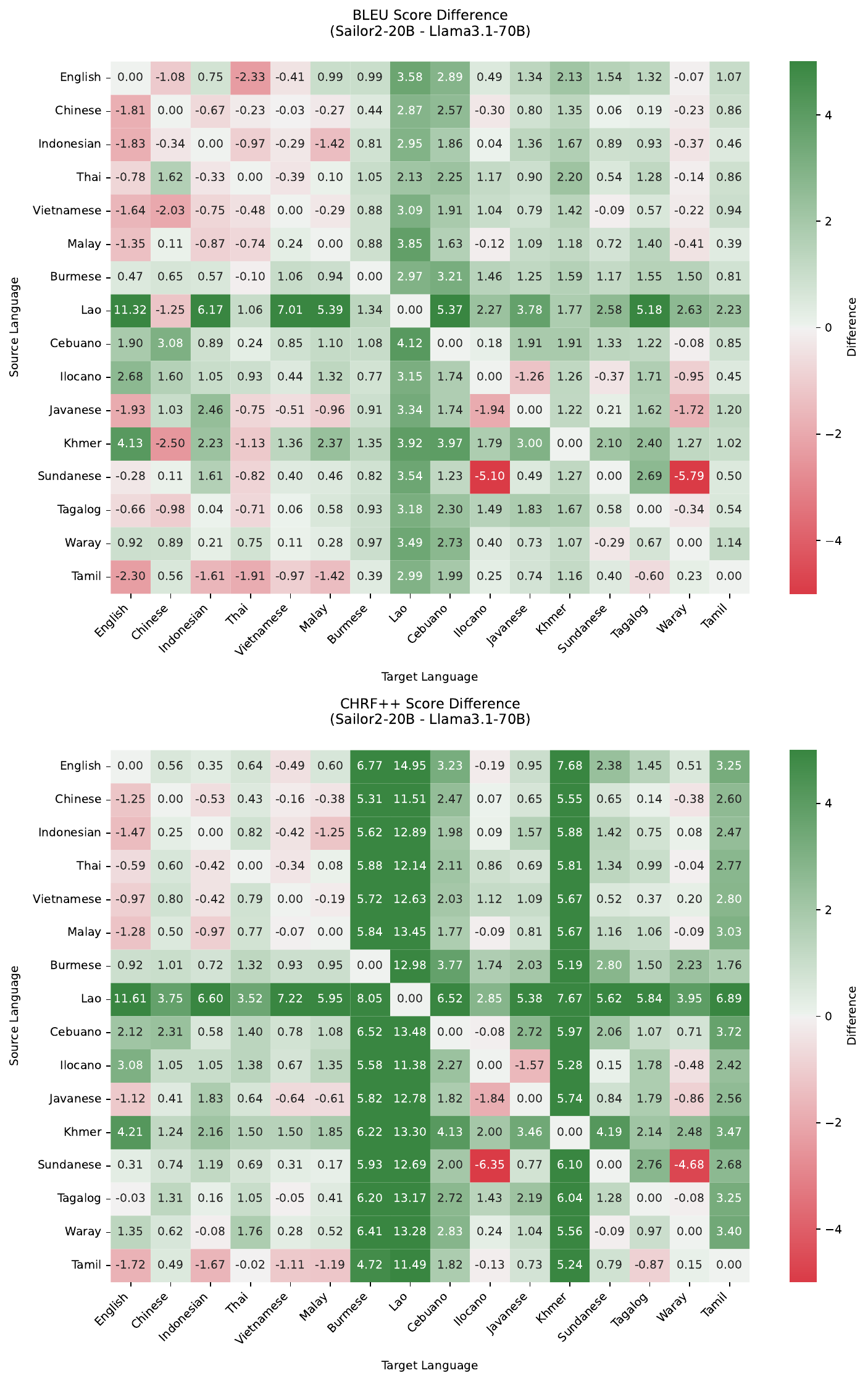}
\end{minipage}
\captionof{figure}{Comparison of BLEU and ChrF++ scores on the Flores Plus Translation Dataset across various source-target language pairs between Sailor2 20B and Llama3.1-70B. BLEU Score Difference = Sailor2 BLEU - Llama3.1 BLEU.}
\label{fig:flores_plus__llama_70b}
\end{figure*}

\begin{figure*}
\centering
\begin{minipage}[b]{\textwidth}
\centering
\includegraphics[width=0.95\textwidth]{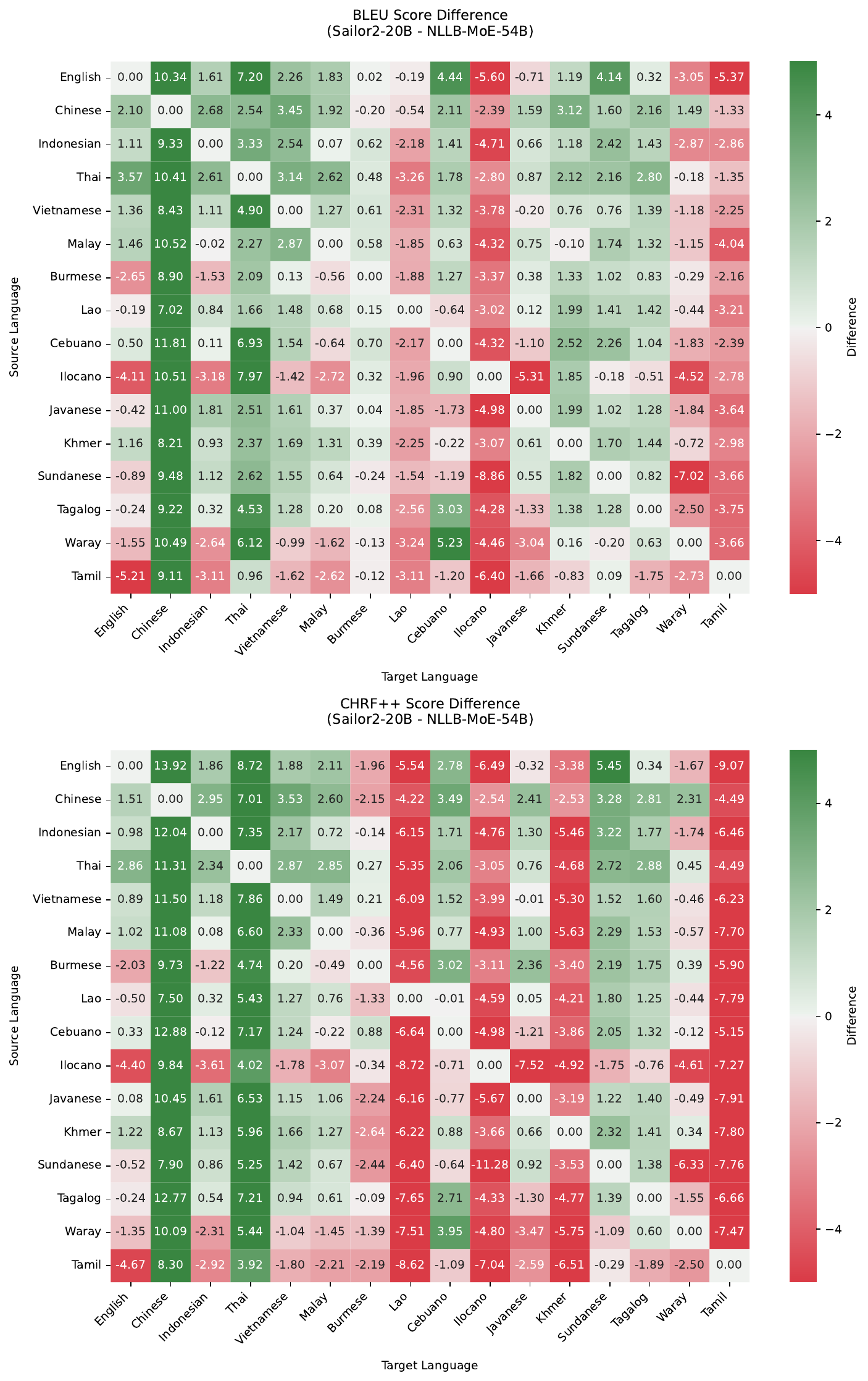}
\end{minipage}
\captionof{figure}{Comparison of BLEU and ChrF++ scores on the Flores Plus Translation Dataset across various source-target language pairs between Sailor2 20B and NLLB-MoE-54B. BLEU Score Difference = Sailor2 BLEU - NLLB BLEU. We noticed NLLB failed to generate complete long Chinese sentences: \url{https://github.com/facebookresearch/fairseq/issues/5549}, and we also found many common Chinese characters and punctuations are tokenized to $<$unk$>$.}
\label{fig:flores_plus__nllb_moe_54b}
\end{figure*}

\input{tables/8_flores_chrf_qwen2_5_32b}
\input{tables/8_flores_chrf_qwen2_5_72b}
\input{tables/8_flores_chrf_llama3_1_70b}
\input{tables/8_flores_chrf_nllb_moe_54b}

%% file: tables/5_1_post_training_data_distribution.tex
\begin{table}[htbp]
\centering
\caption{Data Distribution of SEA-UltraChat (Instruction Tuning) and SEA-UltraFeedback (Preference Tuning) across Languages. SEA-UltraChat has been used by two-stage instruction tuning.}
\begin{tabular}{lcccc}
\toprule
\textbf{Language} & \textbf{SFT Stage 1} & \textbf{SFT Stage 2} & \textbf{SEA-UltraChat} & \textbf{SEA-UltraFeedback} \\
\midrule
English    & 1611404 & 72000  & 1683404 & 8981  \\
Chinese    & 154908  & 72000  & 226908  & 8845  \\
Indonesian & 287719  & 48000  & 335719  & 4687  \\
Thai       & 334194  & 48000  & 382194  & 4696  \\
Vietnamese & 301226  & 48000  & 349226  & 4710  \\
Malay      & 212827  & 48000  & 260827  & 4714  \\
Burmese    & 204495  & 48000  & 252495  & 4614  \\
Lao        & 134228  & 48000  & 182228  & 4393  \\
Cebuano    & 194986  & 1200   & 196186  & 1178  \\
Ilocano    & 1867    & 1200   & 3067    & 1172  \\
Javanese   & 134499  & 48000  & 182499  & 1166  \\
Khmer      & 64421   & 4800   & 69221   & 1151  \\
Sundanese  & 181486  & 1200   & 182686  & 1174  \\
Tagalog    & 217045  & 48000  & 265045  & 1173  \\
Waray      & 177921  & 1200   & 179121  & 0     \\
Tamil      & 82015   & 1200   & 83215   & 1159  \\
\midrule
\textbf{Total} & 4295241 & 538800 & 4834041 & 53813 \\
\bottomrule
\end{tabular}
\label{table:sft_data_distribution}
\end{table}

%% file: tables/8_flores_chrf_qwen2_5_32b.tex
\begin{table}[ht]
\centering
\caption{Performance on Flores Plus (chrF++) for Qwen2.5-32B}
\label{tab:matrix-chrf-qwen2_5_32b}
\resizebox{\textwidth}{!}{%
\begin{tabular}{l *{16}{c}}
\toprule
\makecell[l]{\textbf{Target} ($\rightarrow$) \\ \textbf{Source} ($\downarrow$)}
 & \texttt{eng} & \texttt{cmn} & \texttt{ind} & \texttt{tha} & \texttt{vie} & \texttt{zsm} & \texttt{mya} & \texttt{lao} & \texttt{ceb} & \texttt{ilo} & \texttt{jav} & \texttt{khm} & \texttt{sun} & \texttt{fil} & \texttt{war} & \texttt{tam} \\
\midrule
English (\texttt{eng}) & - & 34.0 & 64.0 & 46.1 & 59.2 & 59.5 & 17.3 & 21.4 & 39.4 & 27.1 & 31.8 & 20.6 & 33.1 & 49.5 & 38.4 & 28.3 \\ 
Chinese (\texttt{cmn}) & 59.0 & - & 52.6 & 40.2 & 51.8 & 48.3 & 16.0 & 18.0 & 33.0 & 23.1 & 27.0 & 18.8 & 28.7 & 41.5 & 30.4 & 24.6 \\ 
Indonesian (\texttt{ind}) & 67.5 & 29.5 & - & 43.1 & 54.3 & 55.0 & 16.2 & 20.6 & 34.9 & 23.5 & 34.9 & 19.6 & 36.2 & 45.2 & 31.7 & 25.4 \\ 
Thai (\texttt{tha}) & 60.6 & 27.9 & 54.8 & - & 52.1 & 50.3 & 15.6 & 22.0 & 33.0 & 23.7 & 28.5 & 19.5 & 29.6 & 42.6 & 30.9 & 24.7 \\ 
Vietnamese (\texttt{vie}) & 62.5 & 29.1 & 55.0 & 41.9 & - & 50.7 & 15.9 & 20.1 & 32.4 & 24.2 & 27.2 & 19.4 & 29.3 & 42.7 & 30.5 & 24.4 \\ 
Malay (\texttt{zsm}) & 67.2 & 28.6 & 59.4 & 42.8 & 53.6 & - & 16.0 & 20.5 & 34.8 & 24.0 & 33.1 & 20.0 & 33.5 & 44.5 & 32.3 & 25.5 \\ 
Burmese (\texttt{mya}) & 39.6 & 16.7 & 38.2 & 29.8 & 35.0 & 36.0 & - & 15.4 & 25.1 & 20.0 & 18.9 & 15.5 & 20.4 & 33.4 & 23.4 & 21.2 \\ 
Lao (\texttt{lao}) & 50.1 & 22.8 & 47.8 & 39.9 & 44.4 & 44.6 & 13.8 & - & 27.6 & 21.2 & 24.3 & 19.1 & 25.9 & 38.9 & 26.1 & 21.6 \\ 
Cebuano (\texttt{ceb}) & 56.0 & 23.7 & 48.8 & 35.9 & 45.0 & 45.2 & 15.2 & 18.6 & - & 27.7 & 22.9 & 17.8 & 22.4 & 45.5 & 44.0 & 23.1 \\ 
Ilocano (\texttt{ilo}) & 43.1 & 19.6 & 39.5 & 30.0 & 36.3 & 36.8 & 13.3 & 16.3 & 31.7 & - & 19.9 & 15.2 & 20.2 & 38.8 & 30.2 & 19.4 \\ 
Javanese (\texttt{jav}) & 51.9 & 22.9 & 49.9 & 35.0 & 43.8 & 45.9 & 14.1 & 17.9 & 28.7 & 21.8 & - & 17.9 & 31.7 & 37.0 & 25.9 & 21.9 \\ 
Khmer (\texttt{khm}) & 49.8 & 22.8 & 47.8 & 37.1 & 43.8 & 44.3 & 12.9 & 20.4 & 28.8 & 21.5 & 24.7 & - & 24.7 & 38.3 & 26.1 & 22.4 \\ 
Sundanese (\texttt{sun}) & 51.7 & 23.1 & 51.2 & 35.9 & 44.7 & 46.1 & 14.5 & 18.1 & 28.6 & 20.4 & 31.1 & 17.4 & - & 37.9 & 27.0 & 21.9 \\ 
Tagalog (\texttt{fil}) & 65.8 & 27.9 & 55.1 & 41.0 & 52.0 & 51.0 & 16.1 & 19.8 & 41.6 & 30.7 & 23.4 & 18.8 & 22.2 & - & 40.1 & 25.5 \\ 
Waray (\texttt{war}) & 56.8 & 23.8 & 49.1 & 36.1 & 45.1 & 45.1 & 14.9 & 18.7 & 44.2 & 29.3 & 23.2 & 17.2 & 23.7 & 45.6 & - & 23.2 \\ 
Tamil (\texttt{tam}) & 47.9 & 20.5 & 44.0 & 32.9 & 40.9 & 41.0 & 16.1 & 16.2 & 28.7 & 21.8 & 22.9 & 16.8 & 24.1 & 37.5 & 26.2 & - \\ 
\bottomrule
\end{tabular}
}
\end{table}

%% file: tables/8_flores_chrf_qwen2_5_72b.tex
\begin{table}[ht]
\centering
\caption{Performance on Flores Plus (chrF++) for Qwen2.5-72B}
\label{tab:matrix-chrf-qwen2_5_72b}
\resizebox{\textwidth}{!}{%
\begin{tabular}{l *{16}{c}}
\toprule
\makecell[l]{\textbf{Target} ($\rightarrow$) \\ \textbf{Source} ($\downarrow$)}
 & \texttt{eng} & \texttt{cmn} & \texttt{ind} & \texttt{tha} & \texttt{vie} & \texttt{zsm} & \texttt{mya} & \texttt{lao} & \texttt{ceb} & \texttt{ilo} & \texttt{jav} & \texttt{khm} & \texttt{sun} & \texttt{fil} & \texttt{war} & \texttt{tam} \\
\midrule
English (\texttt{eng}) & - & 34.3 & 66.5 & 47.9 & 60.8 & 62.4 & 19.9 & 24.8 & 44.4 & 31.8 & 36.8 & 21.9 & 37.2 & 53.1 & 43.7 & 31.2 \\ 
Chinese (\texttt{cmn}) & 59.9 & - & 54.4 & 41.7 & 53.1 & 50.6 & 17.9 & 21.5 & 37.1 & 26.6 & 31.5 & 20.1 & 32.9 & 44.9 & 36.8 & 26.5 \\ 
Indonesian (\texttt{ind}) & 68.7 & 30.1 & - & 44.7 & 56.1 & 57.3 & 19.0 & 23.9 & 40.4 & 30.2 & 38.3 & 21.4 & 39.1 & 48.9 & 38.1 & 28.7 \\ 
Thai (\texttt{tha}) & 62.3 & 28.9 & 56.6 & - & 54.1 & 53.1 & 18.6 & 25.2 & 38.3 & 26.6 & 33.5 & 20.8 & 34.3 & 46.3 & 36.5 & 27.3 \\ 
Vietnamese (\texttt{vie}) & 63.9 & 29.8 & 57.2 & 43.6 & - & 53.3 & 18.3 & 23.3 & 38.3 & 28.2 & 34.0 & 21.1 & 34.6 & 47.1 & 36.5 & 27.4 \\ 
Malay (\texttt{zsm}) & 68.8 & 29.9 & 60.9 & 44.1 & 55.0 & - & 19.1 & 23.8 & 40.0 & 29.5 & 36.9 & 21.6 & 37.1 & 48.7 & 38.2 & 28.9 \\ 
Burmese (\texttt{mya}) & 43.1 & 18.9 & 42.6 & 33.2 & 39.4 & 40.4 & - & 18.4 & 30.1 & 21.2 & 24.8 & 17.7 & 25.2 & 37.8 & 28.3 & 24.3 \\ 
Lao (\texttt{lao}) & 55.0 & 24.8 & 51.9 & 42.9 & 48.9 & 48.6 & 17.1 & - & 34.1 & 24.1 & 29.1 & 21.7 & 30.5 & 43.3 & 32.2 & 25.3 \\ 
Cebuano (\texttt{ceb}) & 60.1 & 26.3 & 52.5 & 39.0 & 48.4 & 49.3 & 17.9 & 20.4 & - & 33.6 & 25.4 & 19.6 & 30.0 & 49.2 & 45.2 & 26.6 \\ 
Ilocano (\texttt{ilo}) & 44.7 & 21.2 & 42.3 & 32.3 & 39.0 & 39.5 & 15.4 & 17.1 & 36.7 & - & 21.8 & 16.8 & 25.4 & 41.8 & 36.4 & 22.0 \\ 
Javanese (\texttt{jav}) & 55.6 & 24.6 & 53.0 & 37.2 & 47.1 & 49.2 & 17.1 & 20.2 & 34.4 & 26.7 & - & 19.5 & 34.6 & 41.6 & 32.8 & 25.3 \\ 
Khmer (\texttt{khm}) & 52.3 & 24.4 & 50.3 & 39.4 & 47.1 & 47.5 & 16.3 & 22.2 & 33.1 & 22.4 & 29.7 & - & 30.3 & 42.8 & 31.3 & 25.6 \\ 
Sundanese (\texttt{sun}) & 54.8 & 24.8 & 53.7 & 37.8 & 47.0 & 48.8 & 16.5 & 20.1 & 34.0 & 26.3 & 35.2 & 19.4 & - & 41.6 & 32.3 & 24.9 \\ 
Tagalog (\texttt{fil}) & 69.0 & 29.1 & 58.4 & 43.2 & 54.4 & 54.7 & 18.9 & 22.0 & 45.1 & 35.3 & 28.5 & 20.8 & 31.6 & - & 42.6 & 28.7 \\ 
Waray (\texttt{war}) & 59.6 & 25.8 & 52.1 & 38.3 & 48.0 & 48.7 & 16.9 & 19.9 & 46.9 & 33.9 & 26.6 & 19.1 & 29.7 & 49.1 & - & 26.4 \\ 
Tamil (\texttt{tam}) & 50.4 & 21.6 & 47.1 & 35.7 & 43.7 & 45.1 & 18.2 & 18.2 & 34.1 & 23.6 & 27.5 & 18.6 & 28.4 & 41.7 & 31.5 & - \\ 
\bottomrule
\end{tabular}
}
\end{table}

%% file: tables/8_flores_chrf_llama3_1_70b.tex
\begin{table}[ht]
\centering
\caption{Performance on Flores Plus (chrF++) for Llama3.1-70B}
\label{tab:matrix-chrf-llama3_1_70b}
\resizebox{\textwidth}{!}{%
\begin{tabular}{l *{16}{c}}
\toprule
\makecell[l]{\textbf{Target} ($\rightarrow$) \\ \textbf{Source} ($\downarrow$)}
 & \texttt{eng} & \texttt{cmn} & \texttt{ind} & \texttt{tha} & \texttt{vie} & \texttt{zsm} & \texttt{mya} & \texttt{lao} & \texttt{ceb} & \texttt{ilo} & \texttt{jav} & \texttt{khm} & \texttt{sun} & \texttt{fil} & \texttt{war} & \texttt{tam} \\
\midrule
English (\texttt{eng}) & - & 32.3 & 67.3 & 47.9 & 61.1 & 66.2 & 21.3 & 22.7 & 54.5 & 46.2 & 51.1 & 21.5 & 45.3 & 57.8 & 52.8 & 37.9 \\ 
Chinese (\texttt{cmn}) & 59.0 & - & 54.2 & 41.4 & 52.8 & 53.4 & 19.3 & 19.2 & 45.2 & 38.9 & 41.9 & 19.6 & 38.8 & 48.0 & 43.2 & 31.9 \\ 
Indonesian (\texttt{ind}) & 68.5 & 27.8 & - & 44.6 & 56.2 & 60.6 & 20.4 & 21.6 & 49.3 & 42.2 & 47.6 & 20.7 & 44.6 & 53.3 & 46.5 & 34.6 \\ 
Thai (\texttt{tha}) & 61.2 & 26.9 & 57.1 & - & 53.8 & 55.9 & 19.6 & 21.8 & 46.9 & 40.2 & 44.0 & 20.3 & 40.0 & 49.8 & 44.5 & 32.6 \\ 
Vietnamese (\texttt{vie}) & 63.0 & 27.2 & 57.9 & 43.6 & - & 56.6 & 19.9 & 20.9 & 47.6 & 40.2 & 44.2 & 20.5 & 41.1 & 51.1 & 44.9 & 33.3 \\ 
Malay (\texttt{zsm}) & 69.4 & 28.2 & 61.4 & 44.1 & 55.6 & - & 20.5 & 21.9 & 49.7 & 42.6 & 48.5 & 20.9 & 43.4 & 53.0 & 47.4 & 34.4 \\ 
Burmese (\texttt{mya}) & 51.0 & 21.9 & 48.5 & 37.0 & 46.0 & 48.5 & - & 16.8 & 41.0 & 35.3 & 37.9 & 18.1 & 33.9 & 45.0 & 39.6 & 31.5 \\ 
Lao (\texttt{lao}) & 49.0 & 22.0 & 49.3 & 41.4 & 45.1 & 49.6 & 17.4 & - & 42.3 & 37.1 & 39.6 & 19.5 & 35.5 & 45.3 & 40.1 & 28.5 \\ 
Cebuano (\texttt{ceb}) & 65.2 & 26.2 & 57.2 & 42.2 & 52.7 & 56.2 & 20.0 & 19.8 & - & 44.6 & 43.0 & 20.4 & 39.4 & 54.4 & 51.0 & 33.2 \\ 
Ilocano (\texttt{ilo}) & 55.0 & 23.7 & 50.6 & 37.4 & 47.3 & 50.3 & 18.8 & 18.3 & 47.2 & - & 39.5 & 18.7 & 36.5 & 49.6 & 45.7 & 31.0 \\ 
Javanese (\texttt{jav}) & 63.1 & 26.0 & 56.9 & 41.0 & 52.0 & 57.0 & 19.4 & 19.8 & 46.5 & 40.0 & - & 20.2 & 41.5 & 50.0 & 43.5 & 32.5 \\ 
Khmer (\texttt{khm}) & 55.4 & 25.2 & 53.3 & 41.4 & 50.0 & 53.0 & 18.6 & 20.2 & 44.7 & 38.1 & 41.3 & - & 37.3 & 48.3 & 41.8 & 31.0 \\ 
Sundanese (\texttt{sun}) & 60.5 & 25.2 & 57.8 & 41.2 & 51.2 & 56.1 & 19.5 & 19.6 & 45.5 & 39.7 & 46.0 & 20.2 & - & 48.3 & 43.0 & 32.5 \\ 
Tagalog (\texttt{fil}) & 69.5 & 27.9 & 60.0 & 43.5 & 55.1 & 59.1 & 20.2 & 20.2 & 52.6 & 44.8 & 45.1 & 20.5 & 41.1 & - & 50.4 & 34.2 \\ 
Waray (\texttt{war}) & 66.4 & 26.6 & 57.4 & 41.9 & 52.8 & 56.9 & 19.9 & 19.9 & 54.0 & 45.5 & 43.4 & 19.9 & 39.9 & 54.9 & - & 33.4 \\ 
Tamil (\texttt{tam}) & 58.2 & 23.8 & 53.0 & 39.3 & 49.5 & 52.7 & 19.9 & 17.3 & 44.9 & 38.6 & 40.8 & 18.5 & 37.2 & 48.4 & 42.7 & - \\ 
\bottomrule
\end{tabular}
}
\end{table}

%% file: tables/8_flores_chrf_nllb_moe_54b.tex
\begin{table}[ht]
\centering
\caption{Translation Performance (chrF++) for NLLB-MoE-54B}
\label{tab:matrix-chrf-nllb_moe_54b}
\resizebox{\textwidth}{!}{%
\begin{tabular}{l *{16}{c}}
\toprule
\makecell[l]{\textbf{Target} ($\rightarrow$) \\ \textbf{Source} ($\downarrow$)}
 & \texttt{eng} & \texttt{cmn} & \texttt{ind} & \texttt{tha} & \texttt{vie} & \texttt{zsm} & \texttt{mya} & \texttt{lao} & \texttt{ceb} & \texttt{ilo} & \texttt{jav} & \texttt{khm} & \texttt{sun} & \texttt{fil} & \texttt{war} & \texttt{tam} \\
\midrule
English (\texttt{eng}) & - & 18.9 & 65.8 & 39.8 & 58.8 & 64.7 & 30.1 & 43.2 & 54.9 & 52.5 & 52.4 & 32.6 & 42.3 & 58.9 & 55.0 & 50.2 \\ 
Chinese (\texttt{cmn}) & 56.2 & - & 50.7 & 34.8 & 49.1 & 50.4 & 26.8 & 35.0 & 44.2 & 41.5 & 40.2 & 27.6 & 36.1 & 45.3 & 40.5 & 39.0 \\ 
Indonesian (\texttt{ind}) & 66.1 & 16.1 & - & 38.0 & 53.6 & 58.6 & 26.2 & 40.7 & 49.6 & 47.0 & 47.9 & 32.0 & 42.8 & 52.3 & 48.3 & 43.6 \\ 
Thai (\texttt{tha}) & 57.7 & 16.2 & 54.4 & - & 50.6 & 53.1 & 25.2 & 39.3 & 47.0 & 44.1 & 44.0 & 30.8 & 38.6 & 47.9 & 44.0 & 39.9 \\ 
Vietnamese (\texttt{vie}) & 61.2 & 16.6 & 56.3 & 36.6 & - & 54.9 & 25.4 & 39.6 & 48.1 & 45.3 & 45.3 & 31.4 & 40.1 & 49.8 & 45.6 & 42.3 \\ 
Malay (\texttt{zsm}) & 67.1 & 17.6 & 60.4 & 38.3 & 53.2 & - & 26.7 & 41.3 & 50.7 & 47.4 & 48.4 & 32.2 & 42.3 & 52.5 & 47.9 & 45.1 \\ 
Burmese (\texttt{mya}) & 54.0 & 13.2 & 50.5 & 33.6 & 46.7 & 49.9 & - & 34.4 & 41.8 & 40.1 & 37.5 & 26.6 & 34.5 & 44.7 & 41.5 & 39.2 \\ 
Lao (\texttt{lao}) & 61.1 & 18.3 & 55.6 & 39.5 & 51.0 & 54.8 & 26.8 & - & 48.8 & 44.6 & 45.0 & 31.4 & 39.4 & 49.9 & 44.5 & 43.2 \\ 
Cebuano (\texttt{ceb}) & 67.0 & 15.7 & 57.9 & 36.5 & 52.3 & 57.5 & 25.6 & 39.9 & - & 49.5 & 47.0 & 30.2 & 39.4 & 54.2 & 51.9 & 42.0 \\ 
Ilocano (\texttt{ilo}) & 62.5 & 14.9 & 55.3 & 34.7 & 49.8 & 54.7 & 24.7 & 38.4 & 50.2 & - & 45.4 & 28.9 & 38.4 & 52.1 & 49.9 & 40.7 \\ 
Javanese (\texttt{jav}) & 61.9 & 15.9 & 57.1 & 35.1 & 50.2 & 55.3 & 27.4 & 38.7 & 49.1 & 43.9 & - & 29.1 & 41.1 & 50.4 & 43.1 & 43.0 \\ 
Khmer (\texttt{khm}) & 58.4 & 17.8 & 54.3 & 37.0 & 49.8 & 53.6 & 27.4 & 39.7 & 47.9 & 43.8 & 44.1 & - & 39.1 & 49.1 & 43.9 & 42.3 \\ 
Sundanese (\texttt{sun}) & 61.3 & 18.1 & 58.2 & 36.6 & 50.1 & 55.6 & 27.9 & 38.7 & 48.1 & 44.6 & 45.9 & 29.8 & - & 49.7 & 44.6 & 42.9 \\ 
Tagalog (\texttt{fil}) & 69.7 & 16.4 & 59.6 & 37.3 & 54.1 & 58.9 & 26.5 & 41.0 & 52.6 & 50.6 & 48.6 & 31.3 & 41.0 & - & 51.8 & 44.1 \\ 
Waray (\texttt{war}) & 69.1 & 17.1 & 59.6 & 38.2 & 54.1 & 58.9 & 27.7 & 40.7 & 52.8 & 50.6 & 47.9 & 31.2 & 40.9 & 55.3 & - & 44.3 \\ 
Tamil (\texttt{tam}) & 61.1 & 15.9 & 54.2 & 35.3 & 50.2 & 53.7 & 26.8 & 37.4 & 47.8 & 45.5 & 44.1 & 30.2 & 38.2 & 49.5 & 45.4 & - \\ 
\bottomrule
\end{tabular}
}
\end{table}

%% file: appendices/culture_eval.tex
\begin{table}[ht]
\centering
\caption{Performance Comparison on CulturalBench across Models.}
\label{tab:culturalbench_performance}
\resizebox{\textwidth}{!}{
\begin{tabular}{l *{16}{c}}
\toprule
\multirow{2}{*}{\textbf{Model}} & \multicolumn{6}{c}{\hspace{3em}\textbf{Hard}} && \multicolumn{6}{c}{\hspace{3em}\textbf{Easy}} && \multirow{2}{*}{\textbf{CulturalBench}} \\ 
\cmidrule{2-15}
 & FIL & ID & MS & SG & TH & VI & AVG & FIL & ID & MS & SG & TH & VI & AVG & \\ \midrule
SEA LION 7B & 72.22 & 58.65 & 27.27 & 66.30 & 74.07 & 45.37 & 57.31 & 11.11 & 15.38 & 18.18 & 13.04 & 7.41 & 14.81 & 13.32 & 35.32 \\ 
Sailor2-1b & 26.11 & 25.00 & 27.27 & 51.09 & 74.07 & 25.00 & 38.09 & 31.11 & 46.15 & 45.45 & 30.43 & 37.04 & 44.44 & 39.10 & 38.60 \\ 
Qwen2.5-0.5b & 26.11 & 25.00 & 27.27 & 55.43 & 74.07 & 25.00 & 38.81 & 42.22 & 53.85 & 45.45 & 43.48 & 55.56 & 33.33 & 45.65 & 42.23 \\ 
Llama-3.1-8B & 28.33 & 31.73 & 27.27 & 57.61 & 25.93 & 39.81 & 35.11 & 53.33 & 73.08 & 54.55 & 60.87 & 74.07 & 48.15 & 60.68 & 47.89 \\ 
SeaLLMs-v3-7B & 52.22 & 27.88 & 27.27 & 60.87 & 74.07 & 31.48 & 45.63 & 42.22 & 57.69 & 54.55 & 56.52 & 66.67 & 51.85 & 54.92 & 50.27 \\ 
SeaLLM-7B-Hybrid & 70.56 & 36.54 & 29.55 & 67.39 & 74.07 & 75.93 & 59.01 & 37.78 & 69.23 & 45.45 & 39.13 & 40.74 & 51.85 & 47.36 & 53.19 \\ 
Gemma-7b & 44.44 & 47.12 & 50.00 & 60.87 & 75.00 & 34.26 & 51.95 & 62.22 & 65.38 & 36.36 & 34.78 & 70.37 & 59.26 & 54.73 & 53.34 \\ 
Mistral-7b & 76.11 & 50.00 & 47.73 & 64.13 & 73.15 & 32.41 & 57.26 & 66.67 & 61.54 & 36.36 & 60.87 & 55.56 & 55.56 & 56.09 & 56.67 \\ 
Sailor-7b & 26.11 & 49.04 & 68.18 & 67.39 & 76.85 & 57.41 & 57.50 & 44.44 & 69.23 & 72.73 & 60.87 & 55.56 & 62.96 & 60.97 & 59.23 \\ 
Sailor2-8b & 48.89 & 41.35 & 34.09 & 59.78 & 80.56 & 43.52 & 51.37 & 64.44 & 69.23 & 63.64 & 60.87 & 81.48 & 74.07 & 68.96 & 60.16 \\ 
Qwen2.5-7b & 45.56 & 66.35 & 36.36 & 63.04 & 81.48 & 60.19 & 58.83 & 62.22 & 61.54 & 36.36 & 65.22 & 85.19 & 66.67 & 62.87 & 60.85 \\ 
Gemma2-9b & 76.67 & 44.23 & 38.64 & 67.39 & 68.52 & 42.59 & 56.34 & 64.44 & 69.23 & 81.82 & 60.87 & 77.78 & 70.37 & 70.75 & 63.55 \\ 
Qwen2.5-14b & 44.44 & 69.23 & 52.27 & 66.30 & 82.41 & 61.11 & 62.63 & 64.44 & 69.23 & 72.73 & 69.57 & 85.19 & 74.07 & 72.54 & 67.58 \\ 
Qwen2.5-32b & 45.56 & 57.69 & 36.36 & 72.83 & 87.04 & 55.56 & 59.17 & 68.89 & 92.31 & 72.73 & 78.26 & 92.59 & 70.37 & 79.19 & 69.18 \\ 
Gemma2-27b & 66.67 & 55.77 & 56.82 & 66.30 & 74.07 & 49.07 & 61.45 & 84.44 & 76.92 & 72.73 & 86.96 & 85.19 & 59.26 & 77.58 & 69.52 \\ 
Sailor2-20b & 65.56 & 64.42 & 54.55 & 66.30 & 83.33 & 52.78 & 64.49 & 80.00 & 84.62 & 81.82 & 73.91 & 77.78 & 77.78 & 79.32 & 71.90 \\ 
Qwen2.5-72b & 61.67 & 63.46 & 59.09 & 83.70 & 88.89 & 65.74 & 70.43 & 82.22 & 80.77 & 100.00 & 82.61 & 92.59 & 77.78 & 86.00 & 78.21 \\ 
\bottomrule
\end{tabular}
}
\end{table}

\begin{table}[ht]
\centering
\caption{Performance Comparison on BLEnD across Models.}
\label{tab:blend_performance}
\resizebox{\textwidth}{!}{
\begin{tabular}{lccccccc}
\toprule
\multicolumn{1}{l}{\textbf{Model}} & \textbf{INST4-ID} & \textbf{PERS3-ID} & \textbf{ID} & \textbf{INST4-JB} & \textbf{PERS4-JB} & \textbf{JB} & \textbf{Overall} \\
\midrule
Qwen2.5-0.5b & 22.71 & 20.83 & 21.77 & 6.11 & 5.68 & 5.90 & 13.83 \\ 
Llama-3.1-8B & 16.25 & 19.58 & 17.92 & 10.70 & 9.83 & 10.27 & 14.09 \\ 
SeaLLMs-v3-7B & 30.21 & 30.21 & 30.21 & 8.95 & 10.04 & 9.50 & 19.85 \\ 
SEA LION 7B & 33.54 & 34.17 & 33.86 & 11.79 & 11.14 & 11.47 & 22.66 \\ 
SeaLLM-7B-Hybrid & 45.83 & 46.04 & 45.94 & 18.56 & 15.07 & 16.82 & 31.38 \\ 
Qwen2.5-7b & 49.58 & 49.17 & 49.38 & 19.00 & 18.78 & 18.89 & 34.13 \\ 
Sailor2-1b & 46.67 & 44.38 & 45.53 & 22.27 & 23.80 & 23.04 & 34.28 \\ 
Mistral-7b & 52.08 & 50.42 & 51.25 & 19.21 & 17.25 & 18.23 & 34.74 \\ 
Gemma-7b & 48.33 & 46.88 & 47.61 & 25.98 & 24.24 & 25.11 & 36.36 \\ 
Sailor-7b & 56.04 & 55.21 & 55.63 & 20.96 & 21.83 & 21.40 & 38.51 \\ 
Qwen2.5-32b & 53.54 & 53.96 & 53.75 & 27.73 & 30.35 & 29.04 & 41.40 \\ 
Qwen2.5-14b & 57.50 & 56.46 & 56.98 & 27.07 & 25.55 & 26.31 & 41.65 \\ 
Gemma2-9b & 60.21 & 59.79 & 60.00 & 34.28 & 31.88 & 33.08 & 46.54 \\ 
Sailor2-8b & 65.00 & 63.33 & 64.17 & 36.68 & 35.15 & 35.92 & 50.04 \\ 
Gemma2-27b & 66.04 & 65.83 & 65.94 & 36.03 & 36.90 & 36.47 & 51.20 \\ 
Qwen2.5-72b & 67.50 & 65.42 & 66.46 & 37.55 & 36.68 & 37.12 & 51.79 \\ 
Sailor2-20b & 68.75 & 66.67 & 67.71 & 38.86 & 38.21 & 38.54 & 53.12 \\ 
\bottomrule
\end{tabular}
}
\end{table}

\begin{table}[ht]
\centering
\caption{Performance Comparison on Global-MMLU across Models.}
\label{tab:global_mmlu_performance}
\begin{tabular}{lccccc}
\toprule
\multicolumn{1}{l}{\textbf{Model}} & \textbf{ID} & \textbf{FIL} & \textbf{MS} & \textbf{VI} & \textbf{Overall} \\ 
\midrule
SEA LION 7B & 24.83 & 27.03 & 24.75 & 25.97 & 25.65 \\ 
Sailor2-1b & 33.56 & 31.26 & 32.40 & 32.52 & 32.44 \\ 
Qwen2.5-0.5b & 36.03 & 27.75 & 33.91 & 36.04 & 33.43 \\ 
SeaLLM-7B-Hybrid & 37.66 & 36.61 & 36.04 & 35.61 & 36.48 \\ 
Sailor-7b & 43.03 & 30.64 & 43.23 & 41.08 & 39.50 \\ 
SeaLLMs-v3-7B & 43.65 & 36.55 & 40.19 & 42.73 & 40.78 \\ 
Mistral-7b & 45.73 & 42.85 & 43.51 & 39.99 & 43.02 \\ 
Llama-3.1-8B & 52.94 & 47.59 & 50.08 & 50.68 & 50.32 \\ 
Gemma-7b & 52.53 & 52.66 & 52.66 & 49.47 & 51.83 \\ 
Sailor2-8b & 59.34 & 57.50 & 56.70 & 56.71 & 57.56 \\ 
Qwen2.5-7b & 61.34 & 53.52 & 57.39 & 60.85 & 58.28 \\ 
Gemma2-9b & 60.72 & 60.54 & 59.63 & 57.05 & 59.49 \\ 
Gemma2-27b & 66.57 & 66.40 & 65.12 & 62.54 & 65.16 \\ 
Qwen2.5-14b & 69.83 & 61.66 & 64.66 & 66.74 & 65.72 \\ 
Sailor2-20b & 69.98 & 68.06 & 68.32 & 66.71 & 68.27 \\ 
Qwen2.5-32b & 74.13 & 67.73 & 70.95 & 72.53 & 71.34 \\ 
Qwen2.5-72b & 76.41 & 74.84 & 77.07 & 77.68 & 76.50 \\ 
\bottomrule
\end{tabular}
\end{table}